\documentclass[twocolumn]{svjour3}       %

\usepackage[utf8]{inputenc}  
\usepackage[T1]{fontenc}
\usepackage{times}
\usepackage{epsfig}
\usepackage{graphicx}
\usepackage{amsmath}
\usepackage{amssymb}
\usepackage[table,xcdraw]{xcolor}
\usepackage[caption=false,font=footnotesize]{subfig}
\usepackage{booktabs}
\usepackage{multirow}
\usepackage{setspace}
\usepackage{stackengine}
\usepackage{microtype}
\usepackage{textcomp}
\usepackage{anyfontsize}
\usepackage{psfrag}
\usepackage{adjustbox}
\usepackage{ulem} 
\usepackage{array}
\newcolumntype{C}[1]{>{\centering\arraybackslash}p{#1}}
\newcolumntype{L}[1]{>{\raggedright\arraybackslash}p{#1}}

\usepackage[pagebackref=true,breaklinks=true,letterpaper=true,colorlinks,bookmarks=false]{hyperref}

\begin{document}

\title{Rain rendering for evaluating and improving robustness to bad weather}

\author{
Maxime Tremblay  \and
Shirsendu Sukanta Halder \and
Raoul de Charette \and \\
Jean-François Lalonde
}

\date{Received: date / Accepted: date}

\maketitle

\begin{abstract}
Rain fills the atmosphere with water particles, which breaks the common assumption that light travels unaltered from the scene to the camera. While it is well-known that rain affects computer vision algorithms, quantifying its impact is difficult. In this context, we present a rain rendering pipeline that enables the systematic evaluation of common computer vision algorithms to controlled amounts of rain. We present three different ways to add synthetic rain to existing images datasets: completely physic-based; completely data-driven; and a combination of both. The physic-based rain augmentation combines a physical particle simulator and accurate rain photometric modeling. We validate our rendering methods with a user study, demonstrating our rain is judged as much as 73\% more realistic than the state-of-the-art. Using our generated rain-augmented KITTI, Cityscapes, and nuScenes datasets, we conduct a thorough evaluation of object detection, semantic segmentation, and depth estimation algorithms and show that their performance decreases in degraded weather, on the order of 15\% for object detection, 60\% for semantic segmentation, and 6-fold increase in depth estimation error. Finetuning on our augmented synthetic data results in improvements of 21\% on object detection, 37\% on semantic segmentation, and 8\% on depth estimation.

\end{abstract}

\section{Introduction}

\begin{figure}	
	\centering
	\footnotesize
	\setlength{\tabcolsep}{0.0025\linewidth}
	\renewcommand{\arraystretch}{0.5}
	\begin{tabular}{ccc}
		\multirow{2}{*}[1cm]{\rotatebox{90}{Object detection~\cite{yang2016exploit}}} &
		\adjincludegraphics[width=0.4775\linewidth,trim={{0.1938\width} 0 0 0},clip]{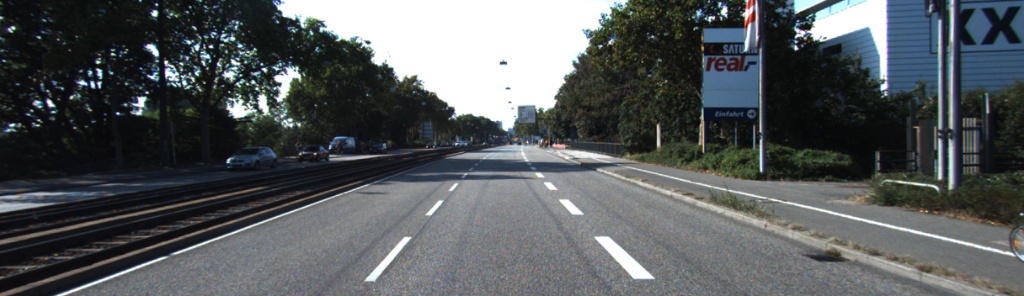} &
		\adjincludegraphics[width=0.4775\linewidth,trim={{0.1938\width} 0 0 0},clip]{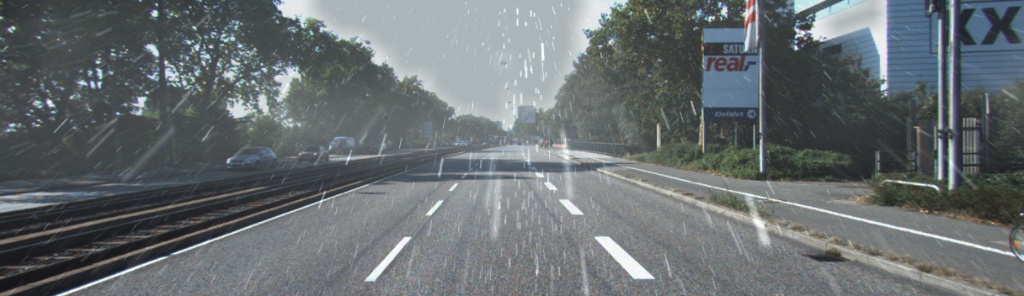}\\
		& 
		\adjincludegraphics[width=0.4775\linewidth,trim={{0.1938\width} 0 0 0},clip]{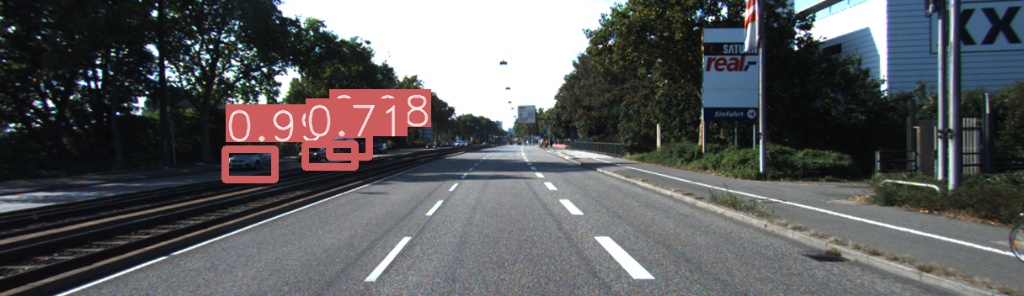} &
		\adjincludegraphics[width=0.4775\linewidth,trim={{0.1938\width} 0 0 0},clip]{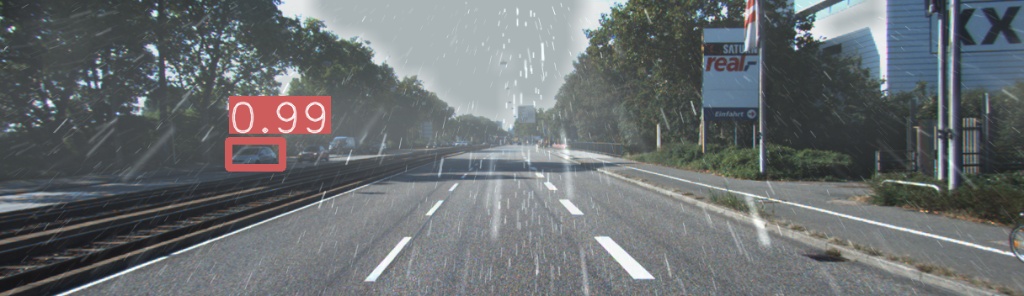}\\
		\midrule
		\multirow{2}{*}[1.5cm]{\rotatebox{90}{Semantic segmentation~\cite{mehta2018espnet}}} &
		\adjincludegraphics[width=0.4775\linewidth,trim={0 {0.110728\height} 0 0},clip]{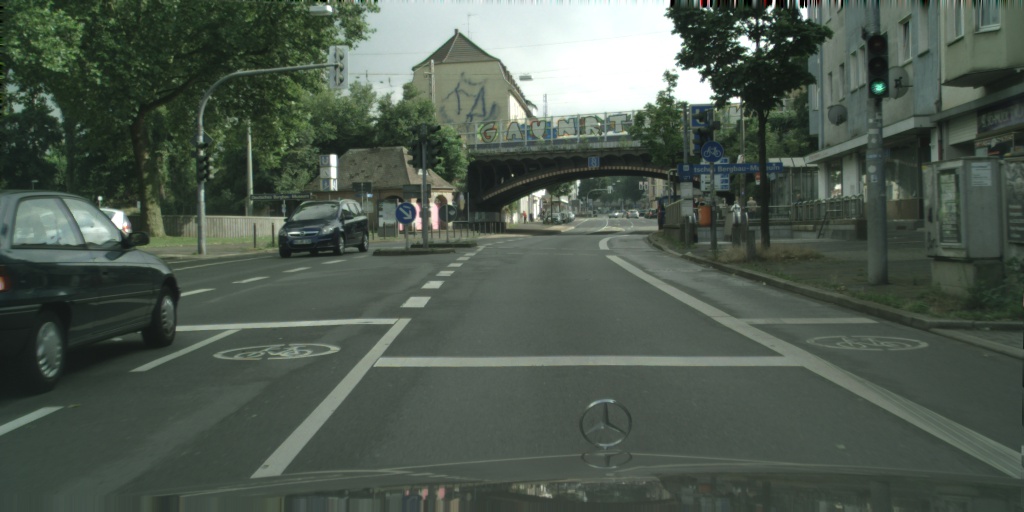} &
		\adjincludegraphics[width=0.4775\linewidth,trim={0 {0.110728\height} 0 0},clip]{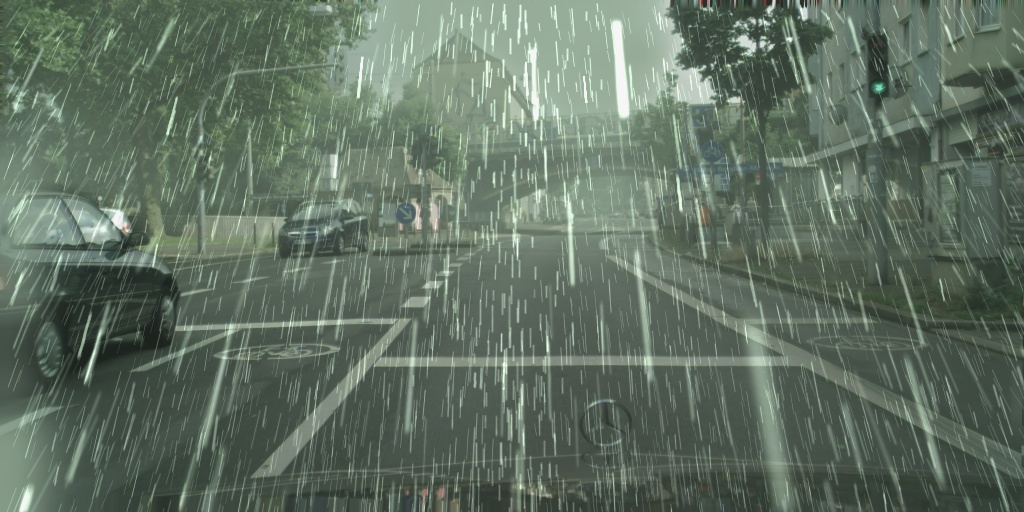}\\
		& 
		\adjincludegraphics[width=0.4775\linewidth,trim={0 {0.110728\height} 0 0},clip]{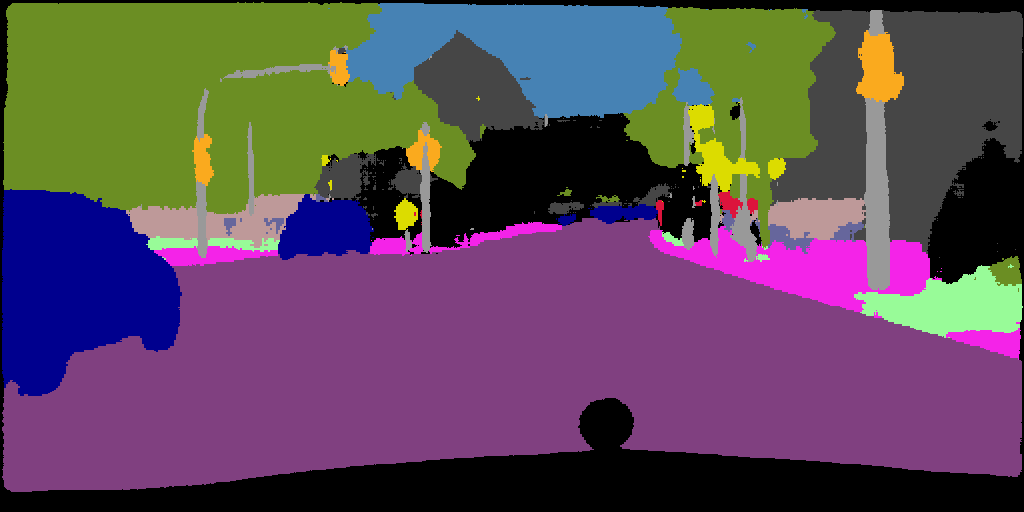} &
		\adjincludegraphics[width=0.4775\linewidth,trim={0 {0.110728\height} 0 0},clip]{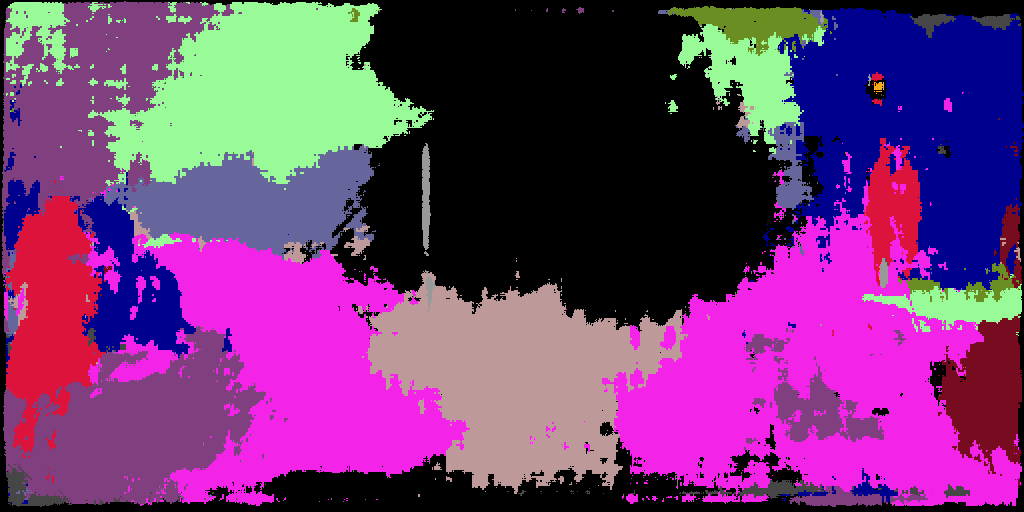}\\
		\midrule
		\multirow{2}{*}[1.15cm]{\rotatebox{90}{Depth estimation~\cite{godard2019digging}}} &
		\adjincludegraphics[width=0.4775\linewidth,trim={0 {0.1\height} 0 {0.1\height}},clip]{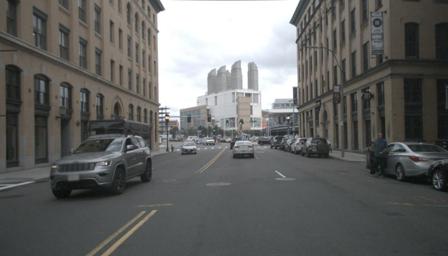} &
		\adjincludegraphics[width=0.4775\linewidth,trim={0 {0.1\height} 0 {0.1\height}},clip]{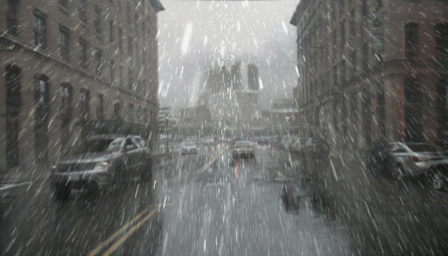}\\
		&
		\adjincludegraphics[width=0.4775\linewidth,trim={0 {0.1\height} 0 {0.1\height}},clip]{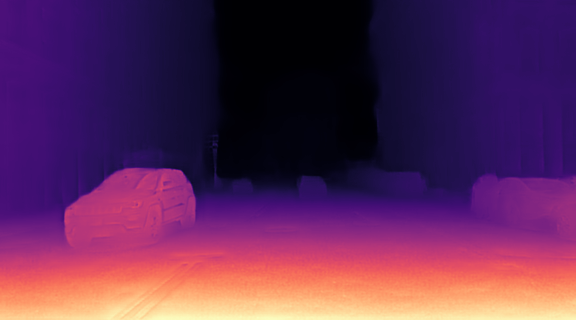} &
		\adjincludegraphics[width=0.4775\linewidth,trim={0 {0.1\height} 0 {0.1\height}},clip]{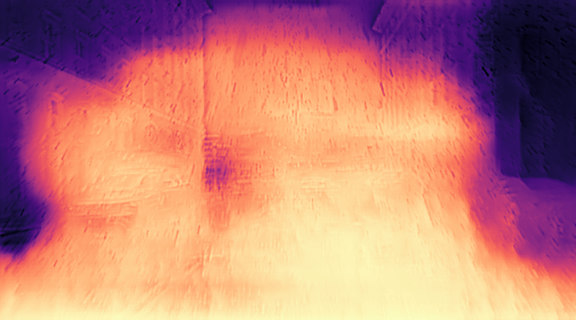}\\
		&
		Clear weather &
		Rain (200~mm/hr) \\
	\end{tabular}
	\vspace{.25em}
	\caption{\textbf{Vision tasks in clear and rain-augmented images.} Our synthetic rain rendering framework allows for the evaluation of computer vision algorithms in challenging bad weather scenarios.  We render physically-based, realistic rain on images from the KITTI~\cite{Geiger2012CVPR} (rows 1-2) and Cityscapes~\cite{Cordts2016Cityscapes} (rows 3-4) datasets with object detection from mx-RCNN~\cite{yang2016exploit} (row 2), semantic segmentation from ESPNet~\cite{mehta2018espnet} (row 4). We also present a combined data-driven and physic-based rain rendering approach which we apply to the nuScenes~\cite{caesar2019nuscenes} (rows 5-6) dataset with depth estimation from Monodepth2~\cite{godard2019digging} (row 6). All algorithms are quite significantly affected by rainy conditions.}
	\label{fig:overview}
	\vspace{-1em}
\end{figure}

A common assumption in computer vision is that light travels unaltered from the scene to the camera. In clear weather, this assumption is reasonable: the atmosphere behaves like a transparent medium and transmits light with very little attenuation or scattering. However, inclement weather conditions such as rain fill the atmosphere with particles producing spatio-temporal artifacts such as attenuation or rain streaks. This creates noticeable changes to the appearance of images (see fig.~\ref{fig:overview}), thus creating additional challenges to computer vision algorithms which must be robust to these conditions. 

While the influence of rain on image appearance is well-known and understood~\cite{garg2005does}, its impact on the performance of computer vision tasks is not.
Indeed, how can one evaluate what the impact of, say, a rainfall rate of 100~mm/hour (a typical autumn shower) is on the performance of an object detector, when our existing databases all contain images overwhelmingly captured under clear weather conditions? To measure this effect, one would need a labeled object detection dataset where all the images have been captured under 100~mm/hour rain. Needless to say, such a "rain-calibrated" dataset does not exist, and capturing one is prohibitive. Indeed, datasets with bad weather information are few and sparse, and typically include only high-level tags (rain or not) without mentioning \emph{how much} rain is falling. While they can be used to improve vision algorithms under adverse conditions by including rainy images in the training set, they cannot help us in systematically evaluating performance degradation under increasing amounts of rain.

Alternatively, one can attempt to remove its effects from images---i.e., create a ``clear weather'' version of the image---prior to applying subsequent algorithms. For example, rain can be detected and attenuated from images~\cite{garg2007vision,barnum2010analysis,yang2017deep,zhang2019image,liu2019dual}. We experiment with this approach in sec.~\ref{sec:deraining}. An alternative approach is to employ programmable lighting to reduce the rain visibility, by shining light between raindrops~\cite{de2012fast}. Unfortunately, these solutions either add significant processing times to already constrained time budgets, or require custom hardware. Instead, if we could systematically study the effect of weather on images, we could better understand the robustness of existing algorithms, and, potentially, increase their robustness afterwards. 

In this paper, we propose methods to realistically \emph{augment} existing image databases with rainy conditions. 
We rely on well-understood physical models as well as on recent image-to-image translations to generate visually convincing results. First, we experiment with our novel physics-based approach, which is the first to allow controlling the \emph{amount} of rain in order to generate arbitrary amounts, ranging from very light rain (5~mm/hour rainfall) to very heavy storms (200+~mm/hour). 
This key feature allows us to produce weather-augmented datasets, where the rainfall rate is known and calibrated. Subsequently, we augment two existing datasets (KITTI~\cite{Geiger2012CVPR} and Cityscapes~\cite{Cordts2016Cityscapes}) with rain, and evaluate the robustness of popular object detection and segmentation algorithms on these augmented databases. Second, we experiment with a combination of physics- and learning-based approaches, where a popular unpaired image-to-image translation method~\cite{zhu2017unpaired} is used to convey a sense of ``wetness'' to the scene, and physics-based rain is subsequently composited on the resulting image. Here, we augment the nuScenes dataset~\cite{caesar2019nuscenes}, and use it to evaluate the robustness of object detection and depth estimation algorithms. 
Finally, we also use the latter to refine algorithms using curriculum learning~\cite{bengio2009curriculum}, and demonstrate improved robustness on real rainy images.

In short, we make the following contributions. 
First, we present two different realistic rain rendering approaches: the first is a purely physic-based method and the second is a combination of a GAN-based approach and this physic-based framework. 
Second, we augment KITTI~\cite{Geiger2012CVPR}, Cityscapes~\cite{Cordts2016Cityscapes}, and nuScenes~\cite{caesar2019nuscenes} datasets with rain. 
Third, we present a methodology for systematically evaluating the performance of 13 popular algorithms---for object detection, semantic segmentation and depth estimation---on rainy images. Our findings indicate that rain affects \emph{all} algorithms: performance drops of 15\% mAP for object detection, 60\% AP for semantic segmentation, and a 6-fold increase in depth estimation error. 
Finally, our augmented database can also be used to finetune these same algorithms in order to \emph{improve} their performance in real-world rainy conditions. 

This paper significantly extends an earlier version of this work published in~\cite{halder2019physics}, by combining physics-based rendering with learning-based image-to-image translation methods, conducting a novel, more in-depth user study, evaluating depth estimation algorithms, comparing to a deraining approach, and by providing a more extensive evaluation of the performance improvement on real images. Our framework is readily usable to augment existing image with realistic rainy conditions. Code and data are available at the following URL: \url{https://team.inria.fr/rits/computer-vision/weather-augment/}.

\section{Related work}

\paragraph{Rain modeling} In their series of influential papers, Garg and Nayar provided a comprehensive overview of the appearance models required for understanding~\cite{garg2007vision} and synthesizing~\cite{garg2006photorealistic} realistic rain. In particular, they propose an image-based rain streak database~\cite{garg2006photorealistic} modeling the drop oscillations, which we exploit in our physics-based rendering framework. 
Other streak appearance models were proposed in~\cite{weber2015multiscale,barnum2010analysis} using a frequency model. Realistic rendering was also obtained with ray-tracing~\cite{rousseau2006realistic} or artistic-based techniques~\cite{tatarchuk2006artist,creus2013r4} but on synthetic data as they require complete 3D knowledge of the scene including accurate light estimation. Numerous works also studied generation of raindrops on-screen, with 3D modeling and ray-casting~\cite{roser2009video,roser2010realistic,halimeh2009raindrop,hao2019learning} or normal maps~\cite{porav2019can} some also accounting for focus blur.

\paragraph{Rain removal} Due to the problems it creates on computer vision algorithms, rain removal in images got a lot of attention initially focusing on photometric models~\cite{garg2004detection}. For this, several techniques have been proposed, ranging from frequency space analysis~\cite{barnum2010analysis} to deep networks~\cite{yang2017deep}. Sparse coding and layers priors were also important axes of research~\cite{li2016rain,luo2015removing,chen2013generalized} due to their facility to encode streak patches. Recently, dual residual networks have been employed~\cite{liu2019dual}.
Alternatively, camera parameters~\cite{garg2005does} or programmable light sources~\cite{de2012fast} can also be adjusted to limit the impact of rain on the image formation process. 
Additional proposals were made for the specific task of raindrops removal on windows~\cite{eigen2013restoring} or windshields~\cite{halimeh2009raindrop,porav2019can,hao2019learning}.

\paragraph{Unpaired image translation} An interesting solution to weather augmentation is the use of data-driven unpaired image translation frameworks. Zhang et al.~\cite{zhang2019image} proposed to use conditional GANs for rain removal. By proposing to add a cyclic loss to the learning process, CycleGAN~\cite{zhu2017unpaired} became a significant paper for unpaired image translation; they produced interesting results in weather and season translation. DualGAN~\cite{yi2017dualgan} uses similar ideas with differences in the network models. The UNIT~\cite{liu2017unsupervised}, MUNIT~\cite{huang2018multimodal}, and FUNIT~\cite{liu2019few} frameworks all, in one way or another, propose to perform image translation with the common idea that data from different sets have a shared latent space. They showed interesting results on adding and removing weather effect to images. 
Since the information in clear and rainy images is symmetrical, many unsupervised image translation approaches could produce decent visual results. 
In \cite{pizzati2020model}, Pizzati et al. learn to disentangle the scene from lens occlusions such as raindrops, which improves both realism and physical accuracy of the translations.
Another strategy for better qualitative translations is to rely on semantic consistency~\cite{li2018semantic,tasar2020semi2i}. 

\paragraph{Weather databases}

In computer vision, few images databases have precise labeled weather information. 
Of note for mobile robotics, the BDD100K~\cite{yu2018bdd100k}, the Oxford dataset~\cite{RobotCarDatasetIJRR}, and Wilddash~\cite{zendel2018wilddash} provide data recorded in various weather conditions, including rain. 
Other stationary camera datasets such as AMOS~\cite{jacobs07amos}, the transient attributes dataset~\cite{Laffont14}, the Webcam Clip Art dataset~\cite{lalonde2009webcam}, or the WILD dataset~\cite{narasimhan2002all} are sparsely labeled with weather information. 
The relatively new nuScenes dataset~\cite{caesar2019nuscenes} have multiple labeled scenes containing rainy images, but variation in rain intensities are not indicated. Gruber et al.~\cite{gruber-3dv-19} recently released a dataset with dense depth labels under a variety of real weather conditions produced by a controlled weather chamber, which inherently limits the variety of scenes (limited to four common scenarios) in the dataset. Note that \cite{bijelic2019seeing} also announced---but at the time of writing, not yet fully available---a promising dataset including heavy snow and rain events.
Still, datasets with rainy data are too small to train algorithms and there exists no dataset with systematically recorded rainfall rates and object/scene labels. 
The closest systematic works in spirit~\cite{johnson2016driving,khan2019procsy} evaluated the effect of simulating weather on vision, but did so in purely virtual environments (GTA and CARLA, respectively) whereas we augment real images.
Of particular relevance to our work, Sakaridis et al.~\cite{SDV18} propose a framework for rendering fog into images from the Cityscapes~\cite{Cordts2016Cityscapes} dataset. Their approach assumes a homogeneous fog model, which is rendered from the depth estimated from stereo. Existing scene segmentation models and object detectors are then adapted to fog. 
In our work, we employ the similar idea of rendering realistic weather on top of existing images, but we focus on rain rather than fog.

\section{Rain Augmentation}
\label{sec:rain_aug}

\begin{figure}
\centering
\includegraphics[width=\linewidth]{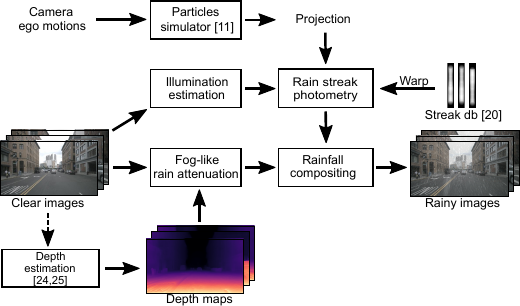}
\caption{\textbf{Physics-Based Rendering for rain augmentation.} We use particles simulation together with depth and illumination estimation to render arbitrarily controlled rainfall on clear images.}
\label{fig:pipeline}
\end{figure}

Broadly speaking, synthesizing rain on images can be achieved using two seemingly antagonistic methods: 1)~physics-based rendering~(PBR) methods~\cite{halimeh2009raindrop,garg2006photorealistic}, which explicitly model the dynamics and the radiometry of rain drops in images; or 2)~learning-based image-to-image translation approaches~\cite{zhu2017unpaired,huang2018multimodal}, which train deep neural networks to ``translate'' an image into its rainy version. While completely different, we argue both these methods offer complementary advantages. On one hand, physics-based approaches are accurate, controllable, can simulate a wide variety of imaging conditions, and do not require any training data. On the other hand, learning-based approaches can realistically simulate important visual cues such as wetness, cloud cover, and overall gloominess typically associated with rainy images. 

In this paper, we propose to first explore the use of both techniques independently, then to \emph{combine} them into a hybrid approach. This section thus first describes our PBR approach~(sec.~\ref{sec:rain_pbr}), followed by the image-to-image translation with a GAN~(sec.~\ref{sec:rain_gan}), and concludes with the hybrid combination of the two GAN+PBR~(sec.~\ref{sec:rain_hybrid}).

\subsection{Physics-Based Rendering (PBR)}
\label{sec:rain_pbr}

Taking inspiration from the vast literature on rain physics, \cite{marshall1948distribution,van1997numerical,garg2007vision,narasimhan2002vision} we simulate the rain appearance in an arbitrary image with the approach summarized in fig.~\ref{fig:pipeline}. Based on the estimated scene depth, a fog-like attenuation layer is first generated. Individual rain streaks are subsequently generated, and composited with the fog-like layer. The final result is blended in the original image to create a realistic, physics-based and controllable rainfall rate. 
 
\subsubsection{Fog-like rain}
\label{sec:rendering.fog-like}

Following the definition of \cite{garg2007vision}, fog-like rain is the set of drops that are too far away and that project on an area smaller than 1 pixel. In this case, a pixel may even be imaging a large number of drops, which causes optical attenuation~\cite{garg2007vision}.
In practice, most drops in a rainfall are actually imaged as fog-like rain\footnote{Assuming a stationary camera with KITTI calibration~\cite{Geiger2013IJRR}, we computed that only 1.24\% of the drops project on 1+ pixel in 50~mm/hr rain, and 0.7\% at 5~mm/hr. This follows logic: the heavier the rain, the higher the probability of having large drops.}, though their visual effect is less dominant.

We render the volumetric attenuation using the model described in \cite{weber2015multiscale} where the per-pixel attenuation $I_\text{att}$ is expressed as the sum of the extinction $L_\text{ext}$ caused by the volume of rain and the airlight scattering $A_\text{in}$ that results of the environmental lighting. Using equations from \cite{weber2015multiscale} to model the attenuation image at pixel $\mathbf{x}$ we get
\begin{equation}
\label{eq:render-foglike}
I_\text{att}(\mathbf{x}) = I L_\text{ext}(\mathbf{x}) + A_\text{in}(\mathbf{x}) \,,
\end{equation}
where
\begin{equation}
\begin{split}
L_\text{ext}(\mathbf{x}) &= e^{-0.312R^{0.67}d(\mathbf{x})}  \,, \\
A_\text{in}(\mathbf{x})  &= \beta_\text{HG}(\theta) \bar{E}_\text{sun} (1 - L_\text{ext}(\mathbf{x}))  \,.
\end{split}
\end{equation}
Here, $R$ denotes the rainfall rate $R$ (in mm/hr), $d(\mathbf{x})$ the pixel depth, $\beta_\text{HG}$ the standard Heynyey-Greenstein coefficient, and $\bar{E}_\text{sun}$ the average sun irradiance which we estimate from the image-radiance relation~\cite{horn1986robot}.

\begin{figure}[!t]
 \centering
 \footnotesize
 \subfloat[Drop FOV]{\includegraphics[width=0.24\linewidth]{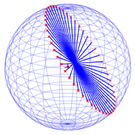}\label{fig:drop_cam_fov_fov}}\hspace{0.05\linewidth}
 \subfloat[Environment map estimation~\cite{Cameron2005}]{\includegraphics[width=0.7\linewidth]{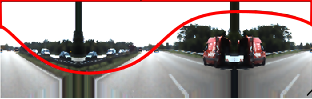}\label{fig:drop_cam_fov_envmap}}
 \vspace{.25em}
 \caption{
     \textbf{Estimation of raindrop photometry.} To estimate the photometric radiance of a drop, we integrate the lighting environment map over the 165$^{\circ}$ drop field of view \protect\subref{fig:drop_cam_fov_fov} relying on an estimate of the environment map $E$ shown in \protect\subref{fig:drop_cam_fov_envmap}. The projected field of view ($F$) of the drop is outlined in red.
 }
 \label{fig:drop_cam_fov}
\end{figure}

\subsubsection{Simulating the physics of raindrops}
\label{sec:raindrop-position}

We use the particles simulator of \cite{de2012fast} to compute the position and dynamics of all raindrops greater than 1~mm for a given fall rate\footnote{The distribution and dynamics of drops vary on earth due to gravity and atmospheric conditions. We selected here the broadly used physical models recorded in Ottawa, Canada~\cite{marshall1948distribution,atlas1973doppler}.}. 
The simulator outputs the position and dynamics (start and end points of streaks) of all visible rain drops in both world and image space, and accounts for intrinsic and extrinsic camera calibration.

\subsubsection{Rendering the appearance of a rain streak}
\label{sec:rwa_appearance}

While ray casting allows exact modeling of drops photometry, this comes at very high processing cost and is virtually only possible in synthetic scenes where the geometry and surface materials are perfectly known~\cite{rousseau2006realistic,halimeh2009raindrop}. What is more, drops oscillate as they fall, which creates further complications in modeling the light interaction. Instead, we rely on the raindrop appearance database of Garg and Nayar~\cite{garg2007vision}, which contains the individual rain streaks radiance when imaged by a stationary camera. For each drop, the streak database also models 10 oscillations due to the airflow, which accounts for much greater realism than Gaussian modeling~\cite{barnum2010analysis}. 

To render a raindrop, we first select a rain streak $S \in \mathcal{S}$ from the streak database $\mathcal{S}$ of \cite{garg2006photorealistic}, which contains 20 different streaks (each with 10 different oscillations) stored in an image format. 
We select the streak that best matches the final drop dimensions (computed from the output of the physical simulator), and randomly select an oscillation. 

The selected rain streak $S$ is subsequently warped to match the drop dynamics from the physical simulator: 
\begin{equation}
S' = \mathcal{H}(S) \,,
\label{eq:drop_warp}
\end{equation}
where $\mathcal{H}(\cdot)$ is the homography computed from the start and end points in image space given by the physical simulator and the corresponding points in the database streak image.

\subsubsection{Computing the photometry of a rain streak}
\label{sec:photometry-rainstreak}

Computing the photometry of a rain streak from a single image is impractical because drops have a much larger field of view than common cameras ($165^\circ$ vs approx. 70--100$^\circ$). To render a drop accurately, we must therefore estimate the environment map (spherical lighting representation) around that drop. 
Sophisticated methods could be used~\cite{holdgeoffroy-cvpr-17,holdgeoffroy-cvpr-19,zhang-cvpr-19} but we employ~\cite{Cameron2005} which approximates the environment map through a series of simple operations on the image. 

From each camera relative 3D drop position, we compute the intersection $F$ of the drop field of view with the environment map $E$, assuming a 10m constant scene distance.
The process is depicted in fig.~\ref{fig:drop_cam_fov}, and geometrical details are provided in appendix~\ref{sec:app-geom-drop-fov}.
Note that geometrically exact drop field of view estimation requires location-dependent environment maps, centered on each drop. However, we consider the impact negligible since drops are relatively close to the camera center compared to the sphere radius used\footnote{We computed that, for KITTI, 98.7\% of the drops are within 4~m from the camera center in a 50~mm/hr rainfall rate. Therefore, computing location-dependent environment maps would not be significantly more accurate,  while being of very high processing cost.}. 

Since a drop refracts 94\% of its field of view radiance and reflects 6\% of the entire environment map radiance~\cite{garg2007vision}, we multiply the streak appearance with a per-channel weight: 
\begin{equation}
S' = S' (0.94 \bar{F} + 0.06 \bar{E}) \,,
\label{eq:photometry_rainstreak}
\end{equation}
where $\bar{F}$ is the mean of the intersection region $F$, and $\bar{E}$ is the mean of the environment map $E$.

\begin{figure}[!t]
 \centering
 \footnotesize
 \setlength{\tabcolsep}{0.02cm}
 \renewcommand{\arraystretch}{0.5}
 \begin{tabular}{ccc}
  \multirow{1}{*}[1.35cm]{\rotatebox{90}{Ground truth}}&\includegraphics[width=0.31\linewidth, height=0.19\linewidth]{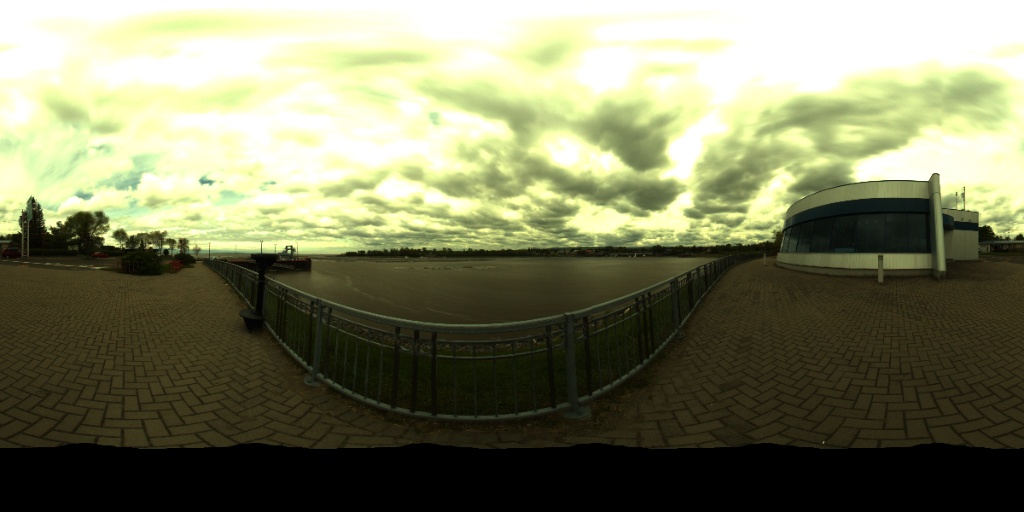}&\includegraphics[width=0.65\linewidth]{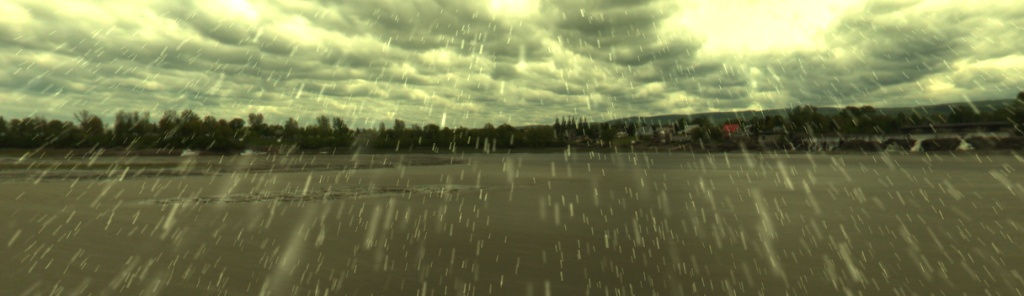}\\
  \multirow{1}{*}[0.85cm]{\rotatebox{90}{Ours}}&\includegraphics[width=0.31\linewidth, height=0.19\linewidth]{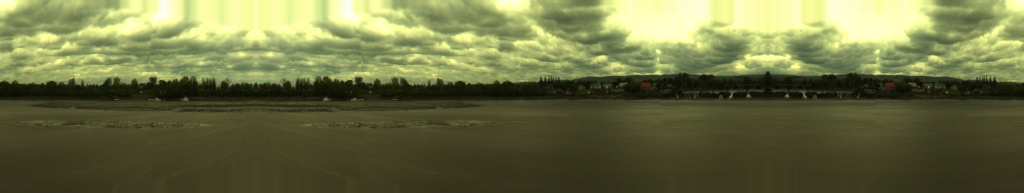}&\includegraphics[width=0.65\linewidth]{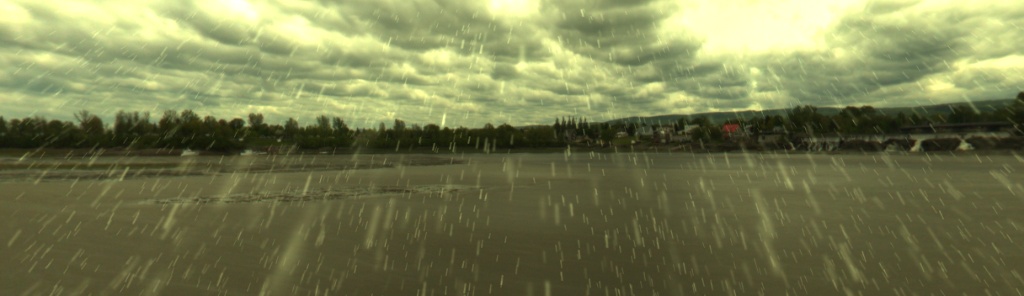}\vspace{0.05in}\\
  \multirow{1}{*}[1.35cm]{\rotatebox{90}{Ground truth}}&\includegraphics[width=0.31\linewidth, height=0.19\linewidth]{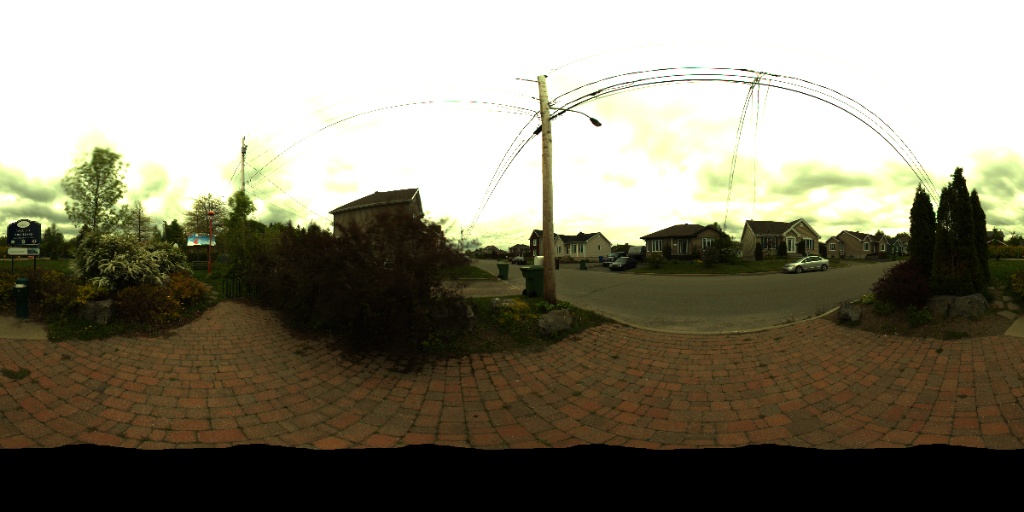}&\includegraphics[width=0.65\linewidth]{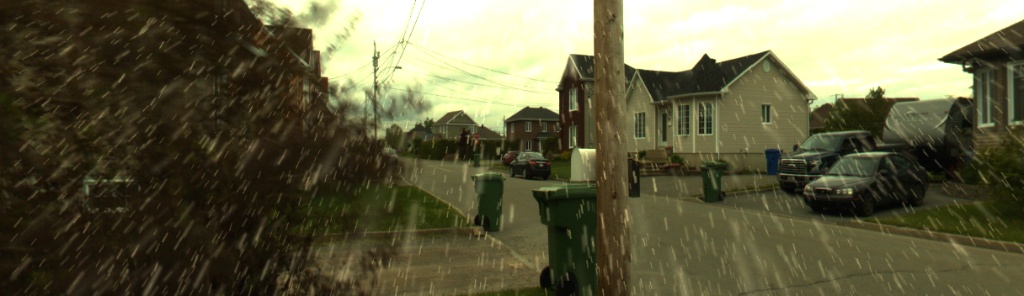}\\
  \multirow{1}{*}[0.85cm]{\rotatebox{90}{Ours}}&\includegraphics[width=0.31\linewidth, height=0.19\linewidth]{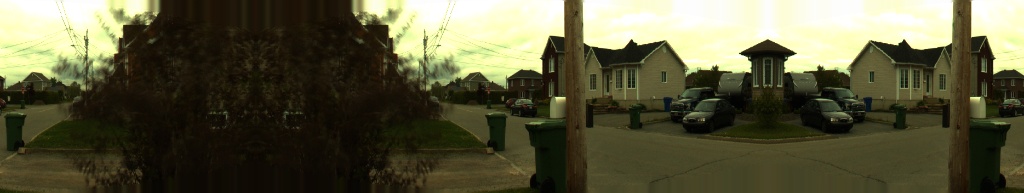}&\includegraphics[width=0.65\linewidth]{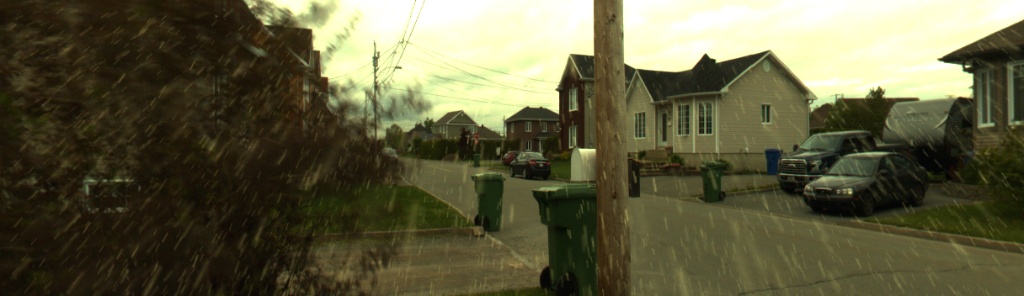}\\
  & Environment map & Our synthesized rain (50mm/hr) \\
 \end{tabular}
 \caption{\textbf{Photometric validation of rain.} Rain rendering using ground truth illumination or our approximated environment map. 
 From HDR panoramas~\cite{holdgeoffroy-cvpr-19}, we first extract limited field of view crops to simulate the point of view of a regular camera. Then, 50mm/hr rain is rendered using either (rows 1, 3) the ground truth HDR environment map or (rows 2, 4) our environment estimation. The environment maps are shown as reference on the left. While our approximated environment maps differ from the ground truth, they are sufficient to generate visually similar rain in images.}
 \label{fig:pano_ours_comparison}
\end{figure}

\subsubsection{Compositing a single rain streak on the image}
\label{sec:rendering.streak-compositing}

Now that the streak position and photometry were determined from the physical simulator and the environment map respectively, we can composite it onto the original image. 
First, to account for the camera depth of field, we apply a defocus effect following~\cite{potmesil1981lens}, convolving the streak image $S'$ with the circle of confusion\footnote{The circle of confusion $C$ of an object at distance $d$, is defined as: $C = \frac{(d - f_{\text{p}}) f^2}{d(f_{\text{p}} - f) f_{\text{N}}}$ with $f_\text{p}$ the focus plane, $f$ the focal and $f_\text{N}$ the lens f-number. $f$ and $f_\text{N}$ are from intrinsic calibration, and $f_\text{p}$ is set to 6~m.} $C$, that is: $S' = S' * C$. 

We then blend the rendered drop with the attenuated background image $I_{\text{att}}$ using the photometric blending model from \cite{garg2007vision}. Because the streak database and the image $I$ are likely to be imaged with different exposures, we need to correct the exposure to match the imaging system used in~$I$.
Suppose a pixel $\mathbf{x}$ of the image $I$ and $\mathbf{x}'$ the overlapping coordinates in streak $S'$, the composite is obtained with
\begin{equation}
I_{\text{rain}}(\mathbf{x}) = \frac{T - S'_{\alpha}(\mathbf{x'})\tau_1}{T}I_{\text{att}}(\mathbf{x}) + S'(\mathbf{x'})\frac{\tau_1}{\tau_0}\,,
\label{eq:rain_blending}
\end{equation}
where $S'_{\alpha}$ is the alpha channel\footnote{While \cite{garg2006photorealistic} does not provide an alpha channel, the latter is easily computed since drops were rendered on black background in a white ambient lighting.} of the rendered streak, $\tau_0 = \sqrt{10^{-3}} / 50$ is the time for which the drop remained on one pixel in the streak database, and $\tau_1$ the same measure according to our physical simulator. We refer to appendix~\ref{sec:app-comp_diff_exp} for details.

\subsubsection{Compositing rainfall on the image}

The rendering of rainfall of arbitrary rates in an image is done in three main steps:
1)~the fog-like attenuated image $I_{\text{att}}$ is rendered (eq.~\ref{eq:render-foglike}), 
2)~the drops outputted by the physical simulator are rendered individually on the image (eq.~\ref{eq:rain_blending}),
3)~the global luminosity average of the rainy image denoted $I_{\text{rain}}$ is adjusted. 
While rainy events usually occur in cloudy weather which consequently decreases the scene radiance, a typical camera imaging system adjusts its exposure to restore the luminosity. Consequently, we adjust the global luminosity factor so as to restore the mean radiance, and preserves the relation $\bar{I} = \bar{I}_{\text{rain}}$, where the overbar denotes the intensity average.

\paragraph{Photometric validation.} A limitation of our physical pipeline is the lighting estimation which impacts the photometry of the rain. 
To measure its effect, in fig.~\ref{fig:pano_ours_comparison} we compare the same rain rendered with our estimated environment map or ground truth illumination obtained from high dynamic range panoramas~\cite{holdgeoffroy-cvpr-19}. 
Overall, our estimation differs from ground truth when the scene is not radially symmetric but we observe that it produces visually similar rain in images.

\subsection{Image-to-image translation (GAN)}
\label{sec:rain_gan}

While our physic-based rendering generates realistic rain streaks and fog-like rain effect, it ignores major rainy characteristics such as wetness, reflections, clouds and thus may fail at conveying the overall look of a rainy scene. Conversely, generative adversarial networks (GAN) excel at learning such visual characteristics as they constitute strong signals for the discriminator in the learning process.

We further learn the $\text{clear}\mapsto{}\text{rain}$ mapping with CycleGAN~\cite{zhu2017unpaired} from a set of unpaired clear/rain images.
We train our model with the $256\times256$ architecture from~\cite{zhu2017unpaired} on images of input size $448\times256$. The generator is similar to~\cite{johnson2016perceptual} with 2 downsampling blocks followed by 9 ResNet blocks and 2 upsampling blocks. The discriminator is a simple 3 hidden layers ConvNet similar to the one used in PatchGAN~\cite{isola2017image}. The model is optimized for 40 epochs with Adam, using batch size of 1, a learning rate of 0.0002, and $\beta = \left\{ 0.5, 0.999 \right\}$. 

\subsection{Combining GAN and PBR}
\label{sec:rain_hybrid}

\begin{figure}
 \centering
 \includegraphics[width=\linewidth]{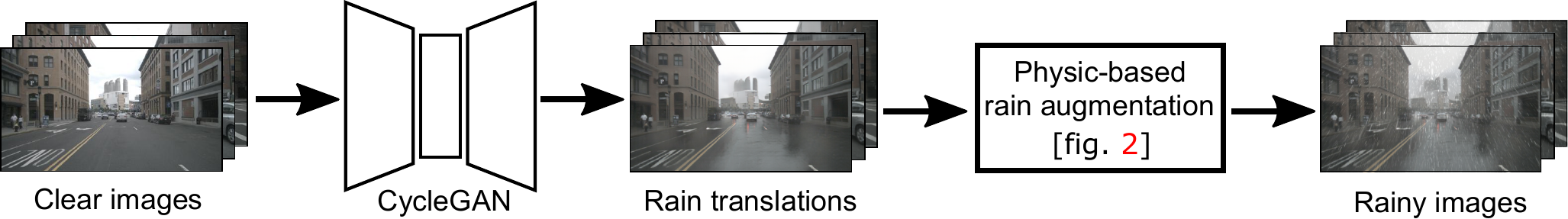}
 \caption{\textbf{GAN+PBR rain-augmentation architecture.} In this hybrid approach, clear images are first translated into rain with CycleGAN~\cite{zhu2017unpaired} and subsequently augmented with rain streaks with our PBR pipeline (see fig.~\ref{fig:pipeline}).}
 \label{fig:hybrid_architecture}
\end{figure}

We combine both the PBR- and the GAN-based rain generation methods together, simply by first translating the image to its rainy version with the GAN, then compositing the rain layer onto the resulting image using PBR (see fig.~\ref{fig:hybrid_architecture}). The sun irradiance estimation $\bar{E}_{sun}$ (sec.~\ref{sec:rendering.fog-like}) of the ``translated'' image is typically darker, which, in turn, makes the fog-like rain more realistic. The estimated environment map is also darker and, consequently, so is its mean value $\bar{E}$. The appearance of rain streaks will thus remain coherent with their environment.

Since rain streaks smaller than 1px in diameter are ignored before the rendering and instead generated as fog-like rain (sec.~\ref{sec:rendering.fog-like}), we need to apply the PBR renderer at full resolution by upsampling the output of the GAN to the image original size. Once the rain rendering is complete, we downsample the augmented image at $448\times256$.
We further refer to this hybrid rendering as GAN+PBR.

\section{Validating rain appearance}

We now validate the appearance of our synthetic rainy images when using either of our rain augmentation pipelines.
We observe visual results and quantify their perceptual realism by comparing them to existing rain augmentation approaches.

\subsection{Qualitative evaluation}
\begin{figure}
 \footnotesize
 \setlength{\tabcolsep}{0.025cm}
 \renewcommand{\arraystretch}{0.5}
 \begin{tabular}{ccc}
  \multicolumn{3}{c}{\textbf{Rain photographs}} \\
  & \adjincludegraphics[width=0.44\linewidth,trim={0 {0.50\height} 0 0},clip]{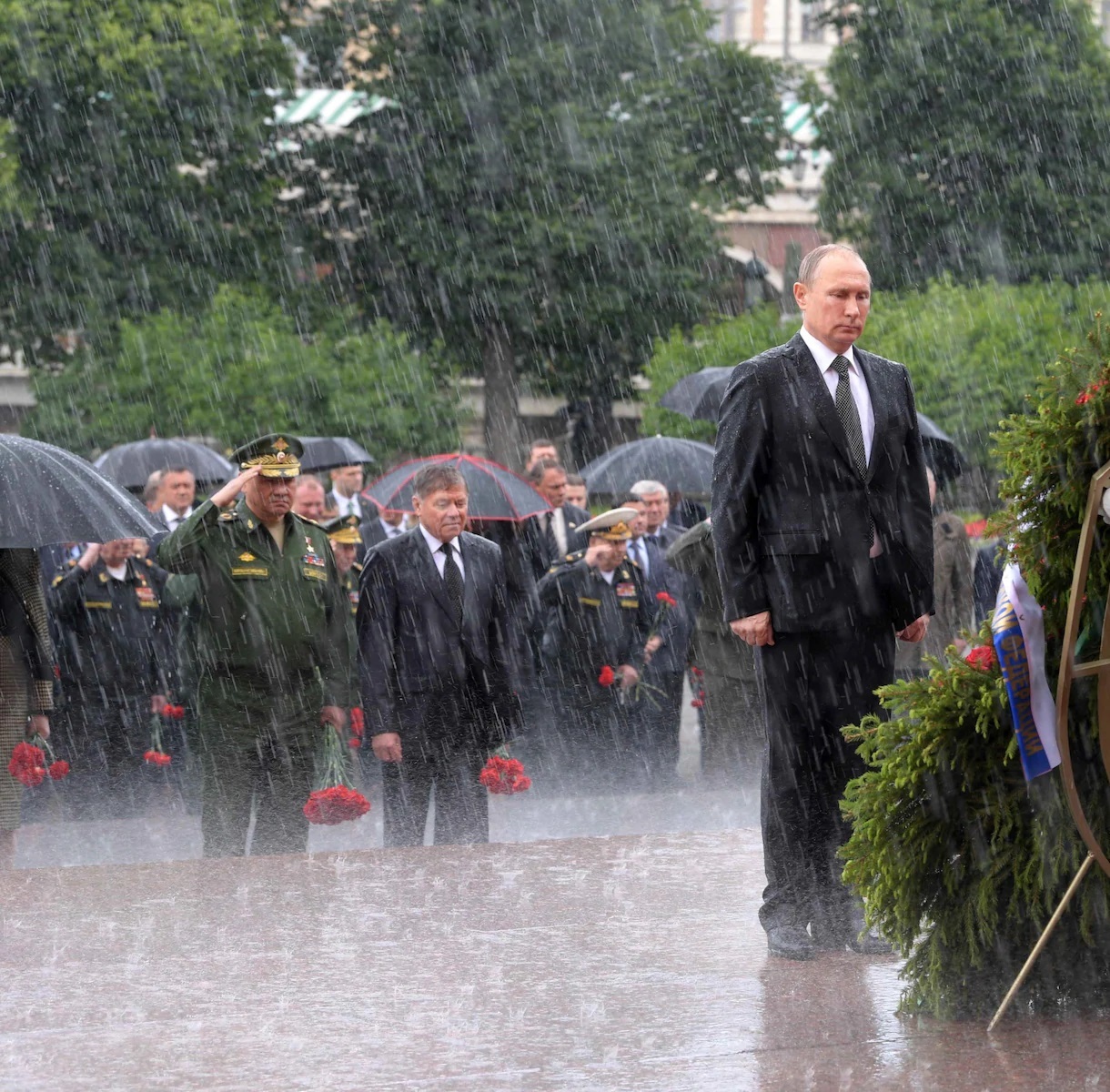} 
  & \adjincludegraphics[width=0.44\linewidth,trim={0 {0.39\height} 0 0},clip]{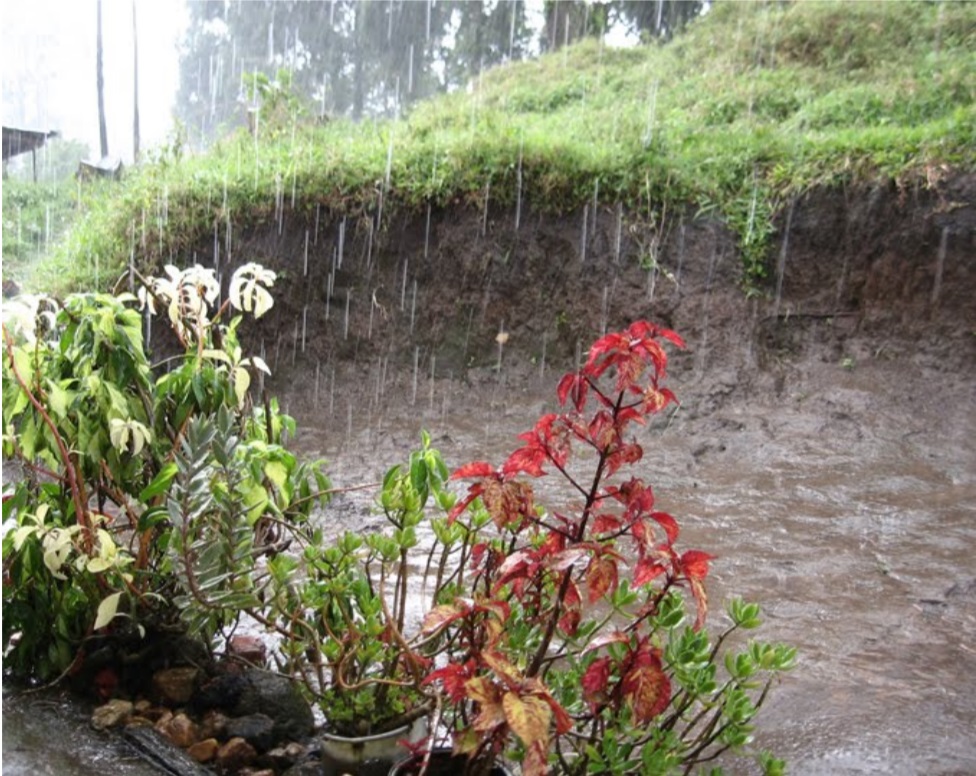}\\
  & \adjincludegraphics[width=0.44\linewidth,trim={{0.1\width} {0.15\height} {0.1\width} {0.15\height}},clip]{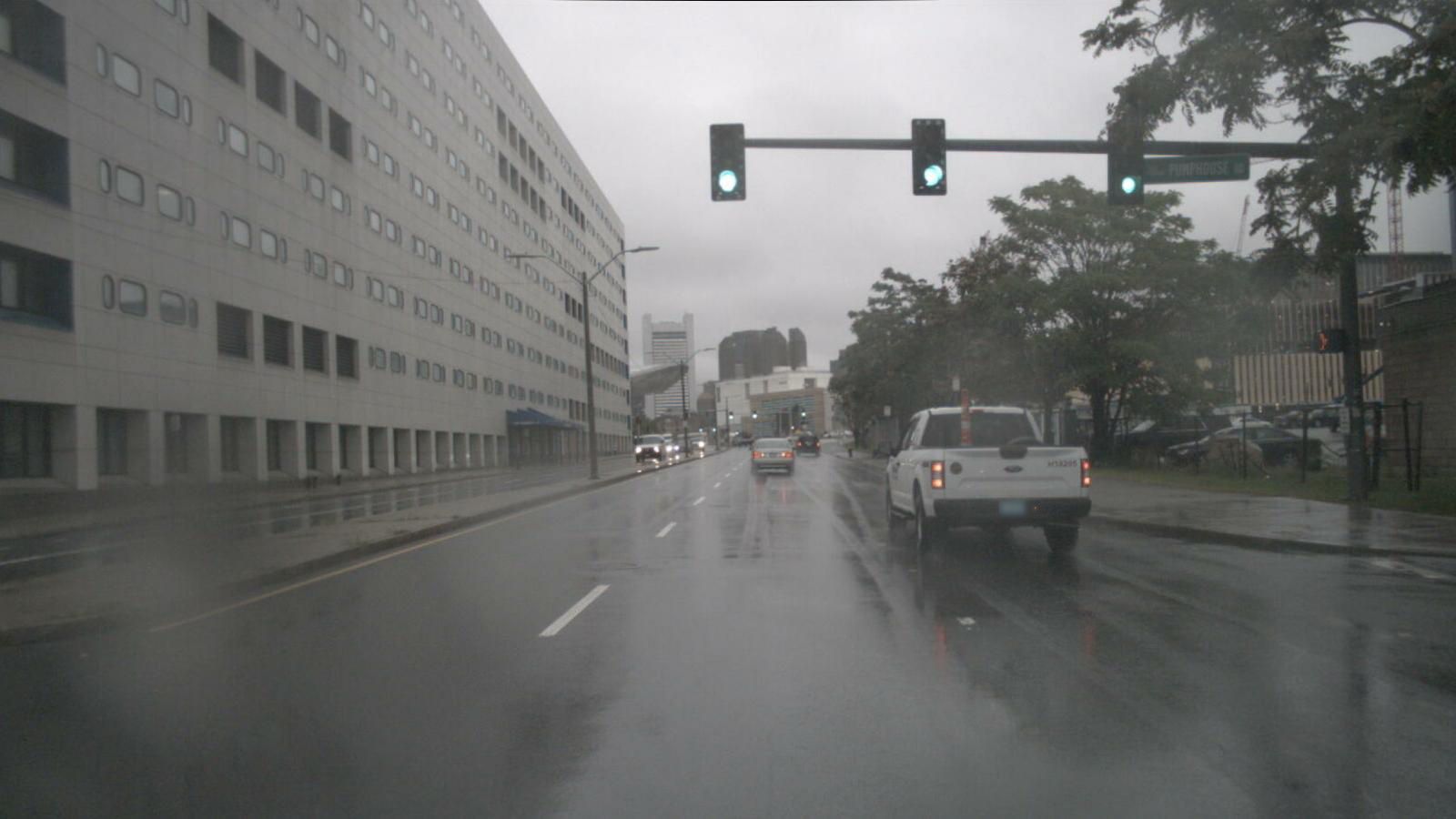} 
  & \adjincludegraphics[width=0.44\linewidth,trim={{0.1\width} {0.05\height} {0.1\width} {0.30\height}},clip]{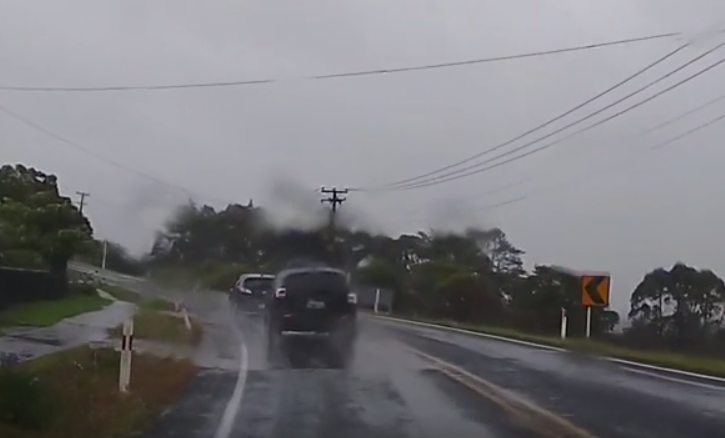}\\
  \toprule
  \multicolumn{3}{c}{\textbf{Our rain rendering}} \\
  \multirow{1}{*}[1.2cm]{\rotatebox{90}{~} \rotatebox{90}{Clear}} & \adjincludegraphics[width=0.45\linewidth,trim={{0.1\width} {0.1\height} {0.1\width} {0.1\height}},clip]{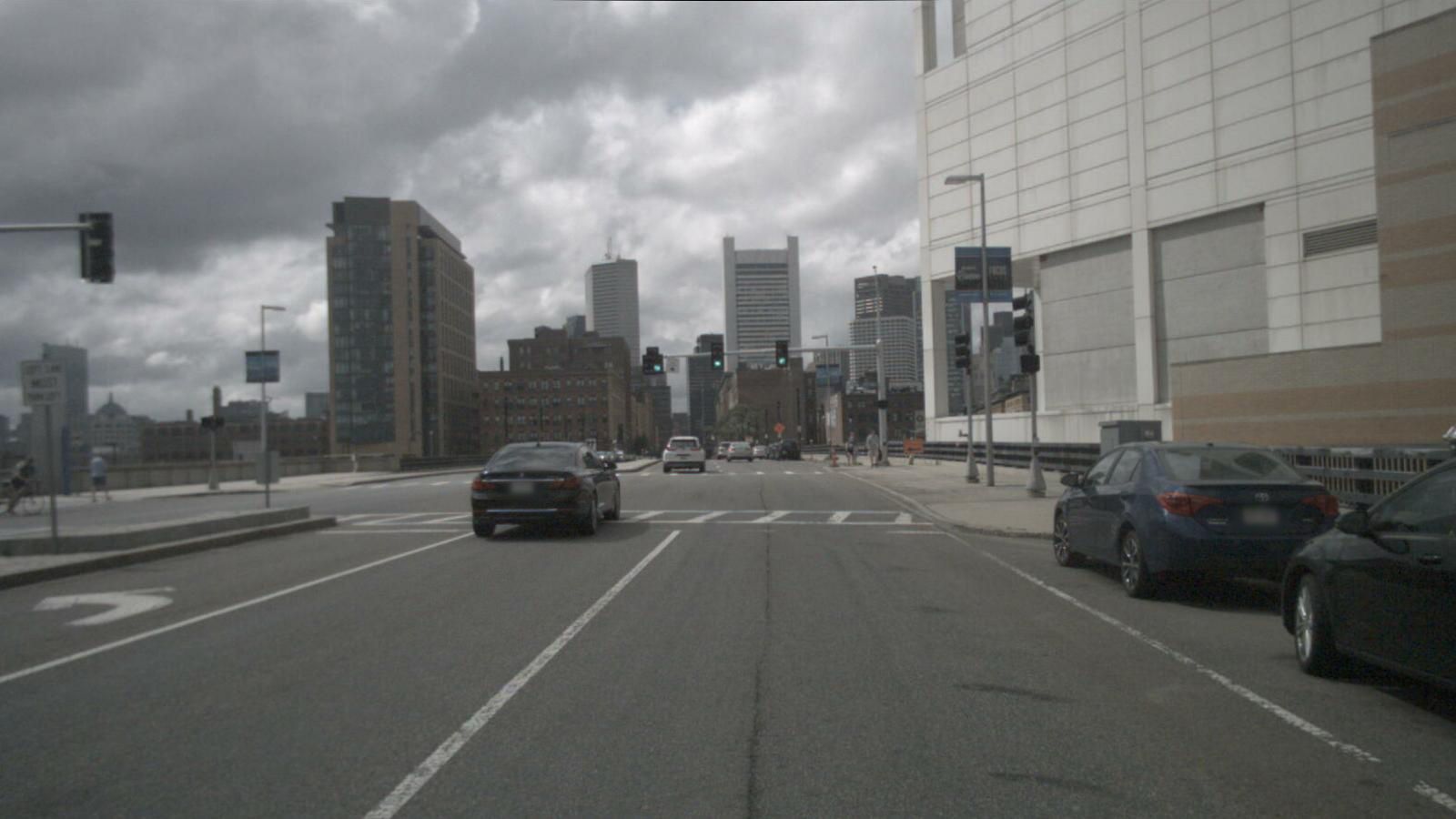} & \adjincludegraphics[width=0.45\linewidth,trim={{0.1\width} {0.1\height} {0.1\width} {0.1\height}},clip]{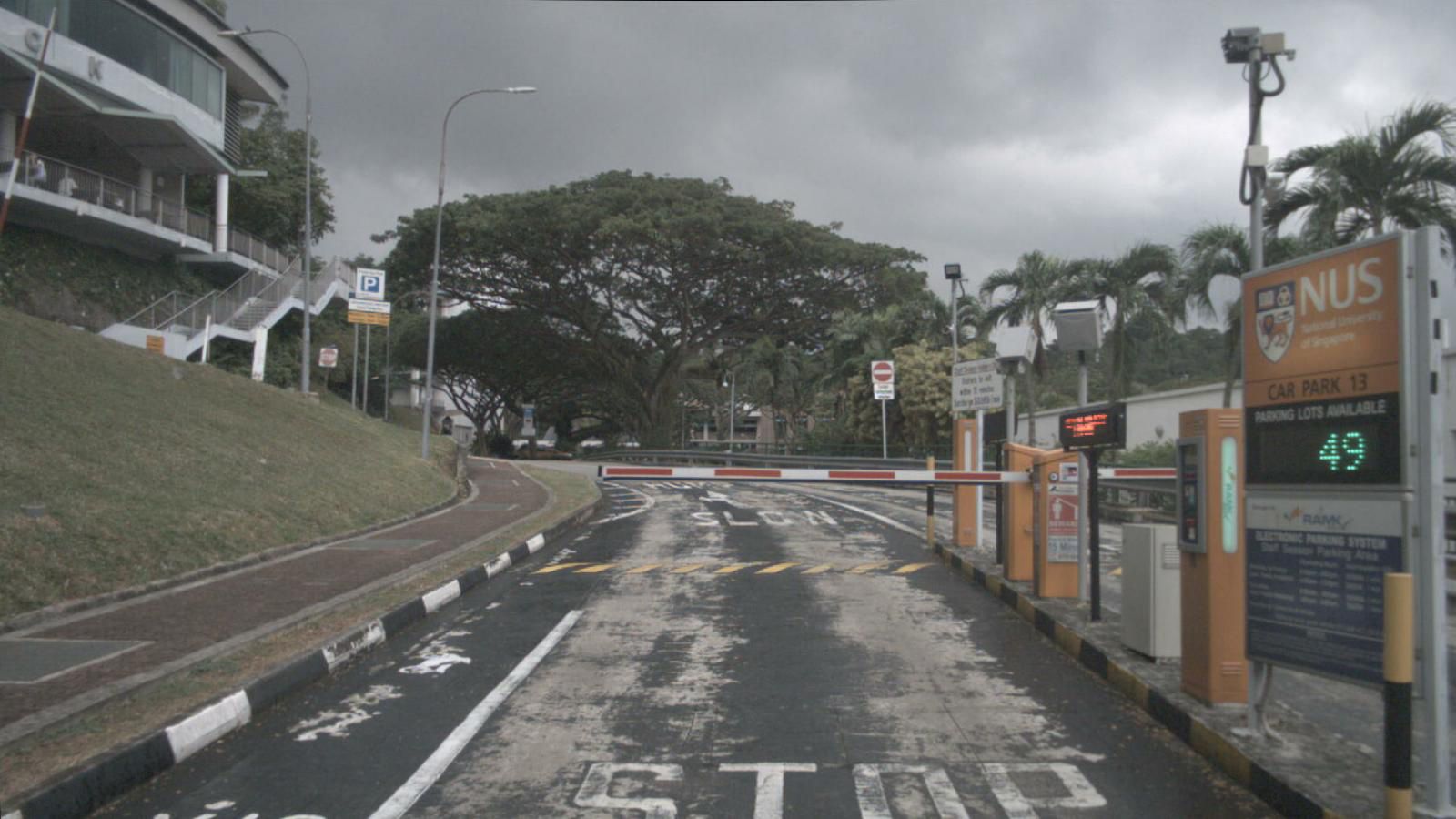} \\
  \multirow{1}{*}[1.6cm]{\rotatebox{90}{~~~~~PBR} \rotatebox{90}{100mm/hr}} & \adjincludegraphics[width=0.45\linewidth,trim={{0.1\width} {0.1\height} {0.1\width} {0.1\height}},clip]{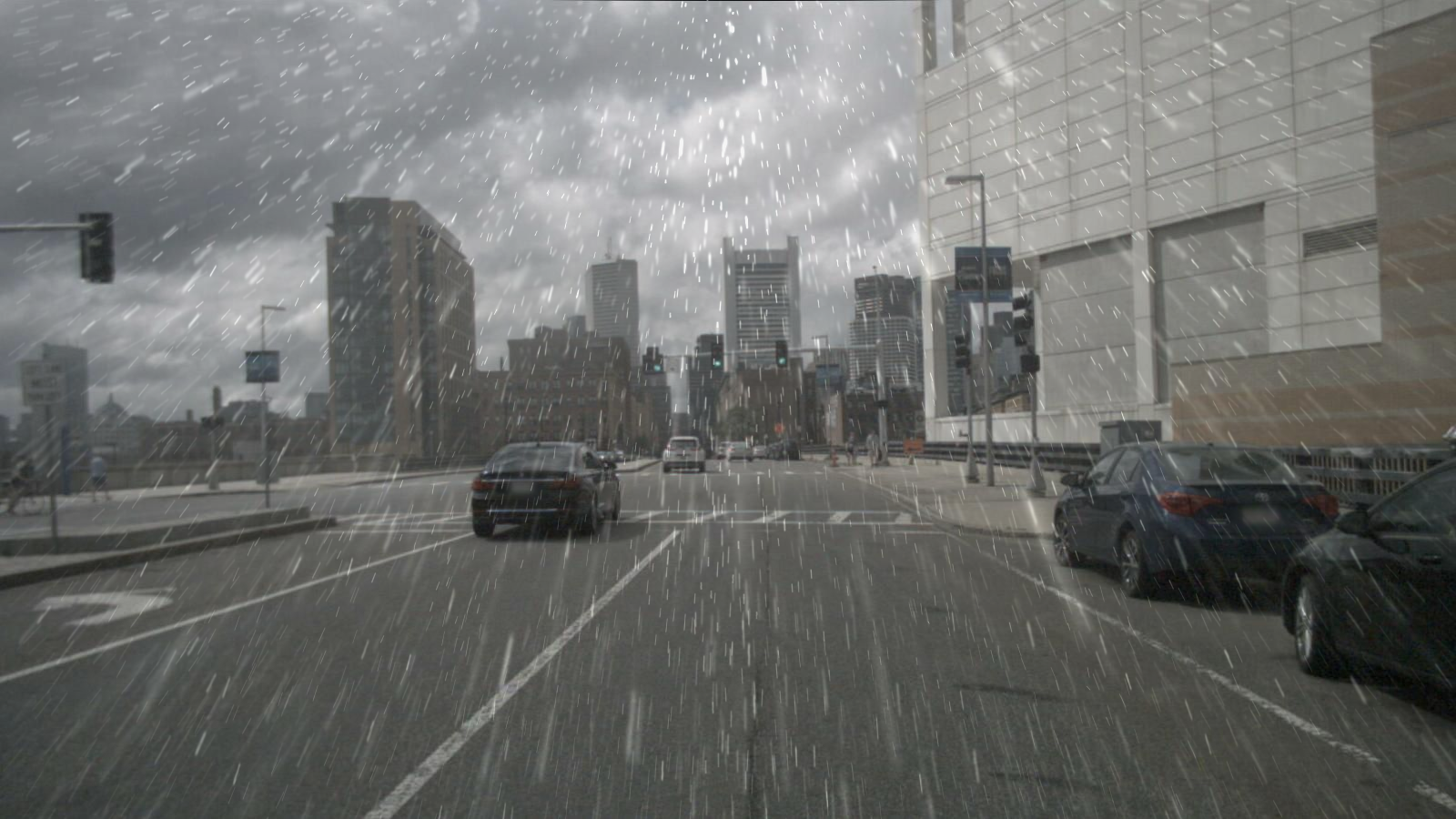} & \adjincludegraphics[width=0.45\linewidth,trim={{0.1\width} {0.1\height} {0.1\width} {0.1\height}},clip]{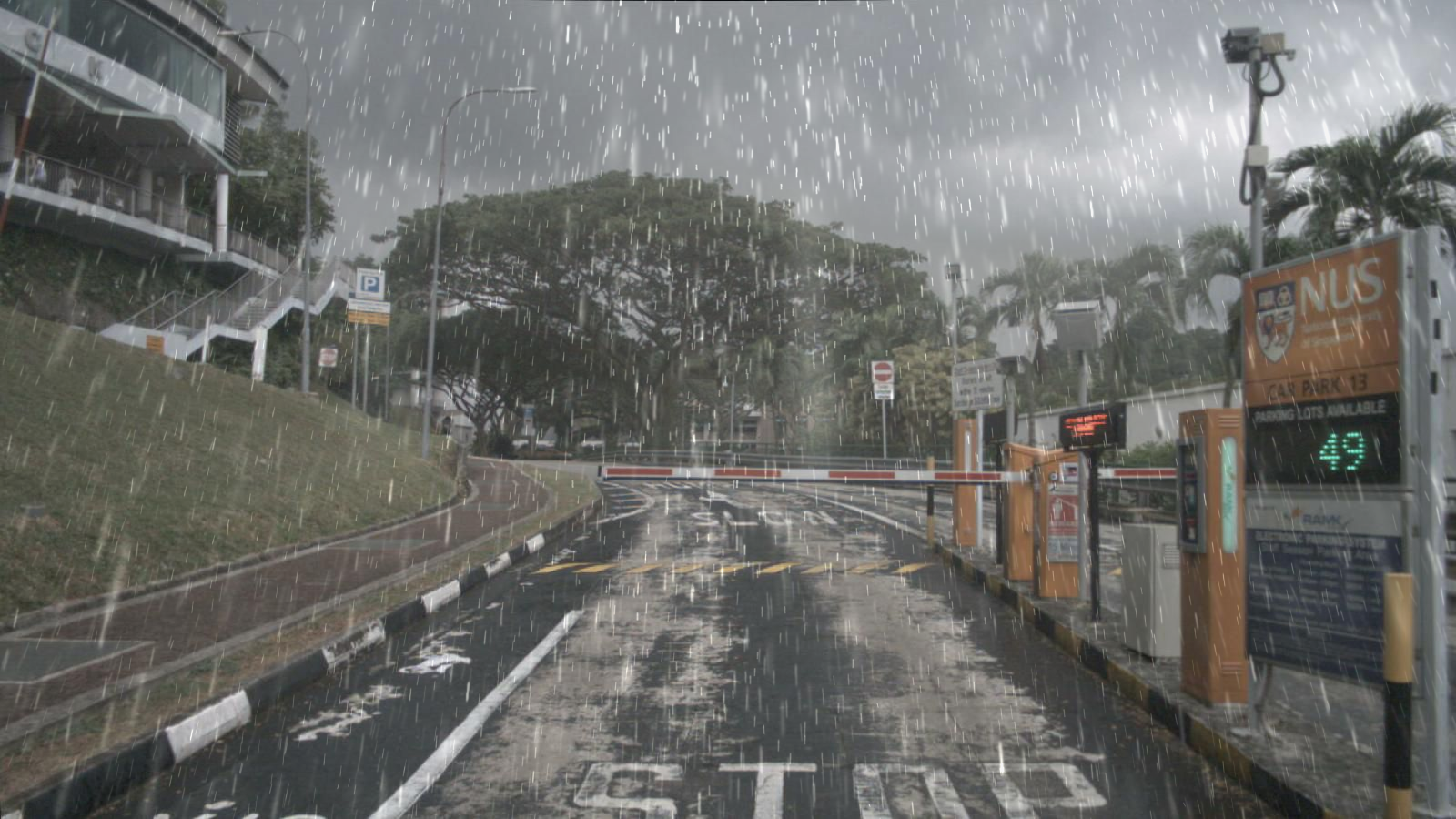} \\
  \multirow{1}{*}[1.6cm]{\rotatebox{90}{~~~~~PBR} \rotatebox{90}{200mm/hr}} & \adjincludegraphics[width=0.45\linewidth,trim={{0.1\width} {0.1\height} {0.1\width} {0.1\height}},clip]{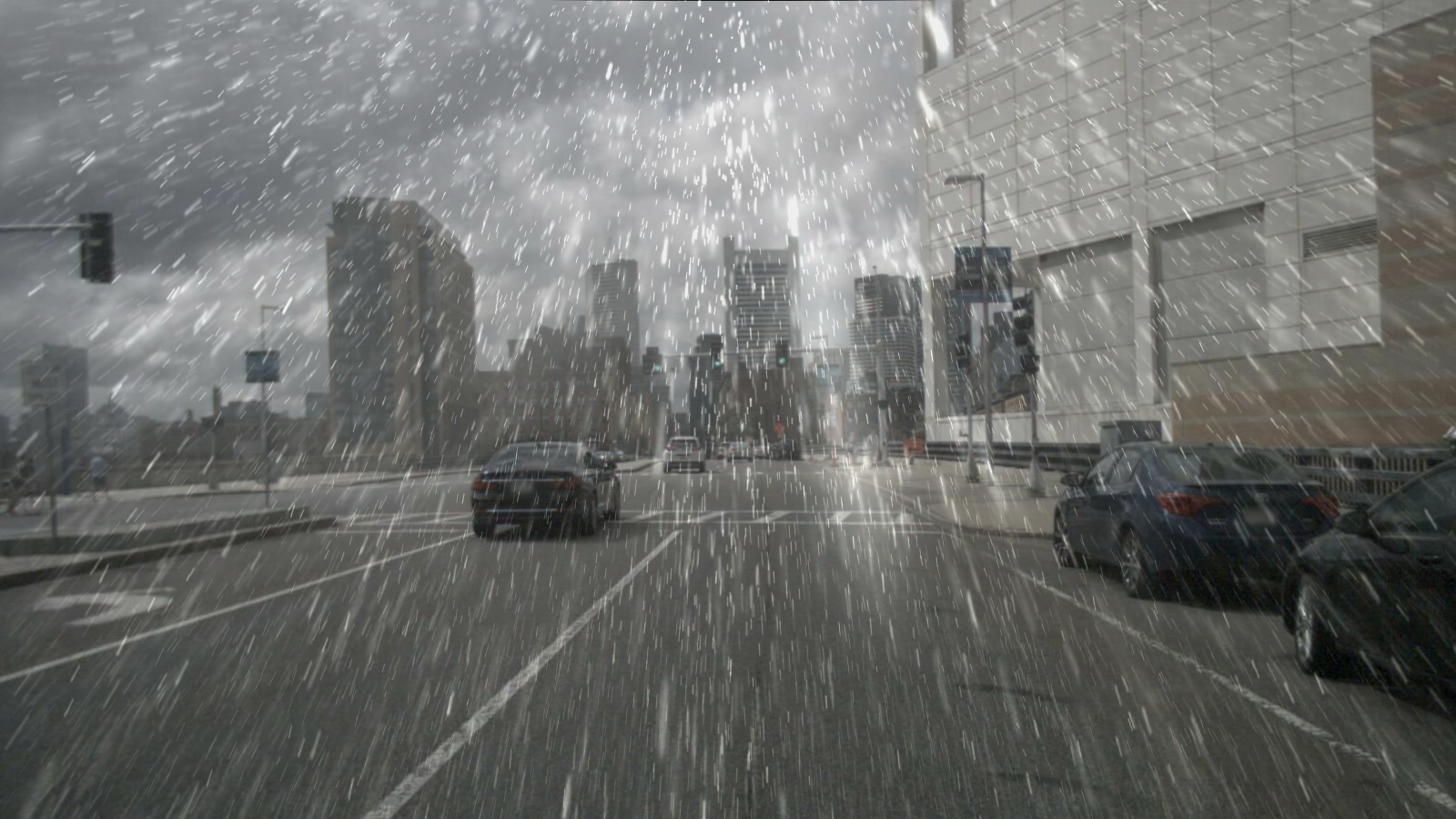} & \adjincludegraphics[width=0.45\linewidth,trim={{0.1\width} {0.1\height} {0.1\width} {0.1\height}},clip]{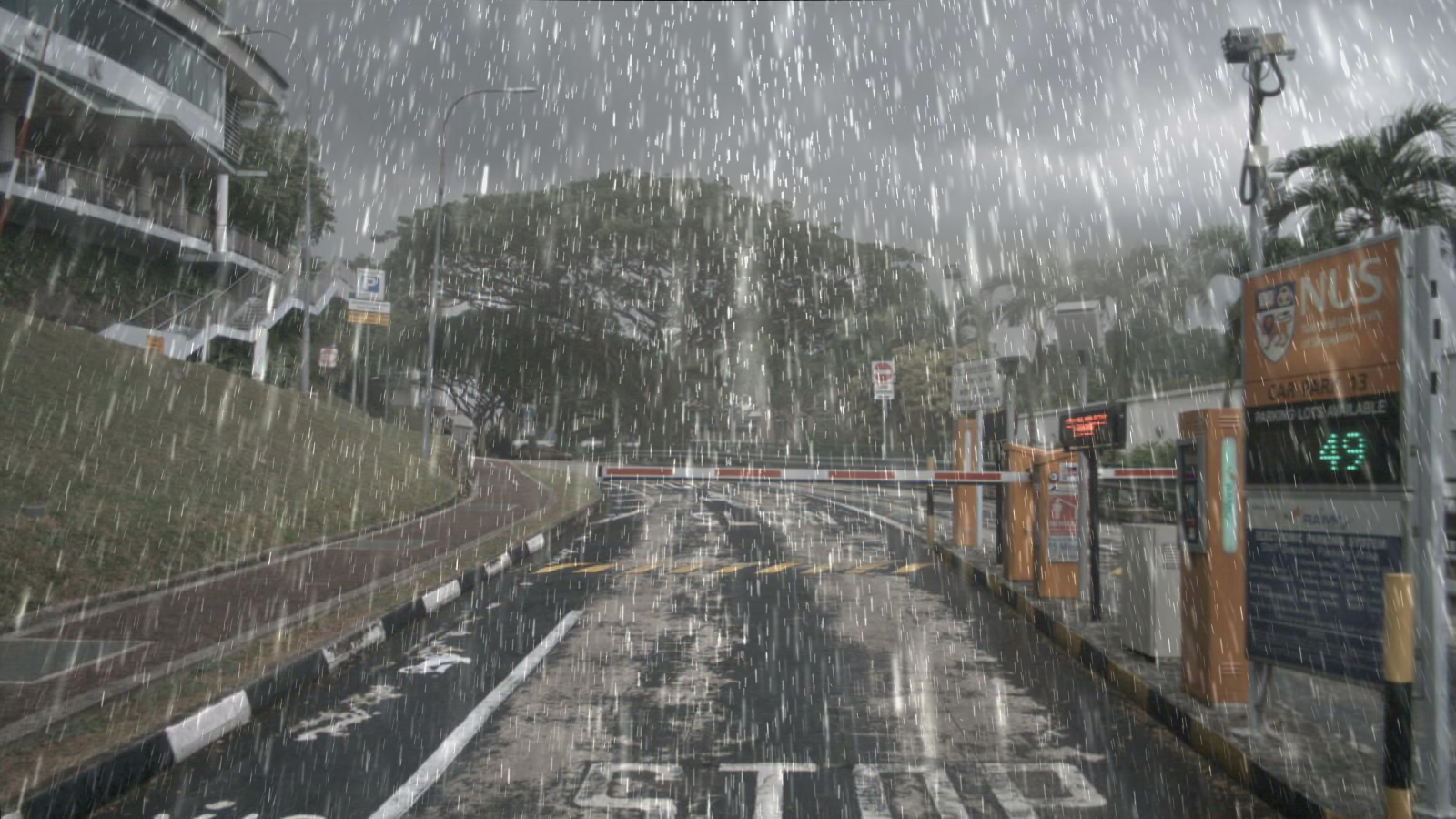} \\
  \multirow{1}{*}[1.2cm]{\rotatebox{90}{~} \rotatebox{90}{GAN}} & \adjincludegraphics[width=0.45\linewidth,trim={{0.1\width} {0.1\height} {0.1\width} {0.1\height}},clip]{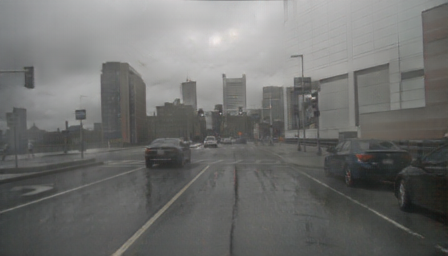} & \adjincludegraphics[width=0.45\linewidth,trim={{0.1\width} {0.1\height} {0.1\width} {0.1\height}},clip]{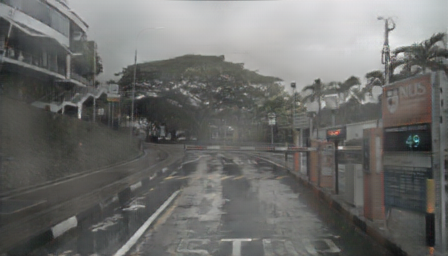} \\
  \multirow{1}{*}[1.6cm]{\rotatebox{90}{GAN+PBR} \rotatebox{90}{~100mm/hr}} & \adjincludegraphics[width=0.45\linewidth,trim={{0.1\width} {0.1\height} {0.1\width} {0.1\height}},clip]{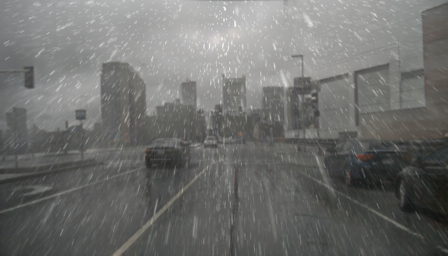} & \adjincludegraphics[width=0.45\linewidth,trim={{0.1\width} {0.1\height} {0.1\width} {0.1\height}},clip]{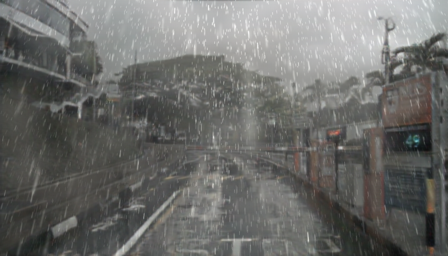} \\
  \multirow{1}{*}[1.6cm]{\rotatebox{90}{GAN+PBR} \rotatebox{90}{~200mm/hr}} & \adjincludegraphics[width=0.45\linewidth,trim={{0.1\width} {0.1\height} {0.1\width} {0.1\height}},clip]{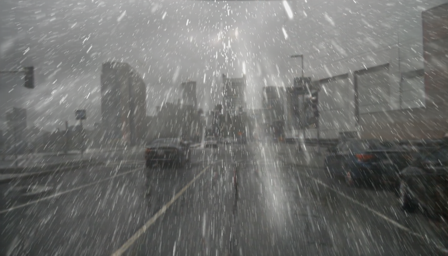} & \adjincludegraphics[width=0.45\linewidth,trim={{0.1\width} {0.1\height} {0.1\width} {0.1\height}},clip]{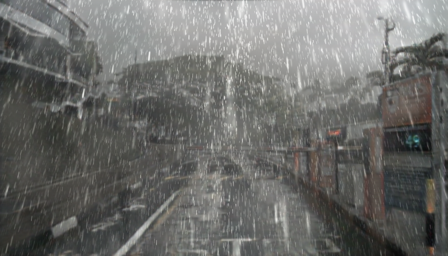} \\
  
 \end{tabular}
 \begin{tabular}{cccc}
  \toprule
  \multicolumn{4}{c}{\textbf{Other physic-based rain rendering}}\\  
  \hspace{0.75em} &\adjincludegraphics[width=0.31\linewidth,trim={0 0 0 {0.03532268746\height}},clip]{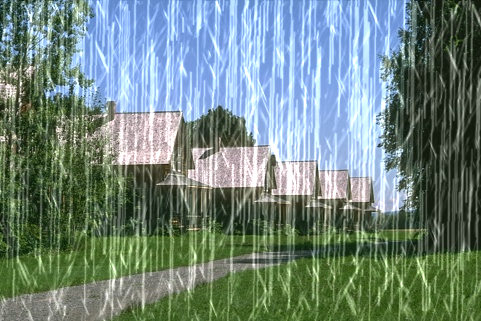}&\adjincludegraphics[width=0.31\linewidth,trim={0 0 0 {0.13\height}},clip]{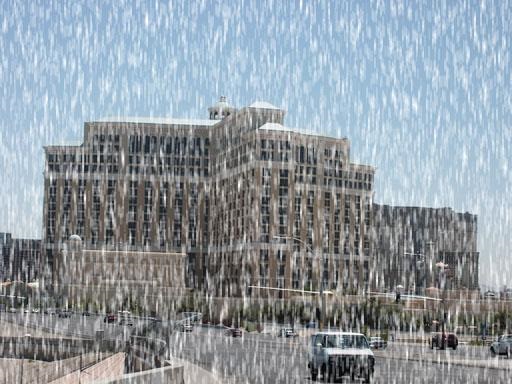}&\adjincludegraphics[width=0.31\linewidth,trim={0cm {0.08795508549\height} 0cm {0.25591017099\height}},clip]{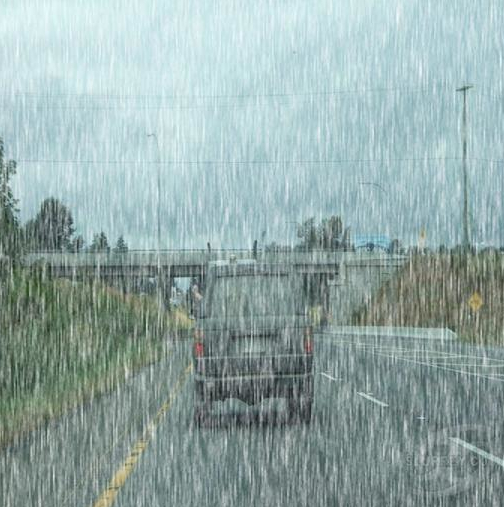}\\
  & rain100H \cite{yang2017deep} & rain800 \cite{zhang2019image} & did-MDN \cite{zhang2018density}
 \end{tabular}
 \caption{\textbf{Comparison of real photographs and our renderings.} Real photographs (source: web, \cite{luo2015removing}, \cite{caesar2019nuscenes}) showing various rain intensity, sample output of our rain rendering (PBR, GAN, and GAN+PBR), and other recent rain rendering methods. Although rain appearance is highly camera-dependent \cite{garg2005does}, results show that both real photographs and our rain generation share volume attenuation and sparse visible streaks which correctly vary with the scene background. As opposed to the other rain rendering methods, our pipeline simulates physical rainfall (here, 100mm/hr and 200mm/hr) and valid particles photometry.}
 \label{fig:comparison-rain}
\end{figure}

Fig.~\ref{fig:comparison-rain} presents real photographs captured under heavy rain, qualitative results of our rain renderings on images from nuScenes~\cite{caesar2019nuscenes} and representative results from 3 recent synthetic rain augmented approaches~\cite{zhang2018density,yang2017deep,zhang2019image}.
From the real rain photographs, it is noticeable that rainy scenes have complex visual appearance, and that the streaks visibility is greatly affected by the imaging device and background. 

Our PBR approach is able to reproduce the complex pattern of streaks with orientation consistent with camera motion and photometry consistent with background and depth. As in the real photographs, the streaks are sparse and only visible on darker backgrounds. The veiling effect caused by the rain volume (i.e. fog-like rain) is visible where the scene depth is larger (i.e. image center, sky) and nearby streaks are accurately defocused. Still, the absence of visible wetness arguably affects the rainy feeling. 

Conversely, the GAN believably renders the wetness appearance of rainy scenes. While some reflections are geometrically incorrect (e.g., a pole is reflected in the middle of the street in the left column of fig.~\ref{fig:comparison-rain} yet no pole is present), the overall appearance is visually pleasant and the global illumination matches that of real photographs. 
A noticeable artifact caused by GAN is the blurry appearance of images, whereas real rain images are only blurred in the distance.
This is explained by the inability of the GAN to disentangle the scene from the lens drops present in the ``rainy'' training images. This leads to blurring the whole image being an easy learning optimum, as highlighted in \cite{pizzati2020model}.
Another limitation we already mentioned is that GAN does not allow to control the amount of rain in the image.

This limitation is circumvented by our GAN+PBR approach which renders controllable rain streaks while preserving the global wetness appearance learned with image translation. Despite the naive GAN and PBR compositing strategy, the drops naturally blend in the scene.

\subsection{User study}
\label{sec:val_userstudy}

\begin{figure}
 \centering
 \includegraphics[width=\linewidth]{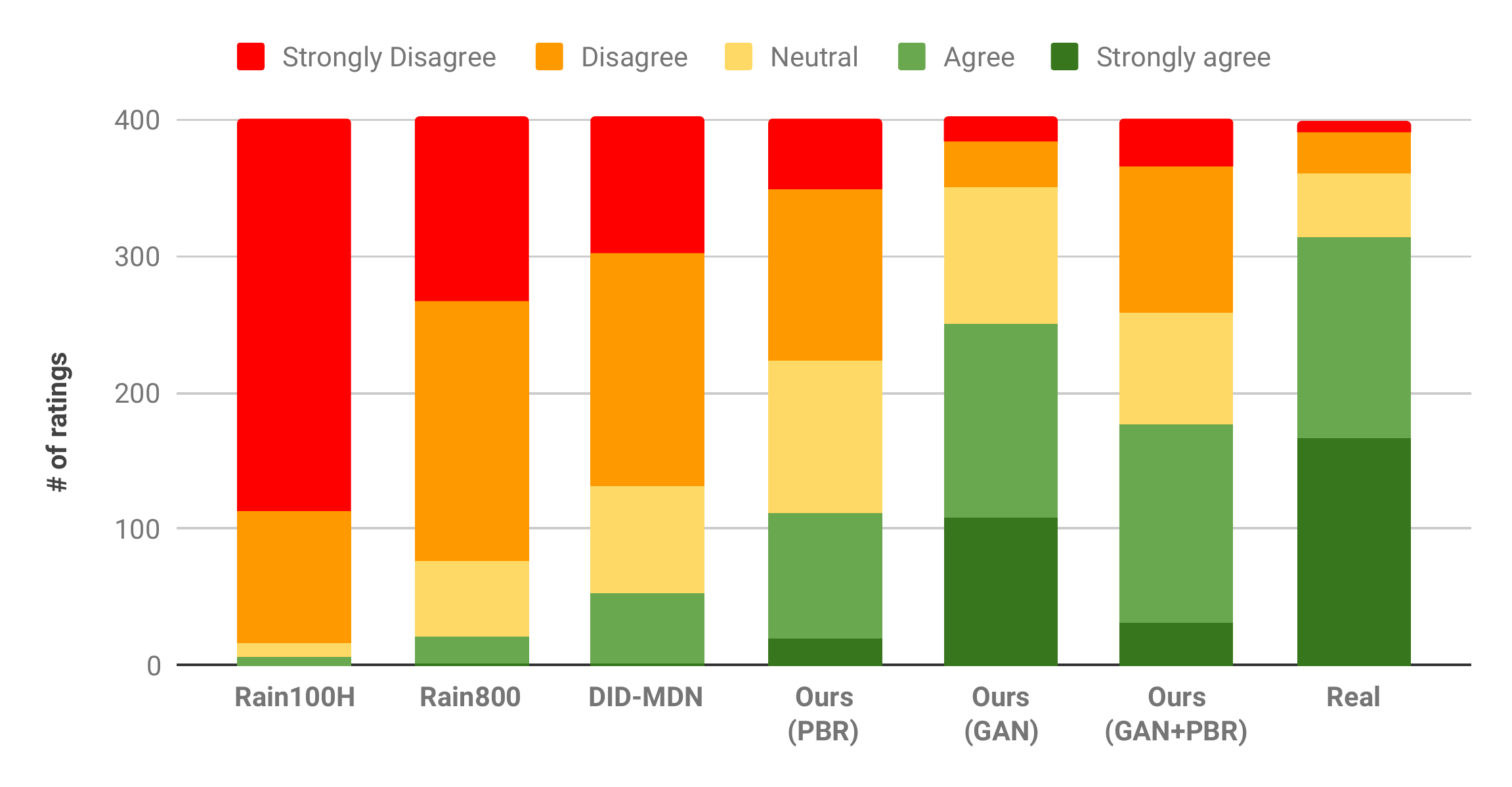}
 \caption{\textbf{User study of rainy images realism.} The y-axis displays ratings to the statement "Rain in this image looks realistic". All of our approaches significantly outperform existing techniques.}
 \label{fig:user_study}
 \includegraphics[width=\linewidth]{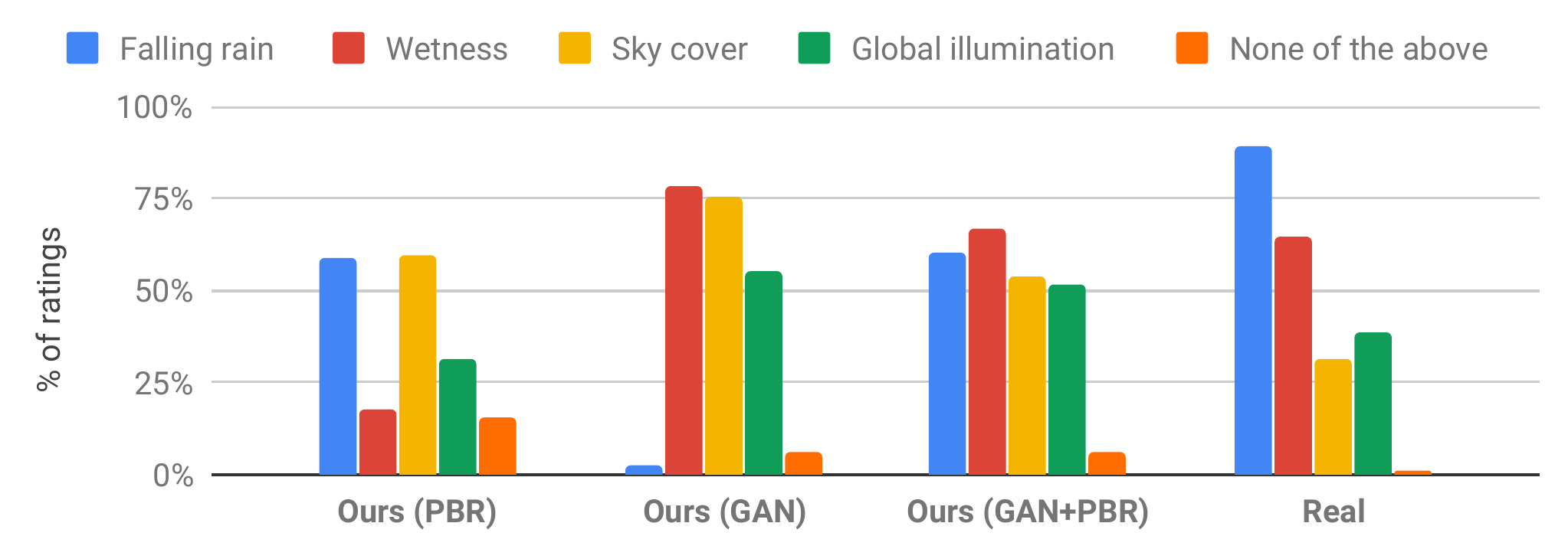}
 \caption{\textbf{User study on images characteristics conveying rain.} The y-axis displays ratings to the statement "Which of these qualities help in determining the realism of the rain".}
 \label{fig:user_study_followup}
\end{figure}

To evaluate the perceptual quality of our rain renderings, we conducted two user studies. In the first, users were shown one image at a time, and asked to rate if rain looks realistic on a 5-point Likert scale. A total of 42 images were shown, that is 6 for each of the following: real rainy photographs, ours (PBR, GAN, GAN+PBR), and previous approaches \cite{yang2017deep,zhang2018density,zhang2019image}. Answers were obtained for a total of 67 participants, aged from 22 to 75 (avg 37.0, std 14.2), with 32.8\% females.

From the Mean Opinion Score (MOS) in fig.~\ref{fig:user_study}, all our rain augmentation approaches are judged to be more realistic than any of the previous approaches. 
Specifically, when converting ratings to the [0,~1] interval, the mean rain realism is 0.77 for real photos, 0.44 for PBR, 0.68 for GAN, 0.52 for GAN+PBR, and 0.30/0.23/0.08 for~\cite{zhang2018density}/\cite{zhang2019image}/\cite{yang2017deep} respectively.
Despite physical and geometrical inconsistencies, the users consistently judged GAN images to be more realistic. This is in favor of using image-to-image translation rather than physics-based rendering for realism purposes.
However, for benchmarking or physical accuracy purposes, GAN+PBR allows us to have an arbitrary control on the rain amount at the cost of slightly lower realism.

In the second study, we asked respondents who participated in the first study to determine, for each of the same images as before (excluding images from the previous work), which visual characteristics influenced their decisions. 52 of the original participants responded, aged from 22 to 72 (avg 36.6, std 13.9) including 28.9\% females. 
Results are reported in fig.~\ref{fig:user_study_followup}.
We note that while \textit{falling rain} and \textit{wetness} are the main characteristics in real rain, GAN---judged the most realistic approach---fails to convey the \textit{falling rain} appearance but excels at rendering \textit{wetness}. The opposite is observed with PBR, though few users indicated \textit{wetness} (despite its absence). The GAN+PBR offers a trade-off balancing all characteristics.

\section{Evaluating the impact of real rain}
\label{sec:real_data}

\begin{figure*}[!t] 
 \centering
 \scriptsize
 \setlength{\tabcolsep}{0.001\linewidth}
 \renewcommand{\arraystretch}{1.25}
 \begin{tabular}{cccccc}
  \multirow{1}{*}[1.15cm]{\rotatebox{90}{Input}} & 
 \includegraphics[width=0.222\linewidth]{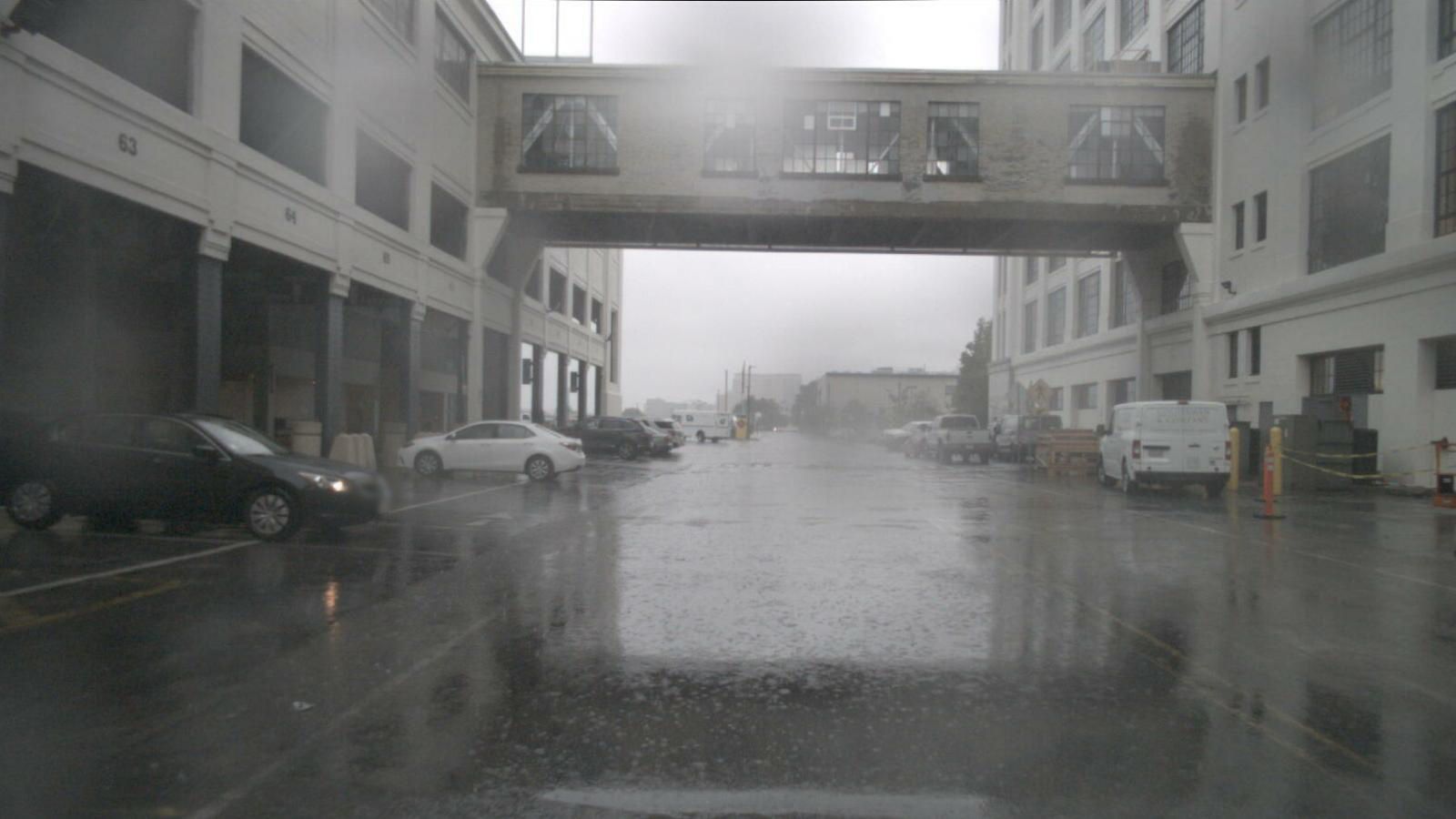} &
 \includegraphics[width=0.222\linewidth]{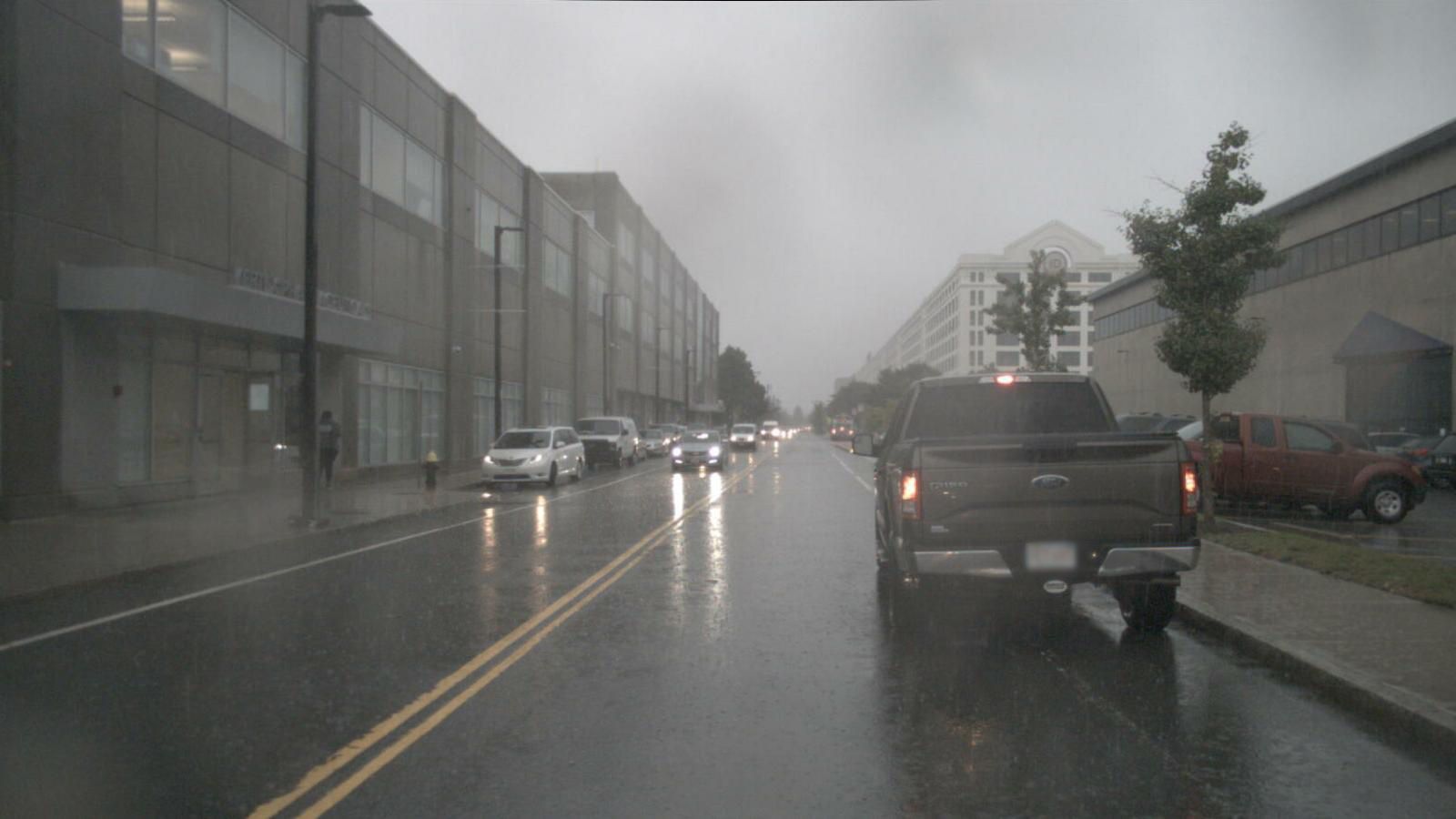} &
 \includegraphics[width=0.222\linewidth]{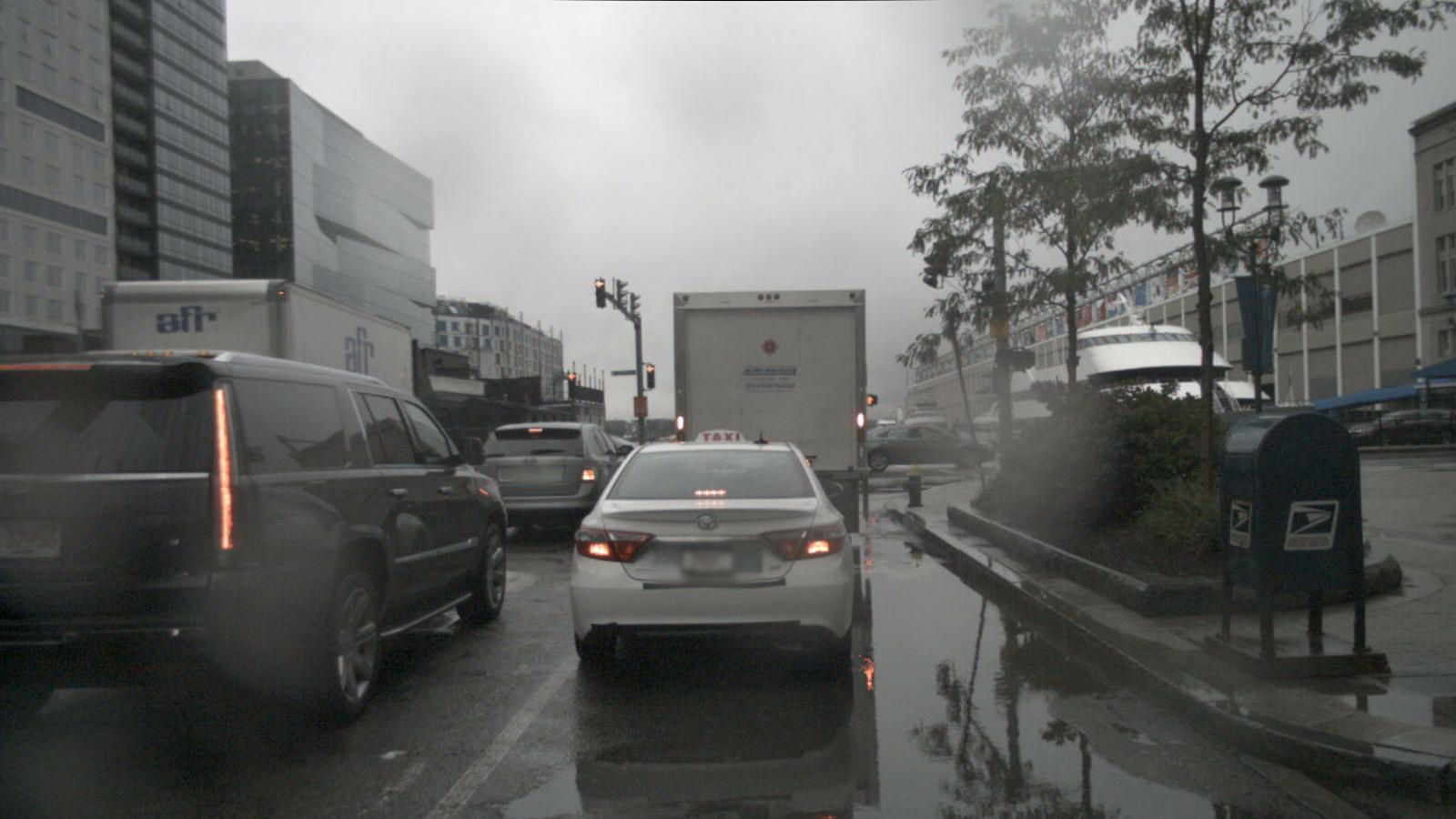} & 
 \includegraphics[width=0.222\linewidth]{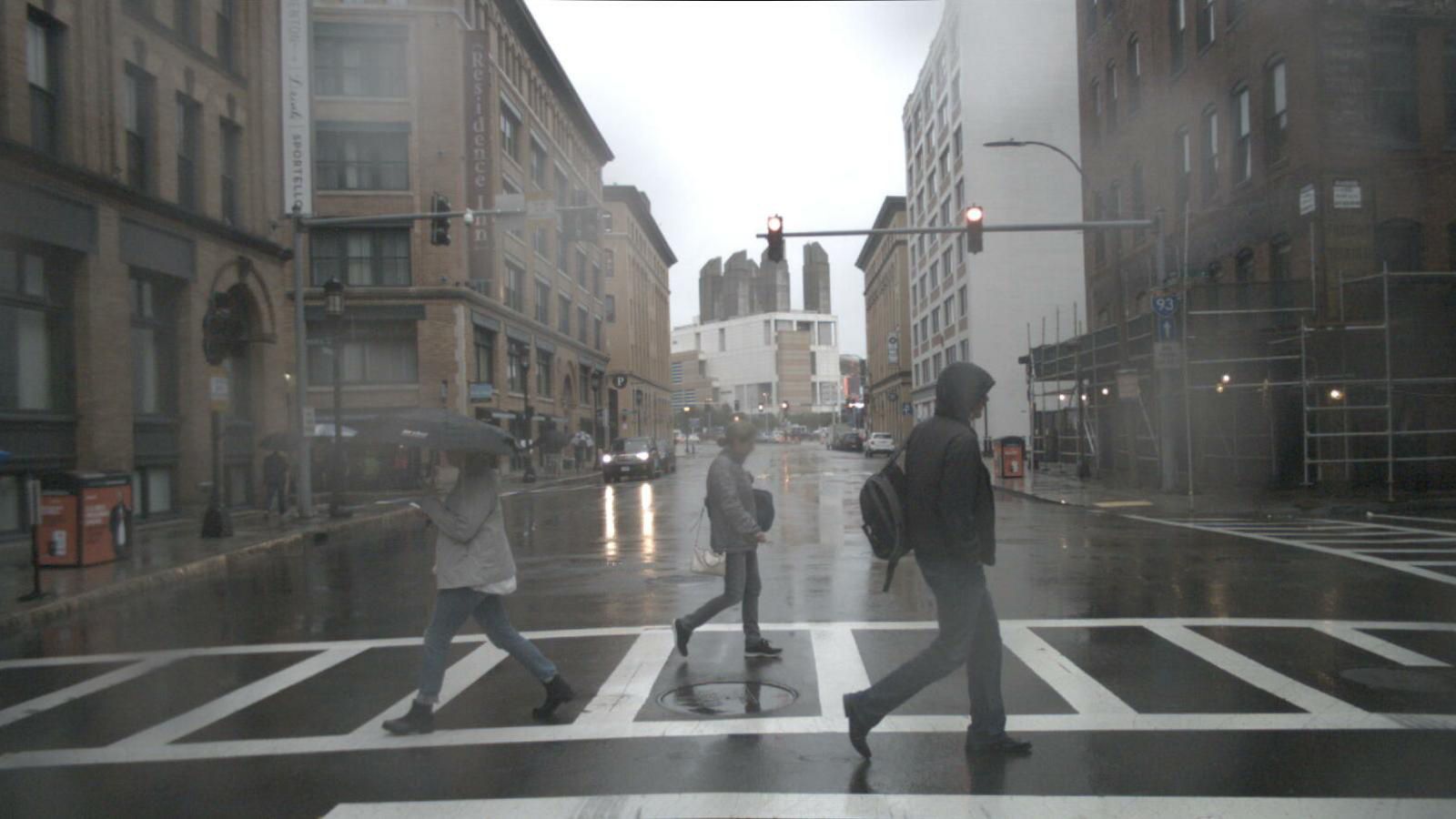} \\
 
  \multirow{1}{*}[1.5cm]{\rotatebox{90}{YOLOv2} \rotatebox{90}{~~~\cite{redmon2017yolo9000}} } & 
 \includegraphics[width=0.222\linewidth]{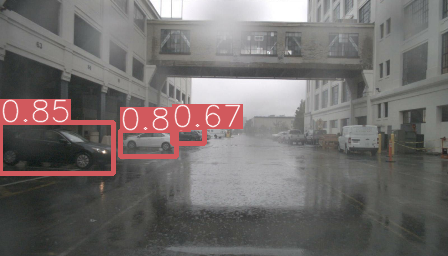} &
 \includegraphics[width=0.222\linewidth]{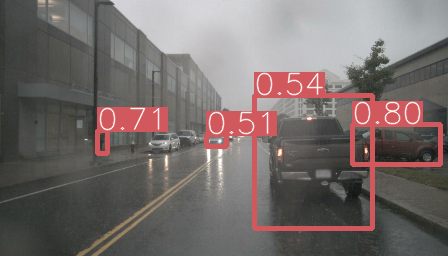} &
 \includegraphics[width=0.222\linewidth]{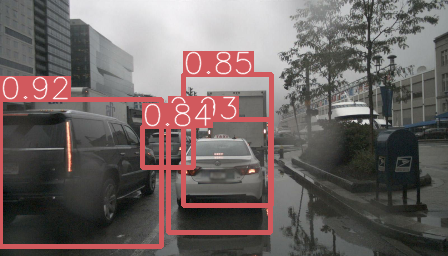} & 
 \includegraphics[width=0.222\linewidth]{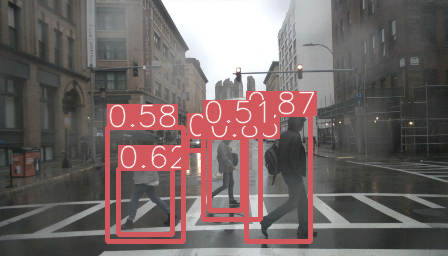} \\
 
  \multirow{1}{*}[1.35cm]{\rotatebox{90}{PSPNet} \rotatebox{90}{~~~\cite{zhao2017pspnet}} } &
 \includegraphics[width=0.222\linewidth]{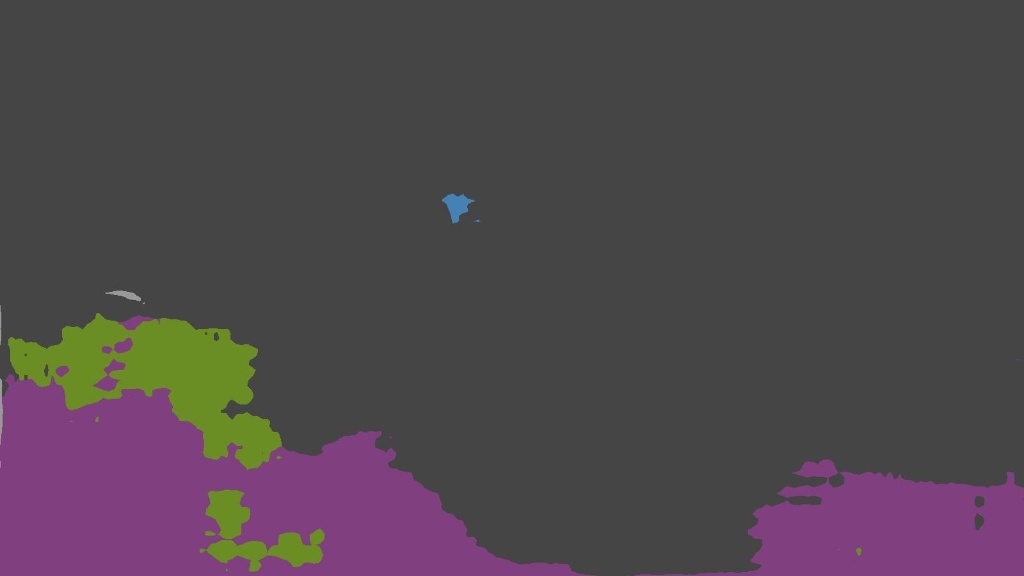} &
 \includegraphics[width=0.222\linewidth]{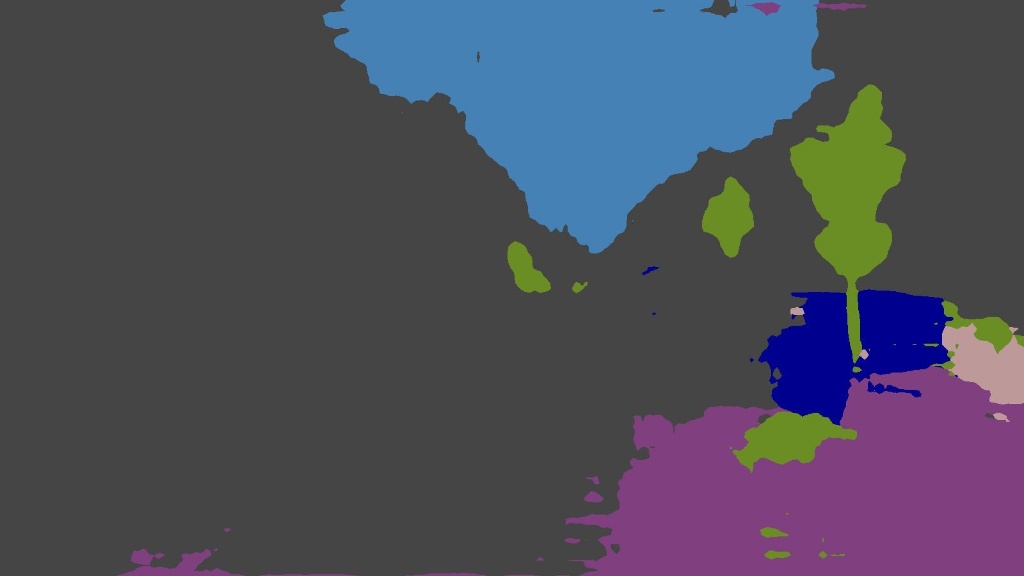} &
 \includegraphics[width=0.222\linewidth]{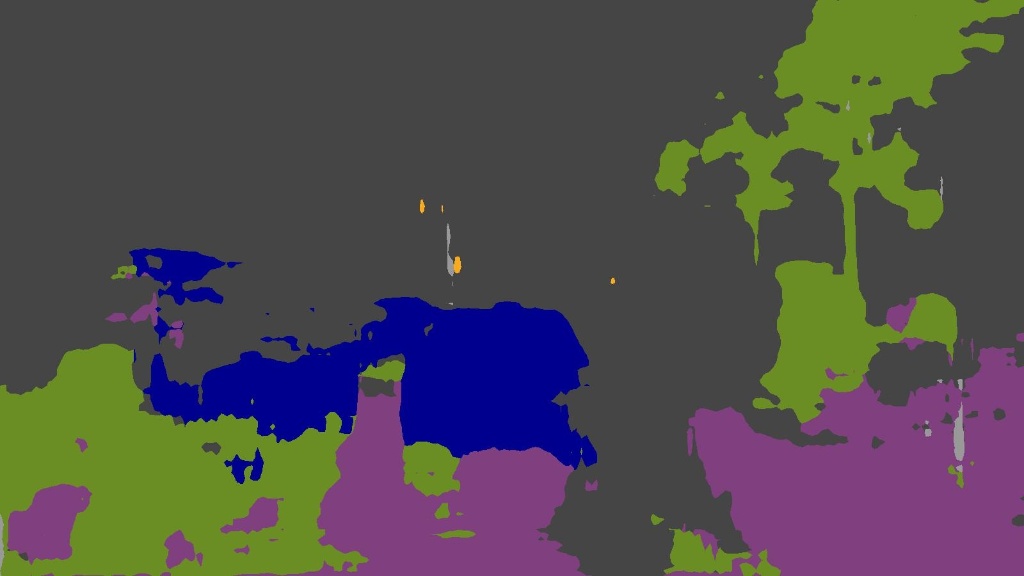} & 
 \includegraphics[width=0.222\linewidth]{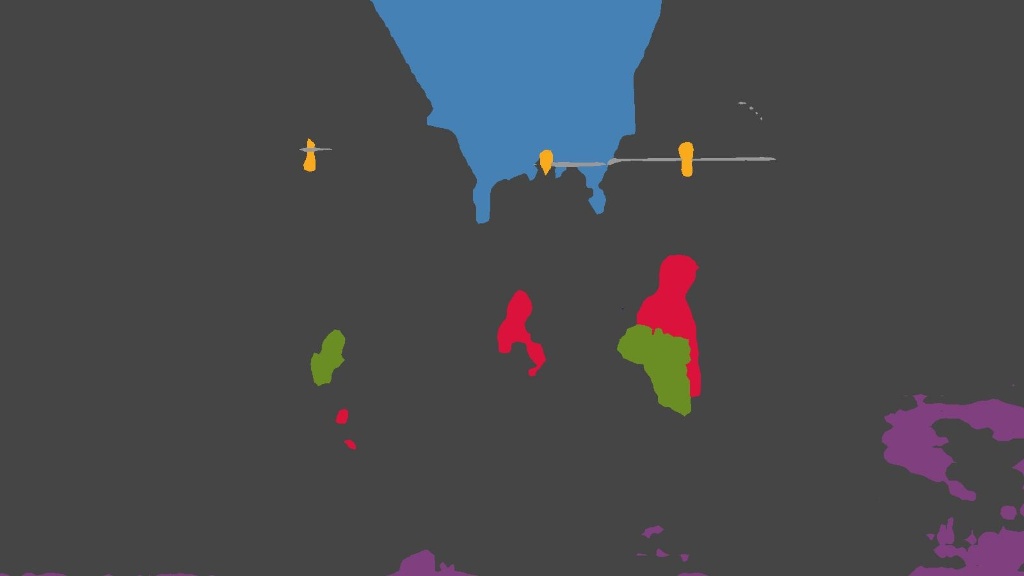} \\

  \multirow{1}{*}[1.625cm]{\rotatebox{90}{Monodepth2} \rotatebox{90}{~~~~~~\cite{godard2019digging}} } & 
 \includegraphics[width=0.222\linewidth]{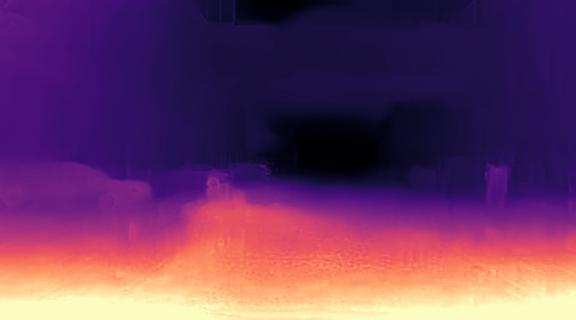} &
 \includegraphics[width=0.222\linewidth]{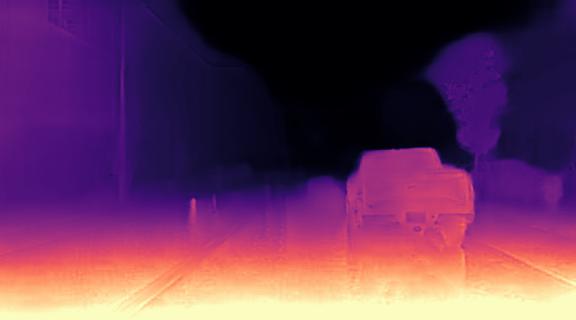} &
 \includegraphics[width=0.222\linewidth]{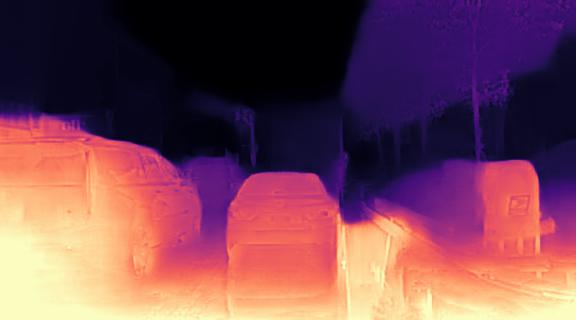} & 
 \includegraphics[width=0.222\linewidth]{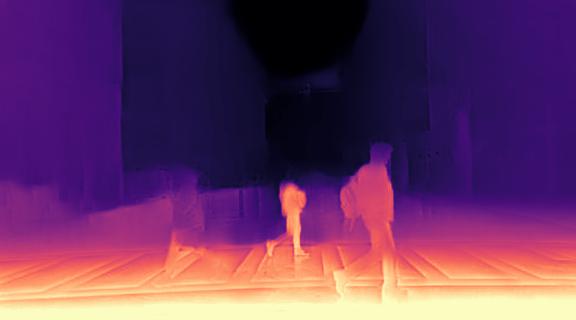} \\
 \end{tabular}
 \caption{\textbf{Qualitative results on real rainy images from the nuScenes dataset.} Shown for different tasks: object detection (top line), semantic segmentation (middle line), and depth estimation (bottom line) } %
 \label{fig:eval_qualitative_real_rain}
\end{figure*}

We now aim at quantifying the impact of rain on three computer vision tasks: object detection, semantic segmentation, and depth estimation. These tasks are critical for any outdoor vision systems such as mobile robotics, autonomous driving, or surveillance. Before we experiment on rain-augmented images with our PBR, GAN, and GAN+PBR approaches, we first experiment on real rainy photographs. 

For this, we use the nuScenes dataset~\cite{caesar2019nuscenes}. Benefiting from coarse frame-wise weather annotations in the dataset, we split nuScenes images\footnote{Note that only front camera and annotated key frames are used for consistency and ground truth accuracy.} in two subsets: ``nuScenes-clear'' (images without rain) and ``nuScenes-rain'' (rainy images). Due to the noisy weather labels, we cross-validated each frame with a historic weather database\footnote{Weather database at \url{https://openweathermap.org}.} using GPS location and time, and only kept frames where the nuScenes label agreed with the weather database. This resulted in sets of 24134 images for nuScenes-clear, and 6028 for nuScenes-rain. In this dataset, rain images are dark, gloomy, the sky is heavily covered, and no falling rain is visible but there are unfocused raindrops occlusions on the lens.

We experiment on these sets of real images with algorithms from each task: YOLOv2~\cite{redmon2017yolo9000} for object detection, PSPNet~\cite{zhao2017pspnet} for semantic segmentation, and Monodepth2~\cite{godard2019digging} for depth estimation. The nuScenes-clear set is split into train/test subsets of 19685/4449 images from 491/110 scenes. The nuScenes-rain is also split into train/test subsets of 5419/609 images from 134/15 scenes. Here, 1000 images from the nuScenes-clear(test) and all images from the nuScenes-rain(test) subset are used for evaluation (train will be used for the GAN in sec.~\ref{sec:rain_setup}). These training and testing sets and subsets are all displayed in table \ref{tab:splits} in appendix~\ref{sec:app-data-splits}. Note that a large number of images were needed to train the CycleGAN for image translation and, to avoid any overlap in image sequences, it unfortunately left a relatively small subset of nuScenes for the evaluation on real rainy images.

For object detection and depth estimation, we first pre-train each algorithm on ImageNet (Darknet53, $448\times448$) and KITTI (monocular, $1024\times320$) respectively, and further finetune them on the nuScenes-clear(train) subset to limit the domain gap between datasets. We then evaluate their performance on the aforementioned test subsets. For segmentation, since semantic labels are not provided on nuScenes, we carefully annotated 25 images from both nuScenes-clear(test) and nuScenes-rain(test). Since we do not have enough labeled data for finetuning, we use our model pretrained on Cityscapes~\cite{Cordts2016Cityscapes}, with the caveat that there may be a significative domain gap between training and evaluation. 

Table~\ref{fig:real_rain_perf} reports the results of this experiment. The performance on real rainy images compared to clear images for all three tasks are a mAP of 16.30\% instead of 32.53\% for object detection, an AP of 18.7\% instead of 40.8\% for semantic segmentation and a square relative error of 3.53\% instead of 2.96\% for depth estimation. Corresponding qualitative results are displayed in fig.~\ref{fig:eval_qualitative_real_rain}. 
As expected, real rain deteriorates the performance of all algorithms on all tasks. However, we cannot evaluate how rain \emph{intensity} affects these algorithms since it would require the accurate measurement of the rainfall rate at the time of capture.

\begin{table}
 \centering
  \scriptsize
  \setlength{\tabcolsep}{0.18cm}
  \renewcommand{\arraystretch}{1.0}
  \begin{tabular}{llcc}
   \toprule
   Tasks & Metric & Clear & Rain \\
   \midrule
   Object detection~\cite{redmon2017yolo9000} & mAP (\%) $\uparrow$ & 24.70 & 11.09 \\
   Semantic segmentation~\cite{zhao2017pspnet} & AP (\%) $\uparrow$ & 40.8 & 18.7 \\
   Depth estimation~\cite{godard2019digging} & Sq. err. (\%) $\downarrow$ & 2.96 & 3.53 \\ 
   \bottomrule
 \end{tabular}\hspace{.025\linewidth}
 \caption{\textbf{Vision tasks on real clear/rain images from nuScenes~\cite{caesar2019nuscenes}.} Significant performance drops are observed in rainy weather.}
 \label{fig:real_rain_perf}
\end{table}

\section{Evaluating the impact of synthetic rain}
\label{sec:impact_synth}

To study how vision algorithms perform under increasing amounts of rain, we leverage our rain synthesis pipeline and augment popular clear-weather datasets. Specifically, our PBR and GAN+PBR frameworks allow us to measure the performance of these algorithms in \emph{controlled} rain settings.

\subsection{Rain generation setup}
\label{sec:rain_setup}

We augment all three of the KITTI, Cityscapes, and nuScenes-clear datasets with PBR. 
We generate rainfall rates ranging from light to heavy storm $R = \{0, 5, 25, 50, 100, 200\}$~mm/hr. 
Only the nuScenes-clear dataset is augmented with GAN and GAN+PBR, since neither KITTI nor Cityscapes contain rainy images to train the GAN. 
Our PBR and GAN+PBR rain augmentation require some preparation as they rely on calibration, depth and camera motion. Our GAN and GAN+PBR rain augmentation require the training of a CycleGAN. These preparations are described below.

\paragraph{Calibration.} For the realistic physical simulator (sec.~\ref{sec:raindrop-position}) and the rain streaks photometric simulation (sec.~\ref{sec:photometry-rainstreak}), intrinsic and extrinsic calibration are used to replicate the imaging sensor. 
We used frame-wise or sequence-wise calibration for KITTI and nuScenes. 
In addition, we use 6mm focal and 2ms exposure for KITTI~\cite{Geiger2012CVPR,Geiger2013IJRR} and assumed 5ms exposure for nuScenes.
As Cityscapes does not provide calibration, we use intrinsic from the camera manufacturer with 5ms exposure and extrinsic is assumed similar to KITTI. 

\paragraph{Depth.} The scene geometry (pixel depth) is also required to model accurately the light-particle interaction and the fog optical extinction. 
We estimate KITTI depth maps from RGB+Lidar with~\cite{jaritz2018sparse}, and Cityscapes/nuScenes from monocular RGB with Monodepth~\cite{monodepth17,godard2019digging}.
While absolute depth is not required, we aim to avoid the critical artifacts along edges, and thus further align RGB with depth using guided filter~\cite{barron2016fast}. 

\paragraph{Camera motion.} We mimic the camera ego motion in the physical simulator to ensure realistic rain streak orientation on still images and preserve temporal consistency in sequences.
Ego speed is extracted from GPS data when provided (KITTI and nuScenes), or drawn uniformly in the $[0, 50]$~km/hr interval for Cityscapes semantics and in the $[0, 100]$~km/hr interval for KITTI object to reflect the urban and semi-urban scenarios, respectively.

\paragraph{CycleGAN.} A CycleGAN is trained for image-to-image rain translation on the train subsets of nuScenes-clear and nuScenes-rain (sec.~\ref{sec:real_data}). In order to make sure that no image is used to both train and evaluate the GAN simultaneously, we use the 4449 images from the nuScenes-clear(test) subset, resize them to $448\times256$, and perform image-to-image translation to generate GAN-augmented rain images. We dub this new set of images ``nuScenes-augment'' for clarity, and will also use this for the GAN+PBR rain augmentation. 

\begin{figure}
 \centering
 \subfloat[Object detection]{\includegraphics[width=0.485\linewidth]{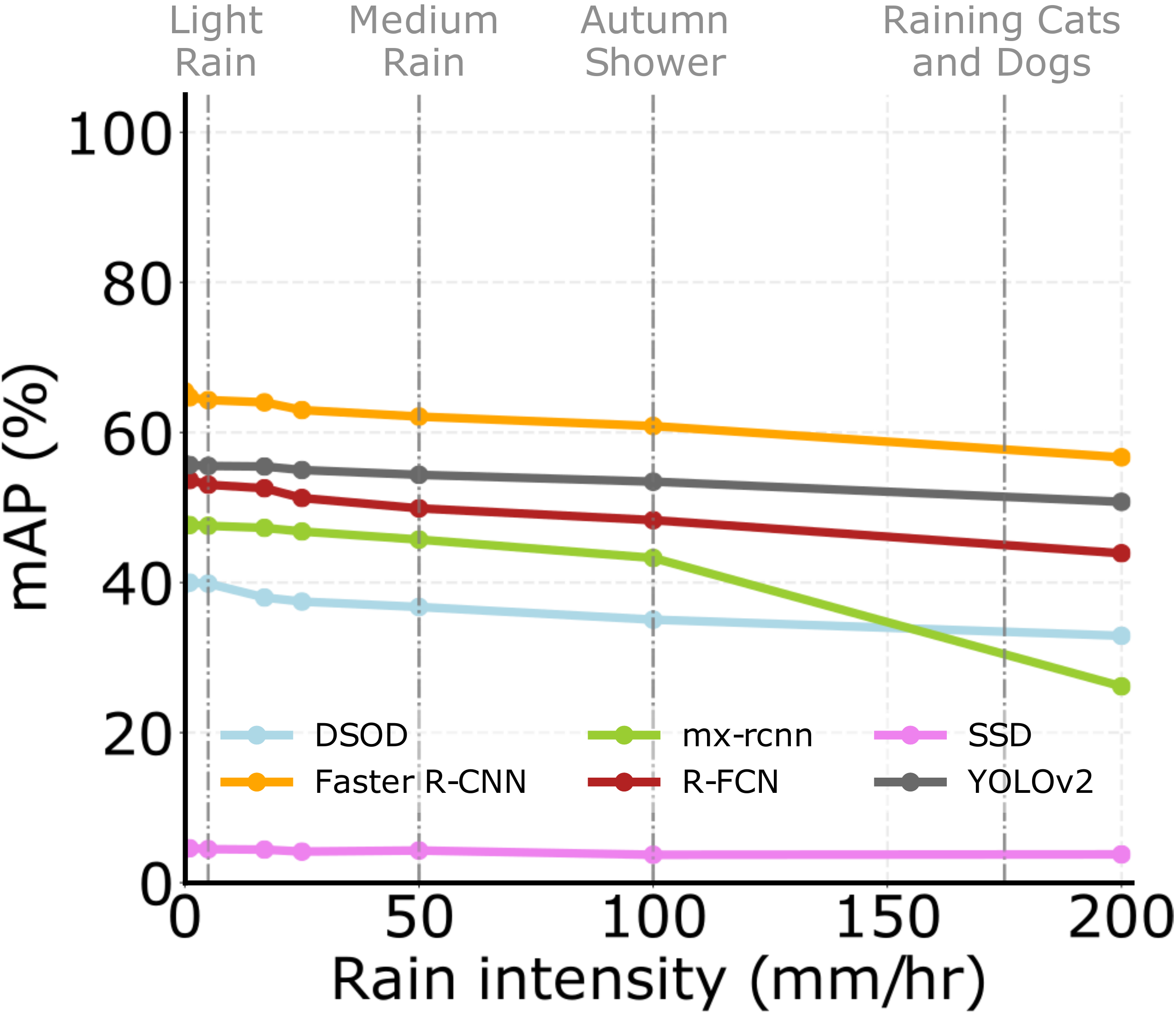}\label{fig:algo_obj_detect_perf}}
 \subfloat[Semantic segmentation]{\includegraphics[width=0.485\linewidth]{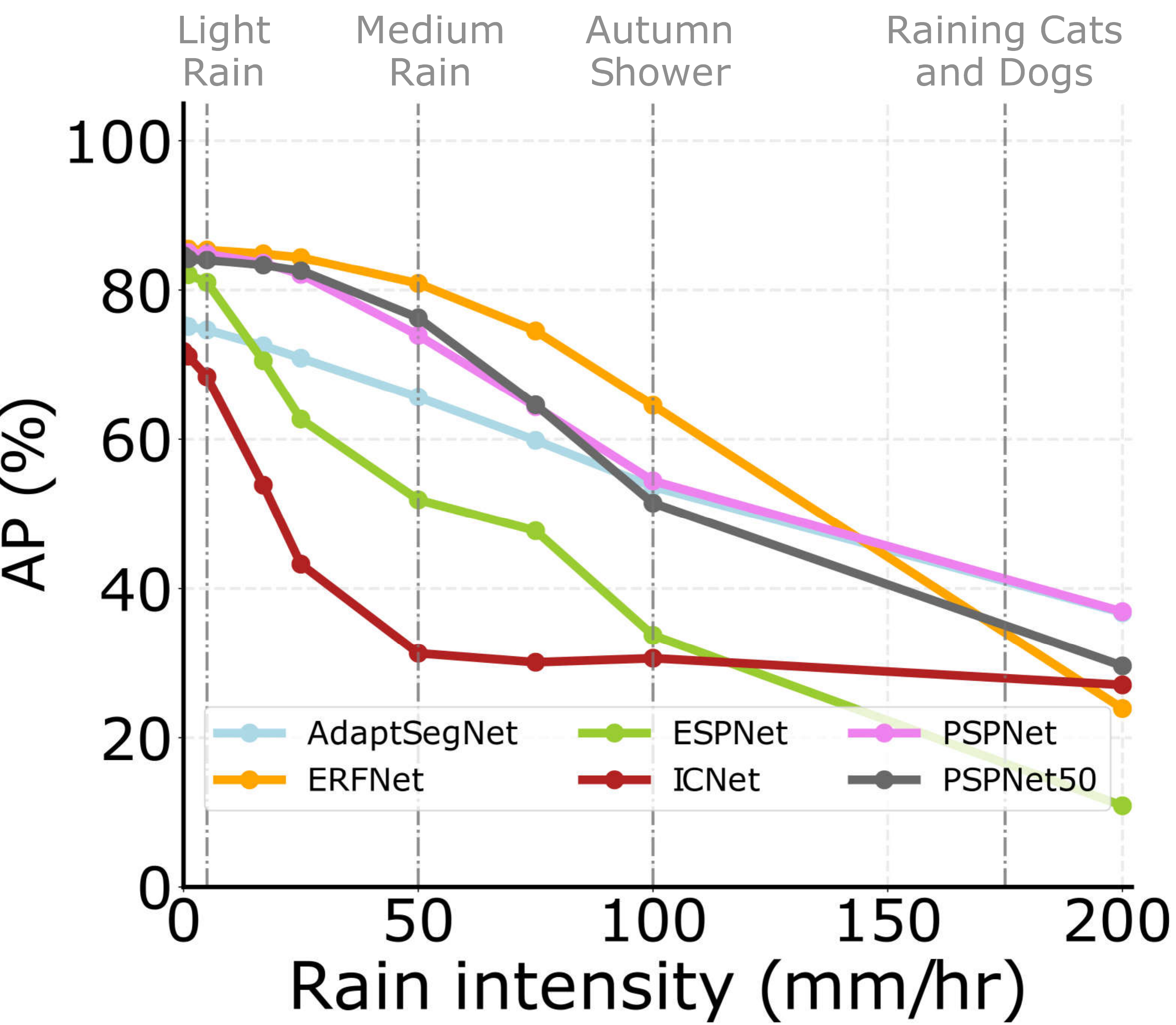}\label{fig:algo_sem_seg_perf}}
 \hfill%
 \subfloat[Depth estimation]{\includegraphics[width=0.485\linewidth]{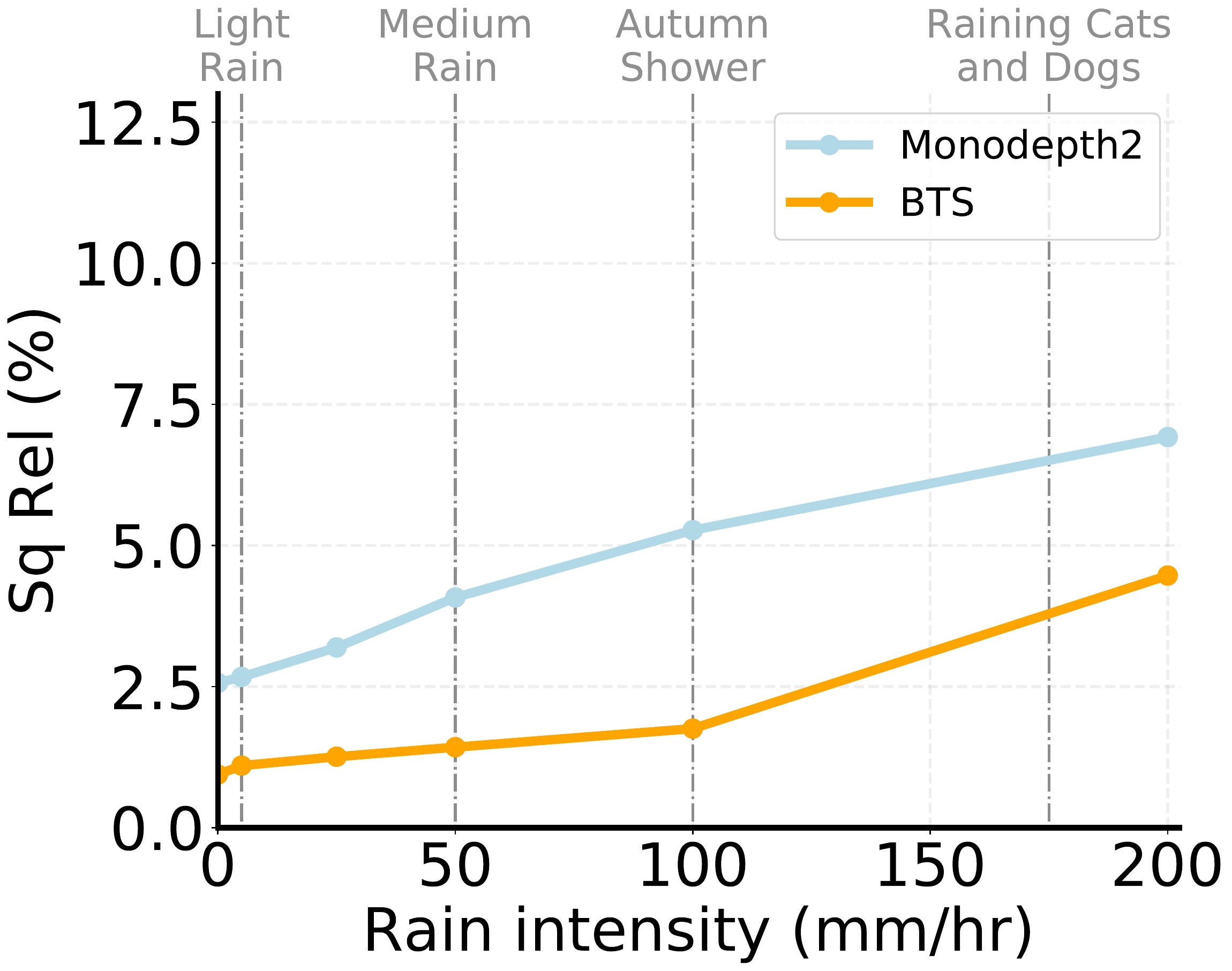}\label{fig:algo_depth_estim_perf}}
 \caption{\textbf{Performance using our PBR rain augmentation}. \protect\subref{fig:algo_obj_detect_perf} Object detection performance on our weather augmented KITTI dataset, \protect\subref{fig:algo_sem_seg_perf} pixel-semantic segmentation performance on our weather augmented Cityscapes dataset, and \protect\subref{fig:algo_depth_estim_perf} depth estimation performance on our weather augmented nuScenes dataset, all of them as a function of rainfall rate. The object detection plot shows the Coco mAP@[.1:.1:.9] (\%) across cars and pedestrians, the semantic segmentation plot shows the AP (\%), and the depth estimation shows the squared relative error (\%). As opposed to object detection which exhibits some robustness, the segmentation and depth estimation tasks are strongly affected by the rain.}
 \label{fig:obj_sem_depth_rain_perf}
\end{figure}

\subsection{Evaluating PBR rain augmentation}

\newcommand\weatherAugmented[4]{
 \adjincludegraphics[width=0.222\linewidth, trim={{0.28346456692\width} 3px {0.12047244094\width} 2px}, clip]{figures/results/qualitative_rain/#2/clear/#1.#3} \hspace{0.085em} &
 \adjincludegraphics[width=0.222\linewidth, trim={{0.28346456692\width} 0 {0.12047244094\width} 0}, clip]{figures/results/qualitative_rain/#2/pbr/50mm/#1.#4} &  
 \adjincludegraphics[width=0.222\linewidth, trim={{0.28346456692\width} 0 {0.12047244094\width} 0}, clip]{figures/results/qualitative_rain/#2/pbr/100mm/#1.#4} & 
 \adjincludegraphics[width=0.222\linewidth, trim={{0.28346456692\width} 0 {0.12047244094\width} 0}, clip]{figures/results/qualitative_rain/#2/pbr/200mm/#1.#4}}

\newcommand\weatherHybridAugmented[4]{
 \adjincludegraphics[width=0.191\linewidth]{figures/results/qualitative_rain/#2/clear_small/#1.#3} \hspace{0.085em} &
 \adjincludegraphics[width=0.191\linewidth]{figures/results/qualitative_rain/#2/gan/#1.#4} &
 \adjincludegraphics[width=0.191\linewidth]{figures/results/qualitative_rain/#2/hybrid/50mm/#1.#4} & 
 \adjincludegraphics[width=0.191\linewidth]{figures/results/qualitative_rain/#2/hybrid/100mm/#1.#4} & 
 \adjincludegraphics[width=0.191\linewidth]{figures/results/qualitative_rain/#2/hybrid/200mm/#1.#4}}

\begin{figure*}
 \newcommand\weatheraugvizObjDetT[3]{
  \adjincludegraphics[width=0.222\linewidth, trim={{0.28346456692\width} 0 {0.12047244094\width} 0}, clip]{figures/results/obj_detection/#2/#1#3_clear_tb.jpg} \hspace{0.085em} & 
  \adjincludegraphics[width=0.222\linewidth, trim={{0.28346456692\width} 0 {0.12047244094\width} 0}, clip]{figures/results/obj_detection/#2/#1#3_50mm_tb.jpg} & 
  \adjincludegraphics[width=0.222\linewidth, trim={{0.28346456692\width} 0 {0.12047244094\width} 0}, clip]{figures/results/obj_detection/#2/#1#3_100mm_tb.jpg} &
  \adjincludegraphics[width=0.222\linewidth, trim={{0.28346456692\width} 0 {0.12047244094\width} 0}, clip]{figures/results/obj_detection/#2/#1#3_200mm_tb.jpg}}
 \newcommand\weatheraugvizObjDet[2]{\weatheraugvizObjDetT{#1}{#2}{_#2}}
 
 \centering
 \scriptsize
 \setlength{\tabcolsep}{0.001\linewidth}
 \renewcommand{\arraystretch}{0.5}
 \begin{tabular}{ccccc}
  & \textbf{\footnotesize{Original}} & \multicolumn{3}{c}{\textbf{\footnotesize{Rain augmented (PBR)}}} \\
  \cmidrule[1pt](l{2pt}r{4pt}){2-2}\cmidrule[1pt](l{2pt}r{2pt}){3-5}
  \multirow{1}{*}[1.15cm]{\rotatebox{90}{Input}} & 
	 \adjincludegraphics[width=0.222\linewidth, trim={{0.28346456692\width} 3px {0.12047244094\width} 2px}, clip]{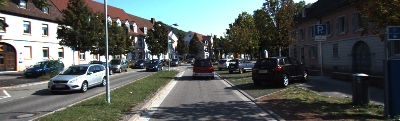} \hspace{0.085em} &
   \adjincludegraphics[width=0.222\linewidth, trim={{0.28346456692\width} 0 {0.12047244094\width} 0}, clip]{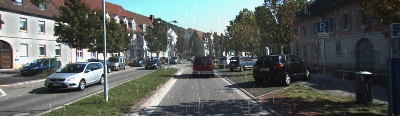} &  
   \adjincludegraphics[width=0.222\linewidth, trim={{0.28346456692\width} 0 {0.12047244094\width} 0}, clip]{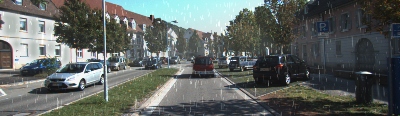} & 
   \adjincludegraphics[width=0.222\linewidth, trim={{0.28346456692\width} 0 {0.12047244094\width} 0}, clip]{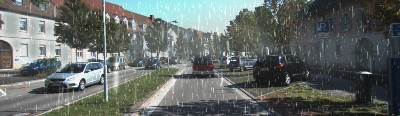} \\

  \multirow{1}{*}[1.45cm]{\rotatebox{90}{FasterRCNN} \rotatebox{90}{~~~~~~\cite{ren2015faster}}} & 
	  \adjincludegraphics[width=0.222\linewidth, trim={{0.28346456692\width} 0 {0.12047244094\width} 0}, clip]{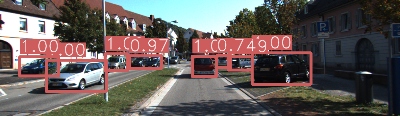} \hspace{0.085em} & 
    \adjincludegraphics[width=0.222\linewidth, trim={{0.28346456692\width} 0 {0.12047244094\width} 0}, clip]{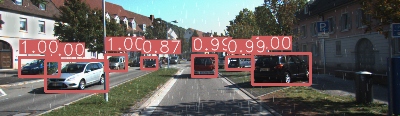} & 
    \adjincludegraphics[width=0.222\linewidth, trim={{0.28346456692\width} 0 {0.12047244094\width} 0}, clip]{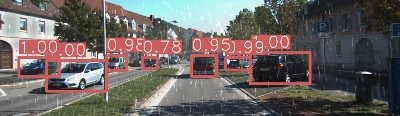} &
    \adjincludegraphics[width=0.222\linewidth, trim={{0.28346456692\width} 0 {0.12047244094\width} 0}, clip]{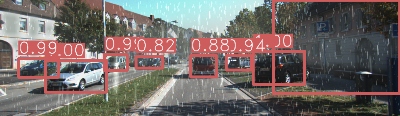} \\
	
  \multirow{1}{*}[1.15cm]{\rotatebox{90}{R-FCN} \rotatebox{90}{~~\cite{dai2016r}}} & 
	\adjincludegraphics[width=0.222\linewidth, trim={{0.28346456692\width} 0 {0.12047244094\width} 0}, clip]{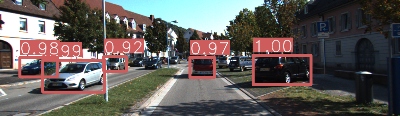} \hspace{0.085em} & 
  \adjincludegraphics[width=0.222\linewidth, trim={{0.28346456692\width} 0 {0.12047244094\width} 0}, clip]{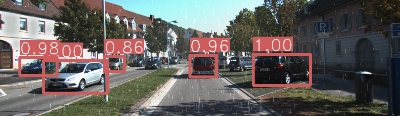} & 
  \adjincludegraphics[width=0.222\linewidth, trim={{0.28346456692\width} 0 {0.12047244094\width} 0}, clip]{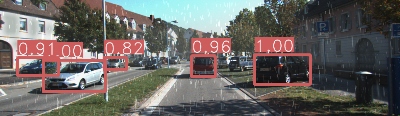} &
  \adjincludegraphics[width=0.222\linewidth, trim={{0.28346456692\width} 0 {0.12047244094\width} 0}, clip]{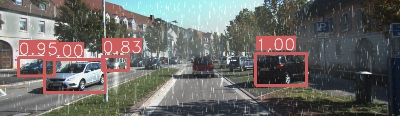} \\
	
  \multirow{1}{*}[1.35cm]{\rotatebox{90}{MX-RCNN} \rotatebox{90}{~~~~~~\cite{yang2016exploit}}} & 
	\adjincludegraphics[width=0.222\linewidth, trim={{0.28346456692\width} 0 {0.12047244094\width} 0}, clip]{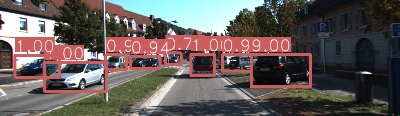} \hspace{0.085em} & 
  \adjincludegraphics[width=0.222\linewidth, trim={{0.28346456692\width} 0 {0.12047244094\width} 0}, clip]{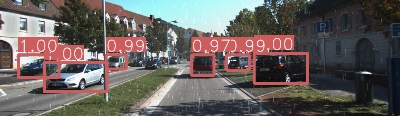} & 
  \adjincludegraphics[width=0.222\linewidth, trim={{0.28346456692\width} 0 {0.12047244094\width} 0}, clip]{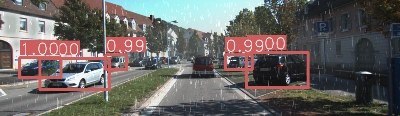} &
  \adjincludegraphics[width=0.222\linewidth, trim={{0.28346456692\width} 0 {0.12047244094\width} 0}, clip]{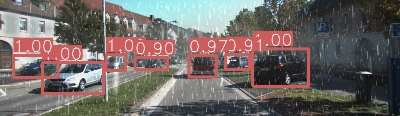} \\
	
  \multirow{1}{*}[1.25cm]{\rotatebox{90}{YOLOv2} \rotatebox{90}{~~~~\cite{redmon2017yolo9000}}} & 
	\adjincludegraphics[width=0.222\linewidth, trim={{0.28346456692\width} 0 {0.12047244094\width} 0}, clip]{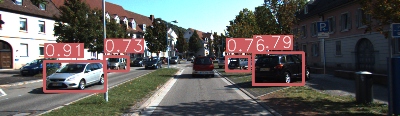} \hspace{0.085em} & 
  \adjincludegraphics[width=0.222\linewidth, trim={{0.28346456692\width} 0 {0.12047244094\width} 0}, clip]{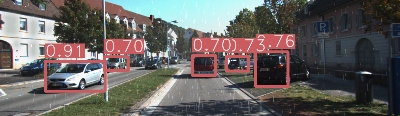} & 
  \adjincludegraphics[width=0.222\linewidth, trim={{0.28346456692\width} 0 {0.12047244094\width} 0}, clip]{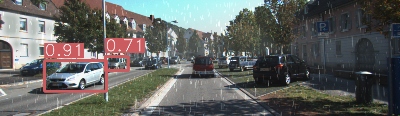} &
  \adjincludegraphics[width=0.222\linewidth, trim={{0.28346456692\width} 0 {0.12047244094\width} 0}, clip]{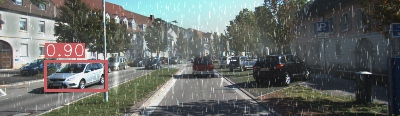} \\
	
  & Clear weather & Moderate rain & Heavy rain     & Shower rain \\
  &    & (50~mm/hr)  & (100~mm/hr)   & (200~mm/hr) \\
 \end{tabular}
 \caption{\textbf{Object detection on PBR rain augmentation of KITTI.} From left to right, the original image (clear) and three PBR augmentations with varying rainfall rates. Images are cropped for visualization.}
 \label{fig:eval_qualitative_objdet}
\end{figure*}

\renewcommand\weatherAugmented[4]{
 \adjincludegraphics[width=0.222\linewidth]{figures/results/qualitative_rain/#2/clear/#1.#3} \hspace{0.085em} &
 \adjincludegraphics[width=0.222\linewidth]{figures/results/qualitative_rain/#2/pbr/50mm/#1.#4} & 
 \adjincludegraphics[width=0.222\linewidth]{figures/results/qualitative_rain/#2/pbr/100mm/#1.#4} & 
 \adjincludegraphics[width=0.222\linewidth]{figures/results/qualitative_rain/#2/pbr/200mm/#1.#4}}

\begin{figure*}
 \newcommand\weatheraugvizSegT[4]{
  \includegraphics[width=0.222\linewidth]{figures/results/semantic_seg/#2/#1#3_clear_tb.#4} \hspace{0.085em} & 
  \includegraphics[width=0.222\linewidth]{figures/results/semantic_seg/#2/#1#3_50mm_tb.#4} & 
  \includegraphics[width=0.222\linewidth]{figures/results/semantic_seg/#2/#1#3_100mm_tb.#4} & 
  \includegraphics[width=0.222\linewidth]{figures/results/semantic_seg/#2/#1#3_200mm_tb.#4}}
 \newcommand\weatheraugvizSeg[2]{\weatheraugvizSegT{#1}{#2}{_#2}{png}}
 
 \centering
 \scriptsize
 \setlength{\tabcolsep}{0.001\linewidth}
 \renewcommand{\arraystretch}{0.5}
 \begin{tabular}{ccccc}
  & \textbf{\footnotesize{Original}}&\multicolumn{3}{c}{\textbf{\footnotesize{Rain augmented (PBR)}}} \\
  \cmidrule[1pt](l{2pt}r{4pt}){2-2}\cmidrule[1pt](l{2pt}r{2pt}){3-5}
  \multirow{1}{*}[1.1cm]{\rotatebox{90}{Input}} & 
	\adjincludegraphics[width=0.222\linewidth]{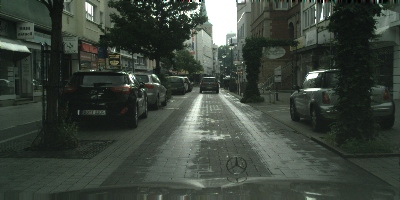} \hspace{0.085em} &
 \adjincludegraphics[width=0.222\linewidth]{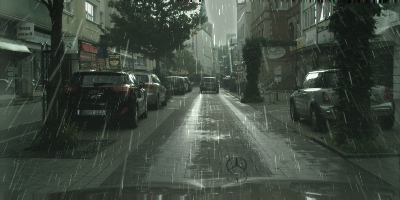} & 
 \adjincludegraphics[width=0.222\linewidth]{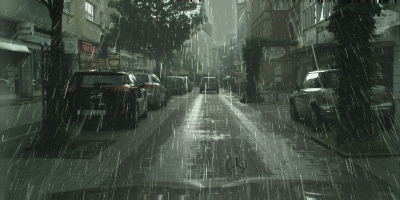} & 
 \adjincludegraphics[width=0.222\linewidth]{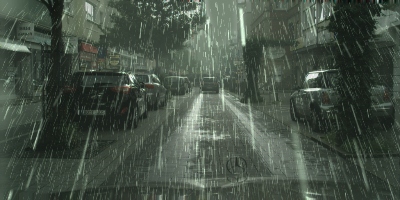} \\
	
  \multirow{1}{*}[1.25cm]{\rotatebox{90}{ERFNet} \rotatebox{90}{~~~\cite{romera2018erfnet}}} &
	\includegraphics[width=0.222\linewidth]{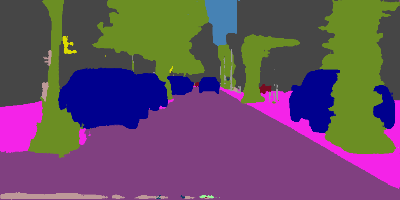} \hspace{0.085em} & 
  \includegraphics[width=0.222\linewidth]{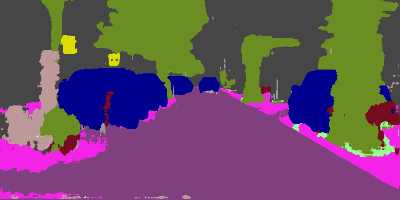} & 
  \includegraphics[width=0.222\linewidth]{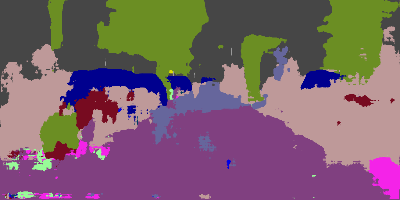} & 
  \includegraphics[width=0.222\linewidth]{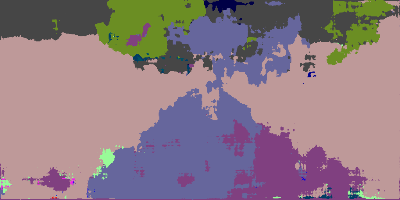} \\
	
  \multirow{1}{*}[1.25cm]{\rotatebox{90}{ESPNet} \rotatebox{90}{~~~\cite{mehta2018espnet}}} & 
	\includegraphics[width=0.222\linewidth]{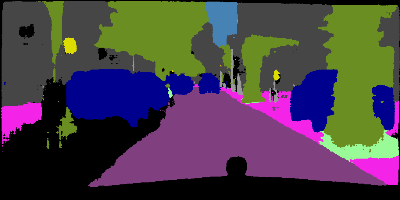} \hspace{0.085em} & 
  \includegraphics[width=0.222\linewidth]{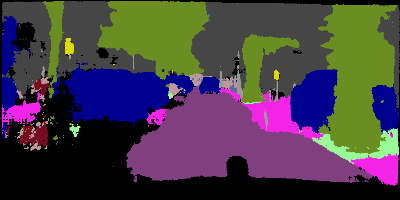} & 
  \includegraphics[width=0.222\linewidth]{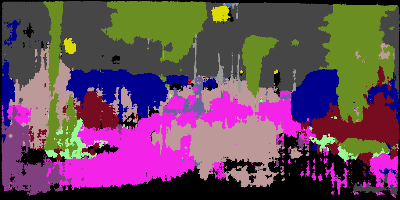} & 
  \includegraphics[width=0.222\linewidth]{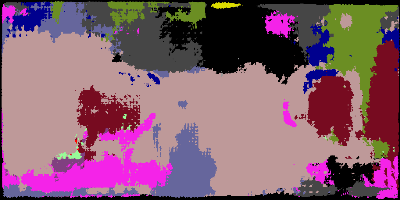} \\
	
  \multirow{1}{*}[1.25cm]{\rotatebox{90}{ICNet} \rotatebox{90}{~\cite{zhao2018icnet}}} & 
	\includegraphics[width=0.222\linewidth]{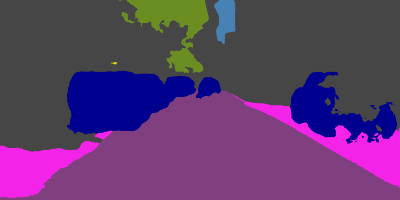} \hspace{0.085em} & 
  \includegraphics[width=0.222\linewidth]{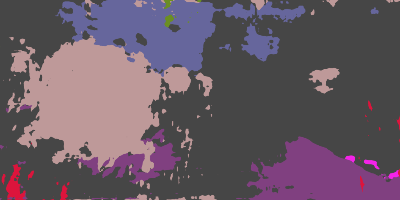} & 
  \includegraphics[width=0.222\linewidth]{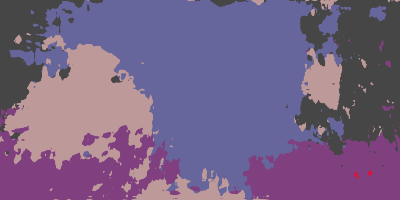} & 
  \includegraphics[width=0.222\linewidth]{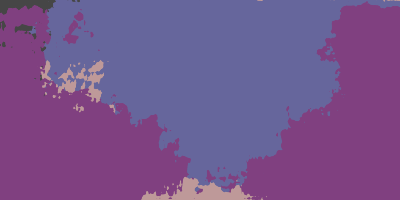} \\
	
  \multirow{1}{*}[1.25cm]{\rotatebox{90}{PSPNet} \rotatebox{90}{~~~\cite{zhao2017pspnet}}} & 
	\includegraphics[width=0.222\linewidth]{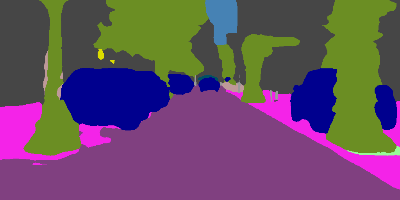} \hspace{0.085em} & 
  \includegraphics[width=0.222\linewidth]{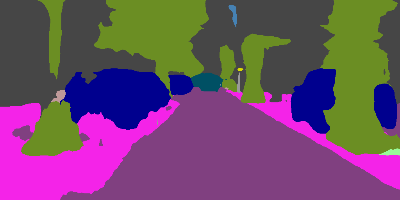} & 
  \includegraphics[width=0.222\linewidth]{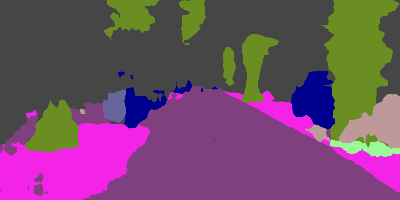} & 
  \includegraphics[width=0.222\linewidth]{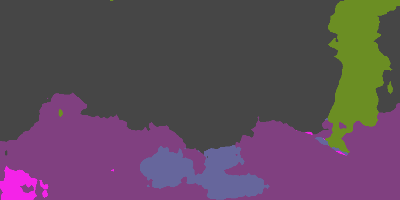} \\
	
  &Clear weather  & Moderate rain  & Heavy rain  & Shower rain \\
  &    & (50~mm/hr)        & (100~mm/hr)  & (200~mm/hr) \\
 \end{tabular}
 \caption{\textbf{Qualitative evaluation of semantic segmentation on our PBR rain augmentation of Cityscapes.} From left to right, the original image (clear) and three PBR augmentations with varying rainfall rates.}
 \label{fig:eval_qualitative_seg}
\end{figure*}

\begin{figure*}
 \newcommand\weatheraugvizDepthEstimT[3]{
  \includegraphics[width=0.222\linewidth]{figures/results/depth_estim/#2/clear/#1.#3} \hspace{0.085em} & 
  \includegraphics[width=0.222\linewidth]{figures/results/depth_estim/#2/50mm/#1.#3} & 
  \includegraphics[width=0.222\linewidth]{figures/results/depth_estim/#2/100mm/#1.#3} & 
  \includegraphics[width=0.222\linewidth]{figures/results/depth_estim/#2/200mm/#1.#3}}
 
 \centering
 \scriptsize
 \setlength{\tabcolsep}{0.001\linewidth}
 \renewcommand{\arraystretch}{0.5}
 \begin{tabular}{ccccc}
  & \textbf{\footnotesize{Original}}&\multicolumn{3}{c}{\textbf{\footnotesize{Rain augmented (PBR)}}} \\
  \cmidrule[1pt](l{2pt}r{4pt}){2-2}\cmidrule[1pt](l{2pt}r{2pt}){3-5}
  \multirow{1}{*}[1.25cm]{\rotatebox{90}{Input}} & 
	\adjincludegraphics[width=0.222\linewidth]{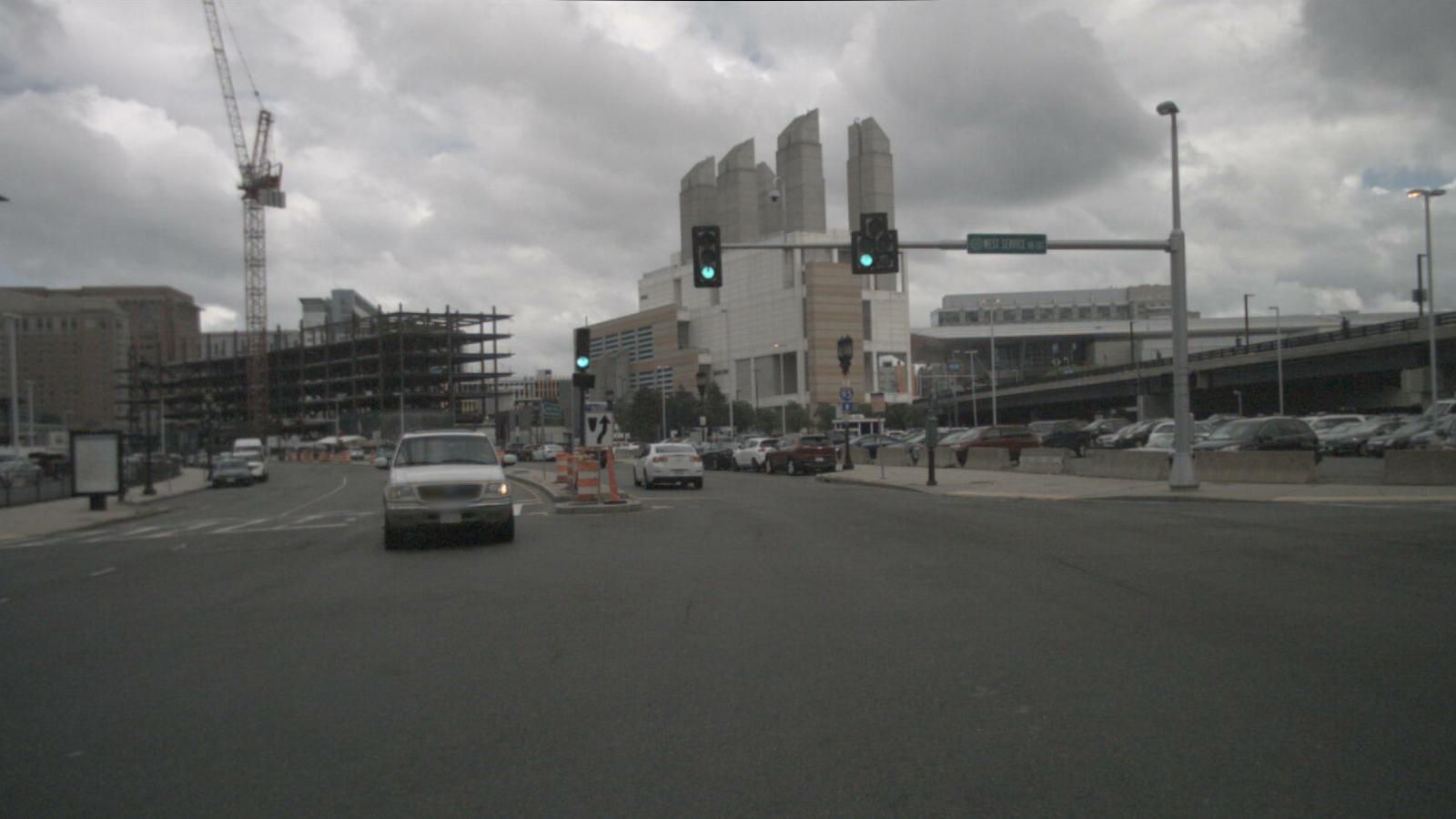} \hspace{0.085em} &
  \adjincludegraphics[width=0.222\linewidth]{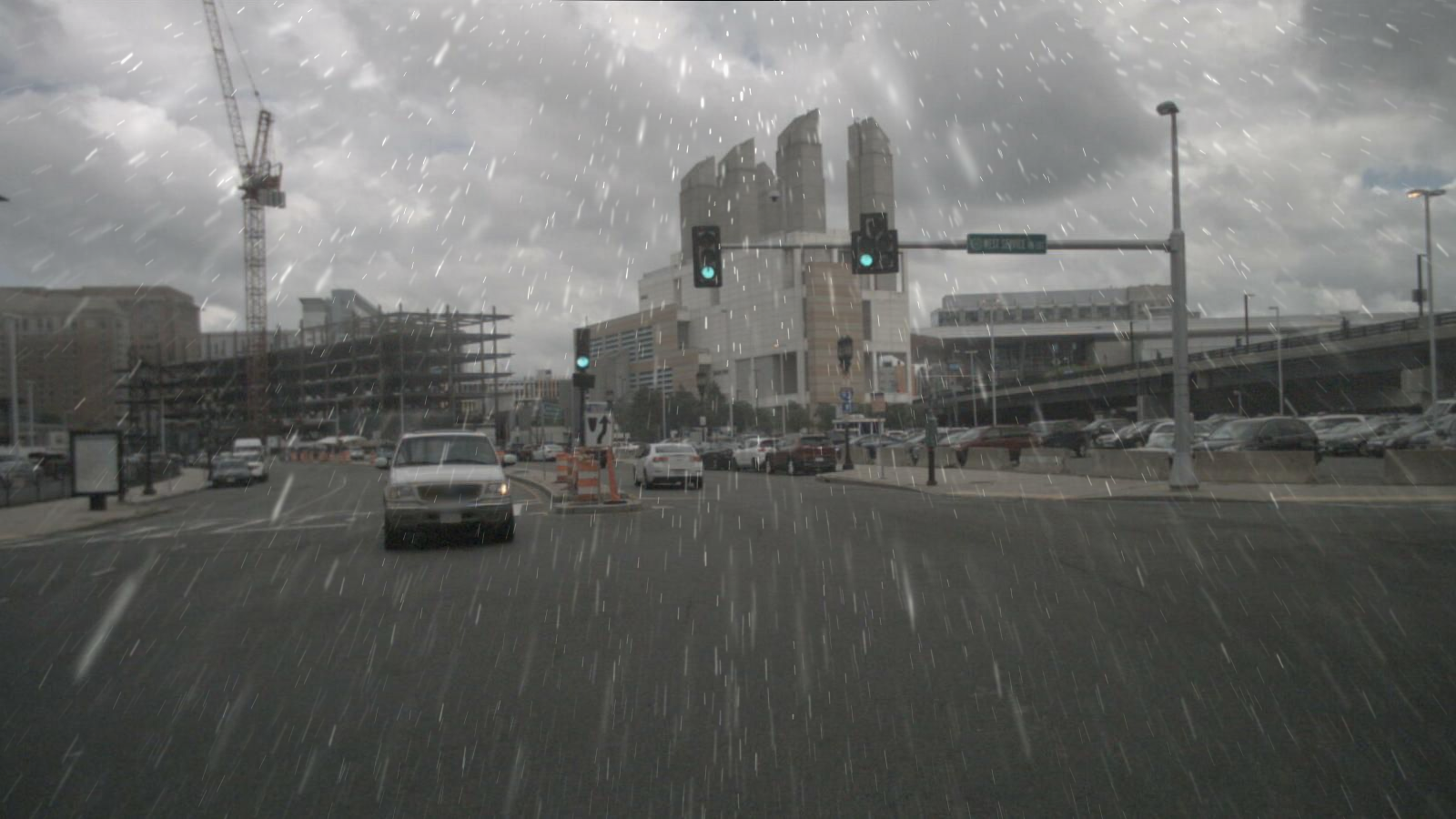} & 
  \adjincludegraphics[width=0.222\linewidth]{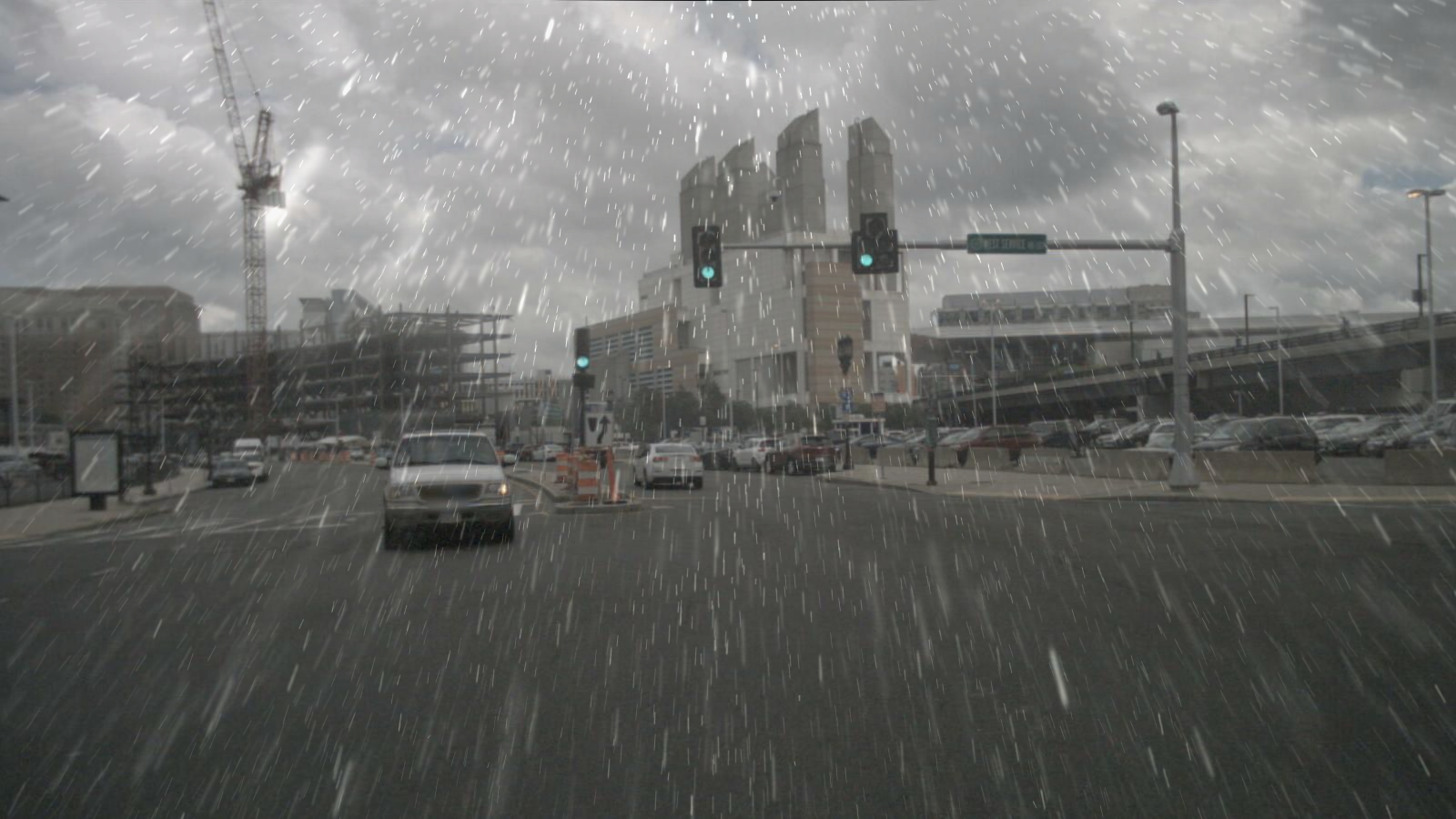} & 
  \adjincludegraphics[width=0.222\linewidth]{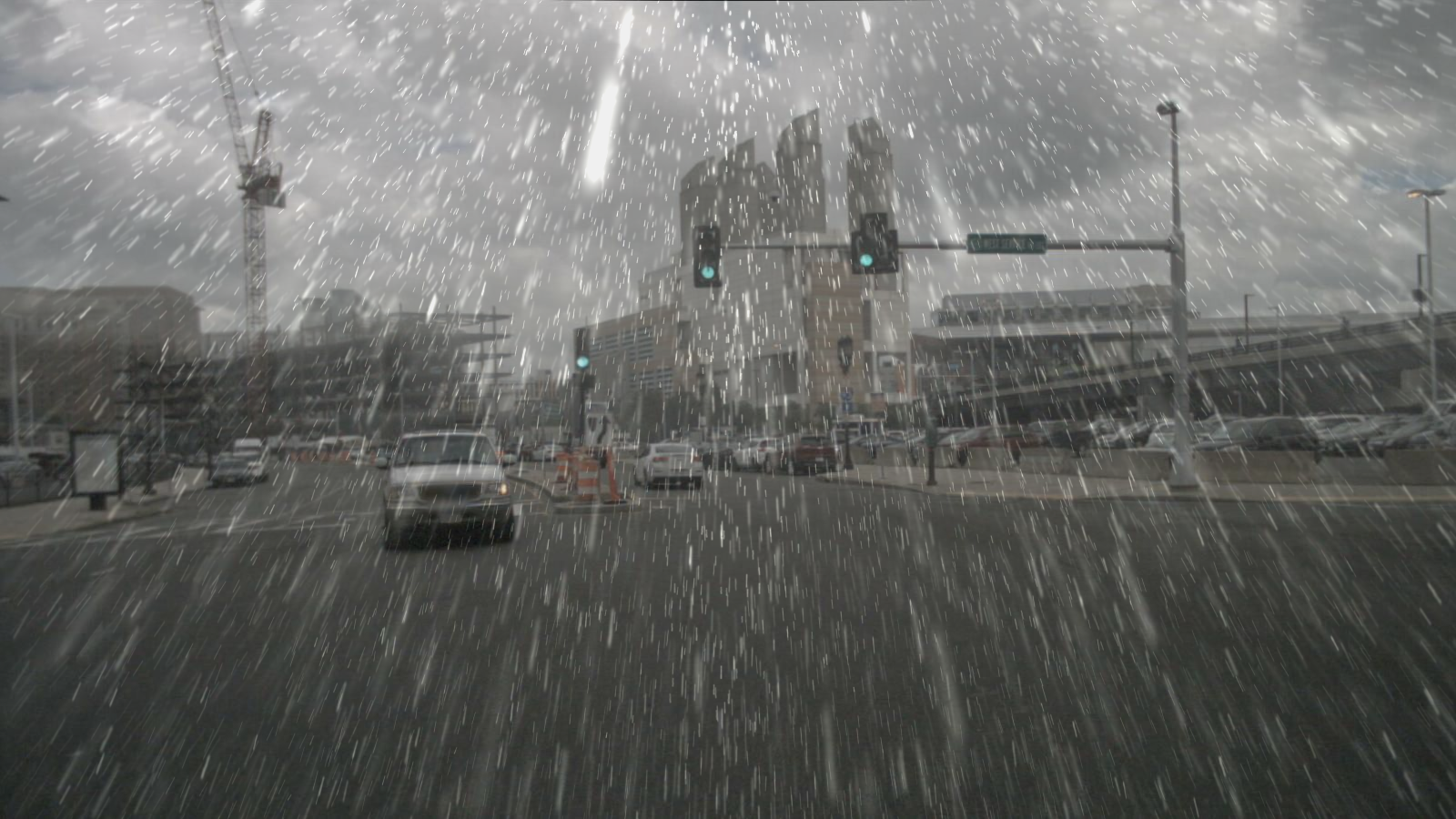} \\
	
  \multirow{1}{*}[1.7cm]{\rotatebox{90}{Monodepth2} \rotatebox{90}{~~~~~~~\cite{godard2019digging}}} &
	\includegraphics[width=0.222\linewidth]{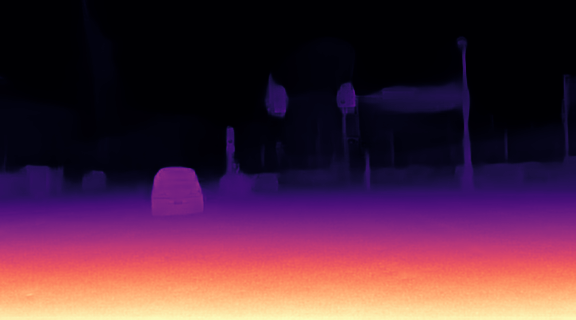} \hspace{0.085em} & 
  \includegraphics[width=0.222\linewidth]{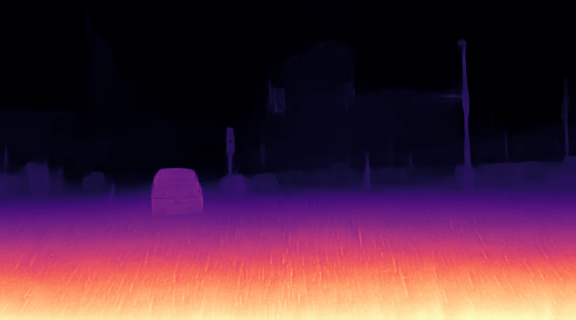} & 
  \includegraphics[width=0.222\linewidth]{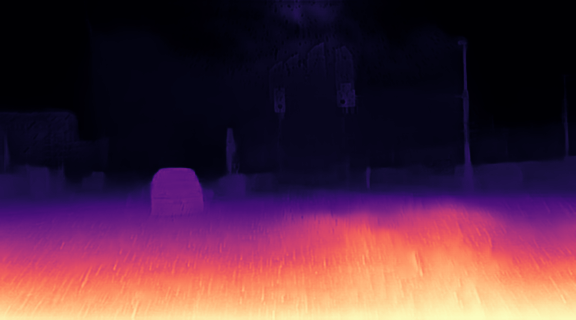} & 
  \includegraphics[width=0.222\linewidth]{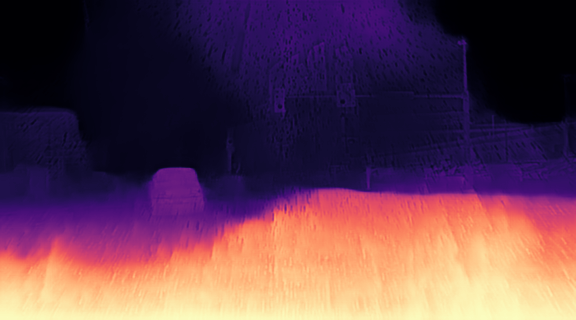} \\
	
  \multirow{1}{*}[1.2cm]{\rotatebox{90}{BTS} \rotatebox{90}{\cite{lee2020big}}} &
	\includegraphics[width=0.222\linewidth]{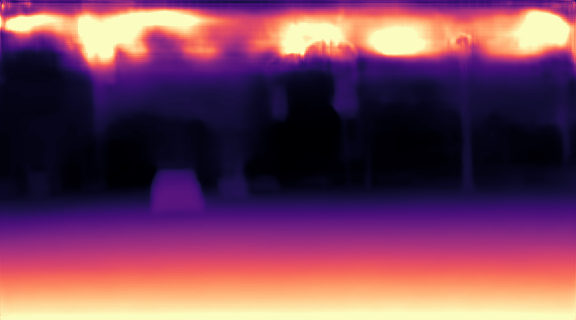} \hspace{0.085em} & 
  \includegraphics[width=0.222\linewidth]{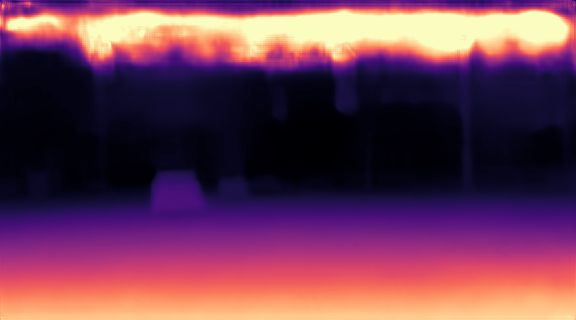} & 
  \includegraphics[width=0.222\linewidth]{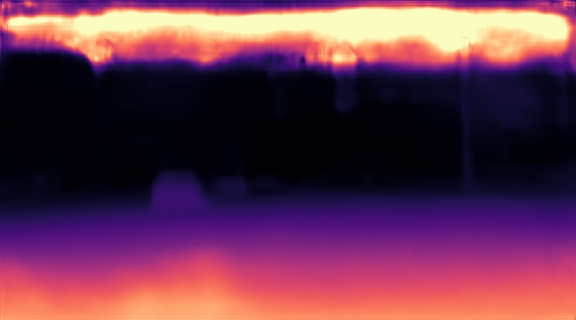} & 
  \includegraphics[width=0.222\linewidth]{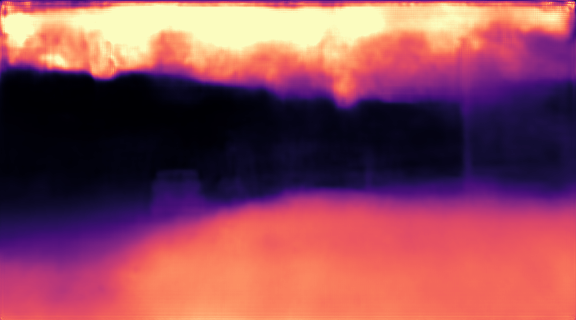} \\
		
  &Clear weather  & Moderate rain  & Heavy rain  & Shower rain \\
  &    & (50~mm/hr)        & (100~mm/hr)  & (200~mm/hr) \\
 \end{tabular}
 \caption{\textbf{Depth estimation on our PBR rain augmentation of nuScenes.} From left to right, the original image (clear) and three PBR augmentations with varying rainfall rates.}
 \label{fig:eval_qualitative_depthestm}
\end{figure*}

We compare the performance on PBR-augmented images for 6~object detection algorithms on KITTI~\cite{Geiger2012CVPR}~(7481 images), 6~segmentation algorithms on Cityscapes~\cite{Cordts2016Cityscapes}~(2995 images) and 2 depth estimation algorithm on nuScenes-augment~(4449 images). 
For all of the algorithms, the clear version always serves as a baseline to which we compare the performance on synthetic rain translations.

\paragraph{Object detection.}
We evaluate the 6 PBR augmented weathers on KITTI for 6~car/pedestrian pre-trained detection algorithms (with $\text{IoU}\geq.7$): DSOD~\cite{Shen2017DSOD}, Faster R-CNN~\cite{ren2015faster}, R-FCN~\cite{dai2016r}, SSD~\cite{liu2016ssd}, MX-RCNN~\cite{yang2016exploit}, and YOLOv2~\cite{redmon2017yolo9000}. Quantitative results for the Coco mAP@[.1:.1:.9] metric across classes are shown in fig.~\ref{fig:algo_obj_detect_perf}. Relative to their clear-weather performance the 200~mm/hr rain is always at least 12\% worse and even drops to 25-30\% for R-FCN, SSD, and MX-RCNN, whereas Faster R-CNN and DSOD are the most robust to changes in fog and rain. 

Representative qualitative results on PBR images are shown in fig.~\ref{fig:eval_qualitative_objdet} for 4 out of 6 algorithms to preserve space. All algorithms are strongly affected by the rain; it has a chaotic effect on object detection results because there can be large variance of occlusion level for objects populating the image. Also, as in real-life, far away objects (which are generally small objects) are more likely to disappear behind fog-like rain. 

\paragraph{Semantic segmentation.}
For semantic segmentation, the PBR augmented Cityscapes is evaluated for: AdaptSegNet~\cite{Tsai_adaptseg_2018}, ERFNet~\cite{romera2018erfnet}, ESPNet~\cite{mehta2018espnet}, ICNet~\cite{zhao2018icnet}, PSPNet~\cite{zhao2017pspnet} and PSPNet(50)~\cite{zhao2017pspnet}.
Quantitative results are reported in fig.~\ref{fig:algo_sem_seg_perf}. As opposed to object detection algorithms which demonstrated significant robustness to moderately high rainfall rates, here the algorithms seem to breakdown in similar conditions. Indeed, all techniques see their performance drop by a minimum of 30\% under heavy fog, and almost 60\% under strong rain. Interestingly, some curves cross, which indicates that different algorithms behave differently under rain. ESPNet for example, ranks among the top 3 in clear weather but drops relatively by a staggering 85\% and ranks last in stormy conditions (200mm/hr). Corresponding qualitative results are shown in fig.~\ref{fig:eval_qualitative_seg} for 4 out of 6 algorithms to preserve space. Although the effect of rain may appear minimal visually, it greatly affects the output of all segmentation algorithms evaluated. 

\paragraph{Depth estimation.}
We evaluate the performance of the recent Monodepth2~\cite{godard2019digging} and BTS~\cite{lee2020big} on the nuScenes-augment subset augmented with our PBR method.
We report the standard squared relative error in fig.~\ref{fig:algo_depth_estim_perf} and note that the error seems to increase linearly with rain. In the extreme 200mm/hr rain conditions, we measure an error of 3x that of clear images.
Qualitative results are shown in fig.~\ref{fig:eval_qualitative_depthestm}. It can be observed that performance drops when raindrops block the view or fog-like limits the visibility in the image.

\subsection{Evaluating GAN and GAN+PBR rain augmentations}

Next, we ascertain the effect of the rain augmentation with our GAN and GAN+PBR augmentation strategies. As mentioned above, semantic segmentation algorithms are not evaluated due to the lack of semantic labels for training. The evaluation is performed on the nuScenes-augment subset. 

\begin{figure}
 \centering
 \subfloat[Object detection~\cite{redmon2017yolo9000}]{\includegraphics[width=0.485\linewidth]{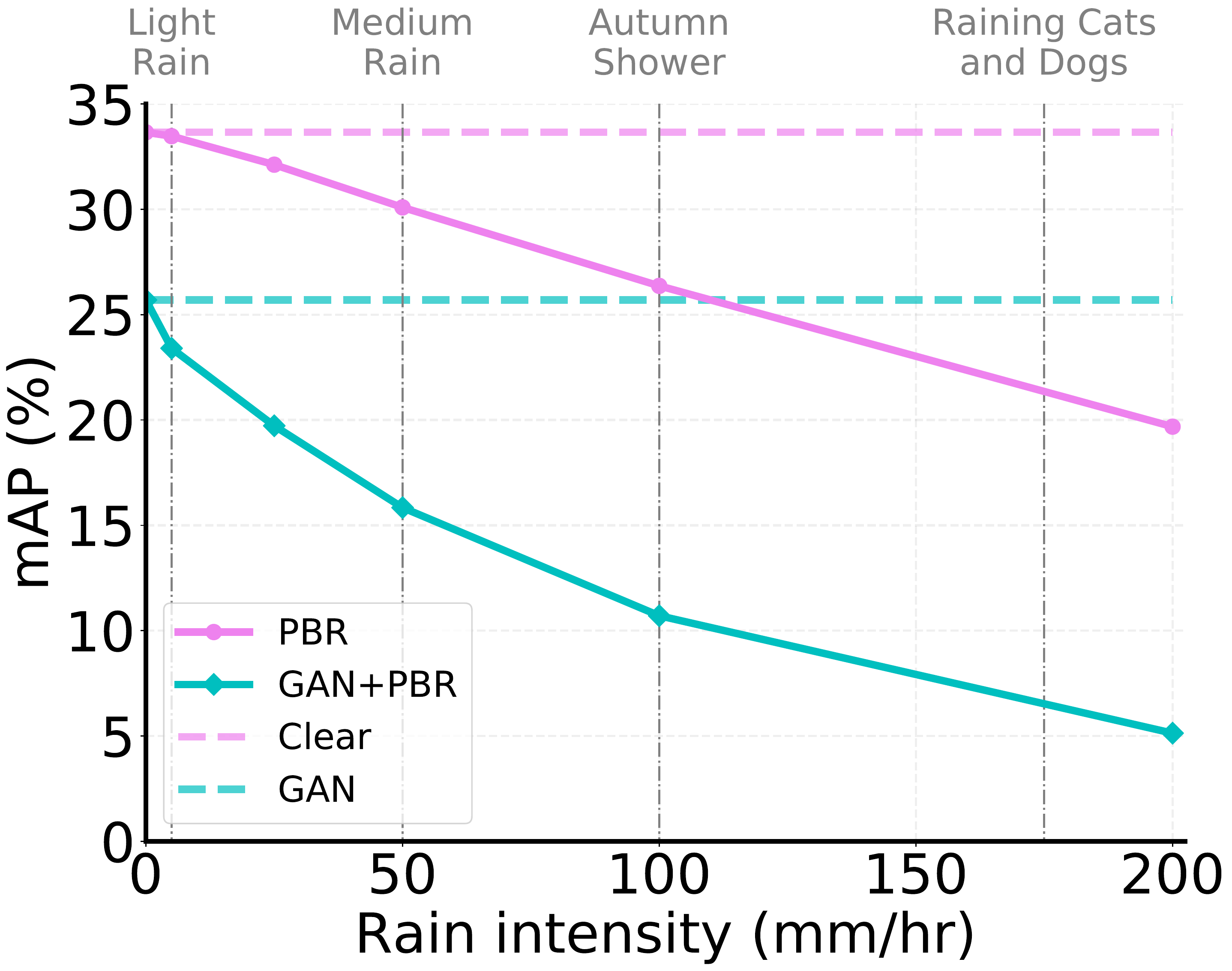} \label{fig:hybrid_obj_detect_perf}}\hfill%
    \subfloat[Depth estimation~\cite{godard2019digging}]{\includegraphics[width=0.485\linewidth]{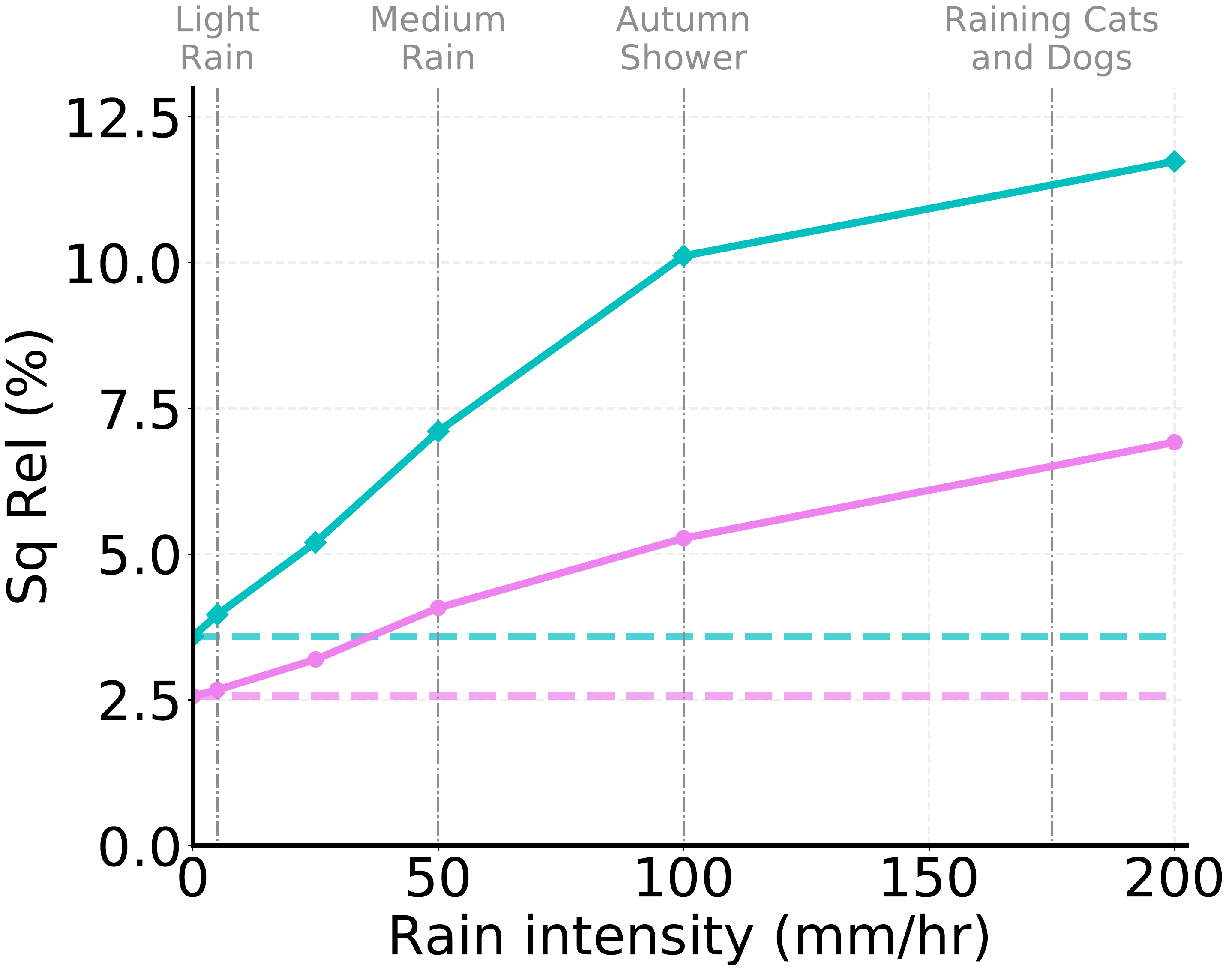} \label{fig:hybrid_depth_estim_perf}} 
 \caption{\textbf{Performance with varying rain intensities and augmentation techniques.} Opposed to PBR, GAN does not allow to control the rain intensity and is reported as dashed line, as for clear performance. Increasing rain intensity translates as a performance drop for \protect\subref{fig:hybrid_obj_detect_perf} object detection with YOLOv2~\cite{redmon2017yolo9000} and \protect\subref{fig:hybrid_depth_estim_perf} depth estimation with Monodepth2~\cite{godard2019digging}.}
 \label{fig:hybrid_perf}
\end{figure}

\begin{figure*}
 \newcommand\weatheraugvizHybridExp[2]{
 \includegraphics[width=0.191\linewidth]{figures/results/hybrid/#2/clear/#1.png} \hspace{0.085em} &
 \includegraphics[width=0.191\linewidth]{figures/results/hybrid/#2/gan/#1.png} &
 \includegraphics[width=0.191\linewidth]{figures/results/hybrid/#2/50mm/#1.png} & 
 \includegraphics[width=0.191\linewidth]{figures/results/hybrid/#2/100mm/#1.png} & 
 \includegraphics[width=0.191\linewidth]{figures/results/hybrid/#2/200mm/#1.png}}
 
 \centering
 \scriptsize
 \setlength{\tabcolsep}{0.001\linewidth}
 \renewcommand{\arraystretch}{0.5}
 \begin{tabular}{cccccc}
  & \textbf{\footnotesize{Original}}&\multicolumn{4}{c}{\textbf{\footnotesize{Rain augmented (GAN+PBR)}}} \\
  \cmidrule[1pt](l{2pt}r{4pt}){2-2}\cmidrule[1pt](l{2pt}r{2pt}){3-6}
  \multirow{1}{*}[1.15cm]{\rotatebox{90}{Input}} & 
	\adjincludegraphics[width=0.191\linewidth]{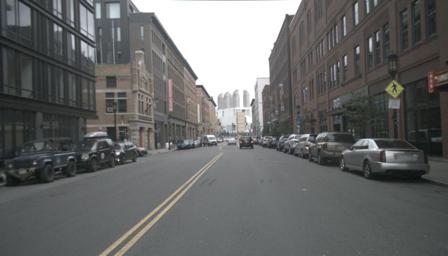} \hspace{0.085em} &
  \adjincludegraphics[width=0.191\linewidth]{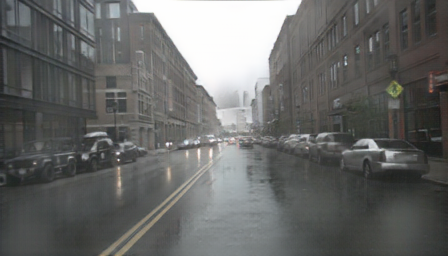} &
  \adjincludegraphics[width=0.191\linewidth]{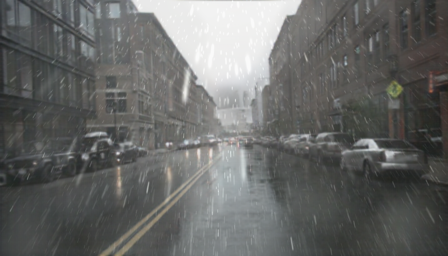} & 
  \adjincludegraphics[width=0.191\linewidth]{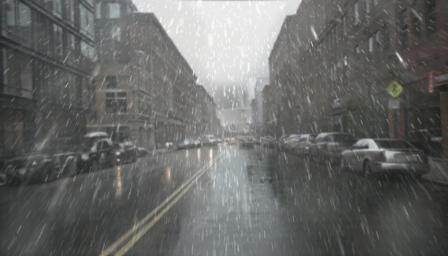} & 
  \adjincludegraphics[width=0.191\linewidth]{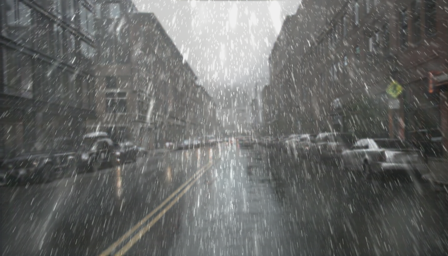} \\
	
  \multirow{1}{*}[1.25cm]{\rotatebox{90}{YOLOv2} \rotatebox{90}{~~~\cite{redmon2017yolo9000}}} &
	\includegraphics[width=0.191\linewidth]{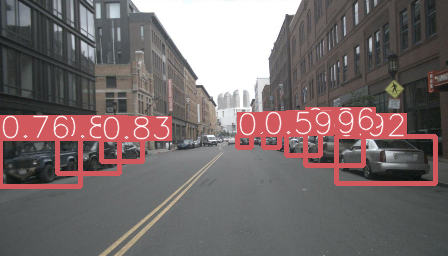} \hspace{0.085em} &
 \includegraphics[width=0.191\linewidth]{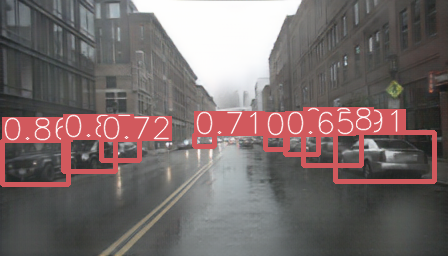} &
 \includegraphics[width=0.191\linewidth]{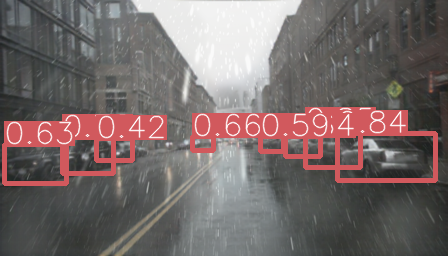} & 
 \includegraphics[width=0.191\linewidth]{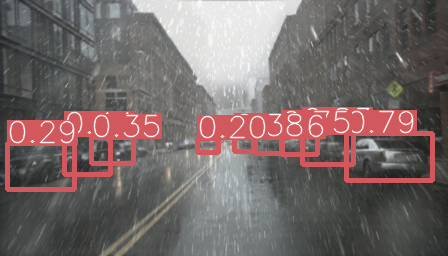} & 
 \includegraphics[width=0.191\linewidth]{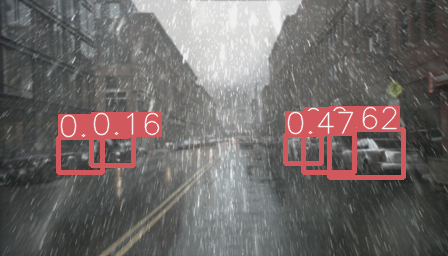} \\
	
  \vspace{0.45em} \\
  \multirow{1}{*}[1.15cm]{\rotatebox{90}{Input}} & 
	\adjincludegraphics[width=0.191\linewidth]{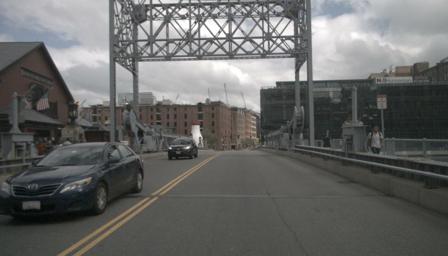} \hspace{0.085em} &
  \adjincludegraphics[width=0.191\linewidth]{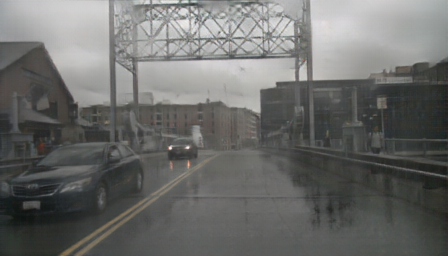} &
  \adjincludegraphics[width=0.191\linewidth]{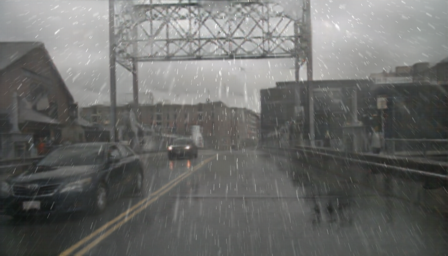} & 
  \adjincludegraphics[width=0.191\linewidth]{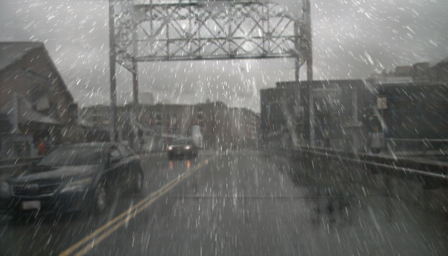} & 
  \adjincludegraphics[width=0.191\linewidth]{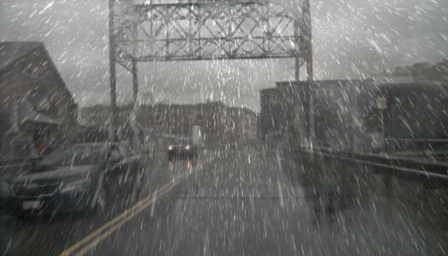} \\

  \multirow{1}{*}[1.25cm]{\rotatebox{90}{YOLOv2} \rotatebox{90}{~~~\cite{redmon2017yolo9000}}} &
	\includegraphics[width=0.191\linewidth]{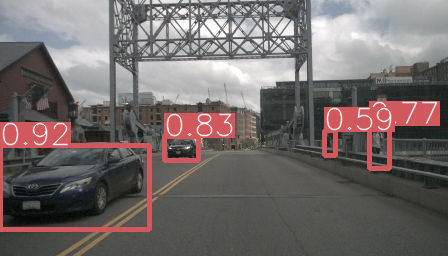} \hspace{0.085em} &
  \includegraphics[width=0.191\linewidth]{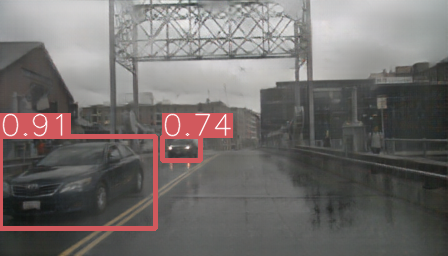} &
  \includegraphics[width=0.191\linewidth]{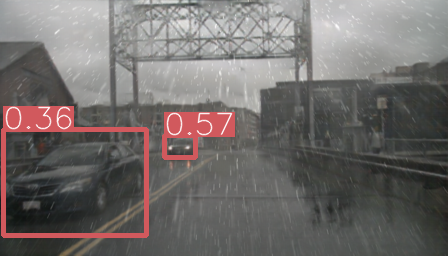} & 
  \includegraphics[width=0.191\linewidth]{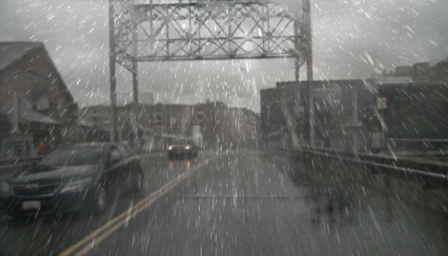} & 
  \includegraphics[width=0.191\linewidth]{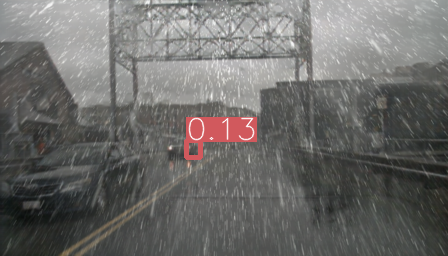} \\

  & Clear weather  & GAN only & Moderate rain  & Heavy rain  & Shower rain \\
  &      &     & (50~mm/hr)     & (100~mm/hr)  & (200~mm/hr) \\
 \end{tabular}
 \caption{\textbf{Object detection on our GAN+PBR augmented nuScenes.} From left to right, the original image (clear), the GAN augmented image and three GAN+PBR images.}
 \label{fig:eval_obj_detect_qualitative_hybrid}
\end{figure*}

\paragraph{Object detection.}
YOLOv2~\cite{redmon2017yolo9000} is evaluated, and the resulting mAP as a function of rainfall rates are reported in fig.~\ref{fig:hybrid_obj_detect_perf}. Qualitative results are shown in fig.~\ref{fig:eval_obj_detect_qualitative_hybrid}. We note that GAN augmented images have similar performance than PBR 100mm/hr images and that performance deterioration is stronger and steeper with GAN+PBR images. 

Observe that the particles physical simulation is the same in both PBR and GAN+PBR. Still, it is interesting to notice that the decrease is different with GAN+PBR compared to PBR (i.e. curves are shifted but also exhibit different slopes). This may be the result of the two cumulative domain shifts (i.e. wetness + streaks) leading to a non-linear effect.

\begin{figure*}
 \newcommand\weatheraugvizHybridExp[2]{
 \includegraphics[width=0.191\linewidth]{figures/results/hybrid/#2/clear/#1.png} \hspace{0.085em} &
 \includegraphics[width=0.191\linewidth]{figures/results/hybrid/#2/gan/#1.png} &
 \includegraphics[width=0.191\linewidth]{figures/results/hybrid/#2/50mm/#1.png} & 
 \includegraphics[width=0.191\linewidth]{figures/results/hybrid/#2/100mm/#1.png} & 
 \includegraphics[width=0.191\linewidth]{figures/results/hybrid/#2/200mm/#1.png}}
 
 \centering
 \scriptsize
 \setlength{\tabcolsep}{0.001\linewidth}
 \renewcommand{\arraystretch}{0.5}
 \begin{tabular}{cccccc}
  & \textbf{\footnotesize{Original}}&\multicolumn{4}{c}{\textbf{\footnotesize{Rain augmented (GAN+PBR)}}} \\
  \cmidrule[1pt](l{2pt}r{4pt}){2-2}\cmidrule[1pt](l{2pt}r{2pt}){3-6}
  \multirow{1}{*}[1.15cm]{\rotatebox{90}{Input}} & 
	\adjincludegraphics[width=0.191\linewidth]{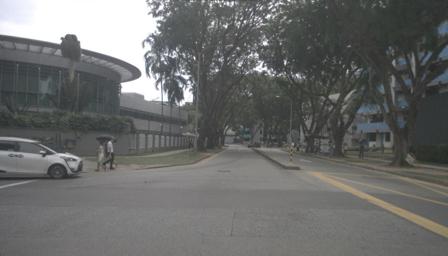} \hspace{0.085em} &
  \adjincludegraphics[width=0.191\linewidth]{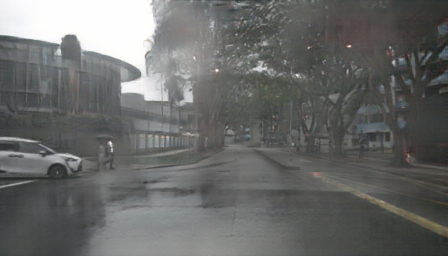} &
  \adjincludegraphics[width=0.191\linewidth]{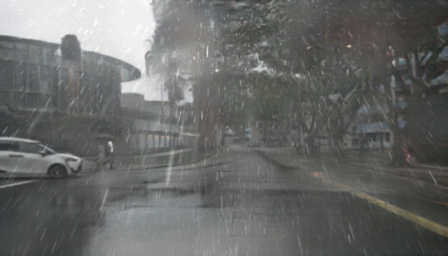} & 
  \adjincludegraphics[width=0.191\linewidth]{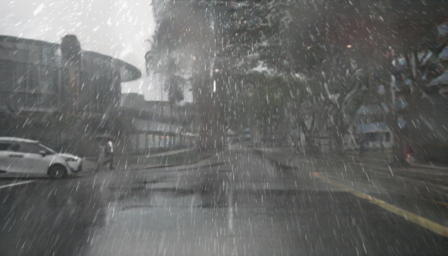} & 
  \adjincludegraphics[width=0.191\linewidth]{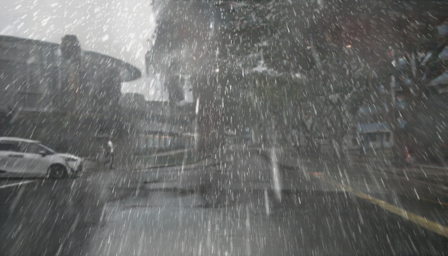} \\ 
	
  \multirow{1}{*}[1.50cm]{\rotatebox{90}{Monodepth2} \rotatebox{90}{~~~~~\cite{godard2019digging}}} &
	\includegraphics[width=0.191\linewidth]{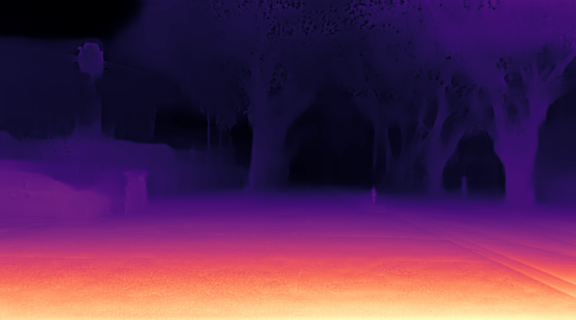} \hspace{0.085em} &
  \includegraphics[width=0.191\linewidth]{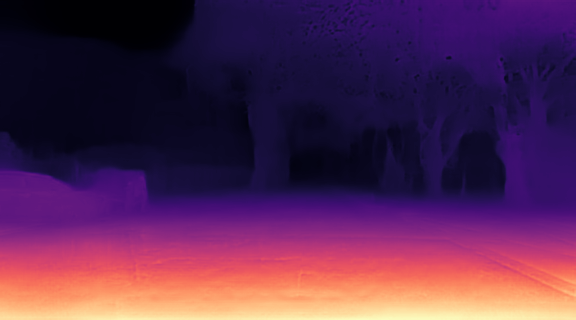} &
  \includegraphics[width=0.191\linewidth]{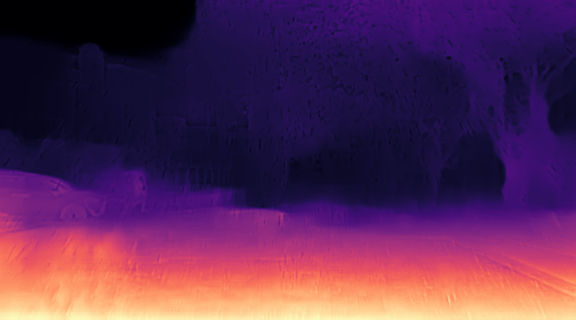} & 
  \includegraphics[width=0.191\linewidth]{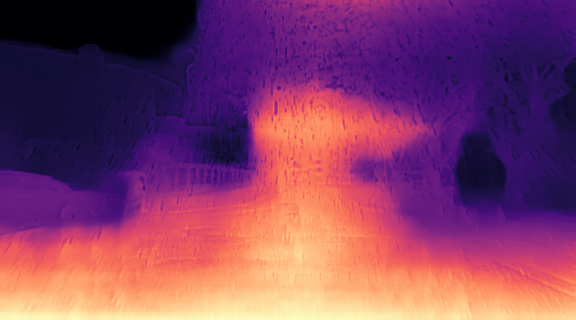} & 
  \includegraphics[width=0.191\linewidth]{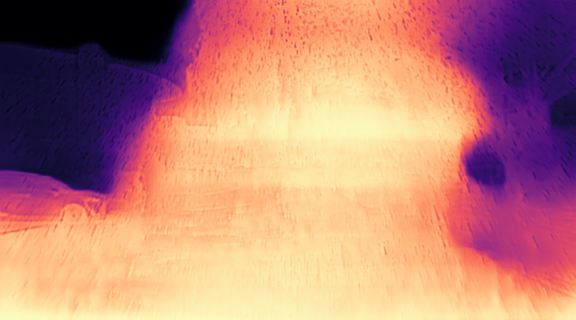} \\
	
  \vspace{0.45em} \\
  \multirow{1}{*}[1.15cm]{\rotatebox{90}{Input}} & 
	\adjincludegraphics[width=0.191\linewidth]{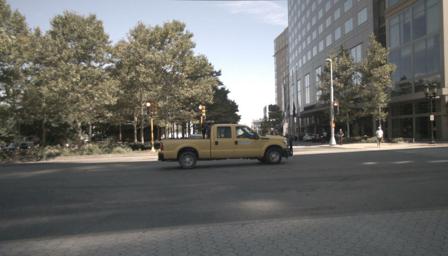} \hspace{0.085em} &
  \adjincludegraphics[width=0.191\linewidth]{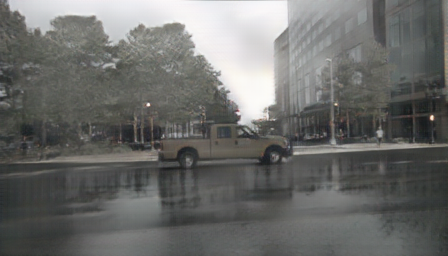} &
  \adjincludegraphics[width=0.191\linewidth]{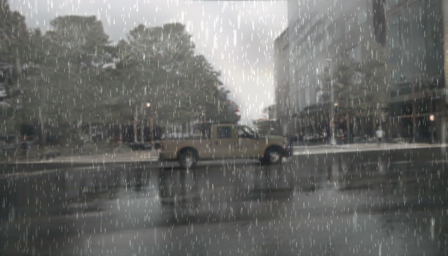} & 
  \adjincludegraphics[width=0.191\linewidth]{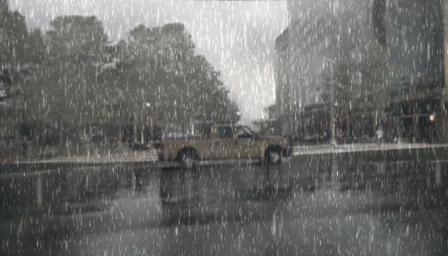} & 
  \adjincludegraphics[width=0.191\linewidth]{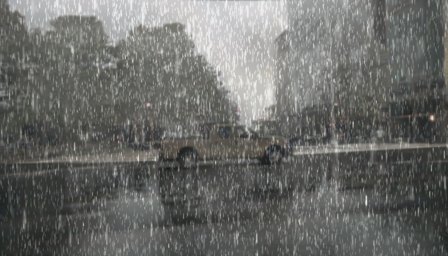} \\
	
  \multirow{1}{*}[1.50cm]{\rotatebox{90}{Monodepth2} \rotatebox{90}{~~~~~\cite{godard2019digging}}} &
	\includegraphics[width=0.191\linewidth]{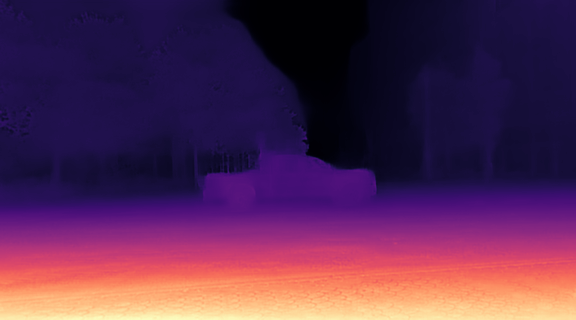} \hspace{0.085em} &
  \includegraphics[width=0.191\linewidth]{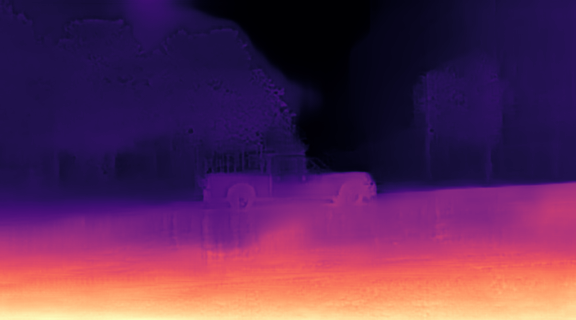} &
  \includegraphics[width=0.191\linewidth]{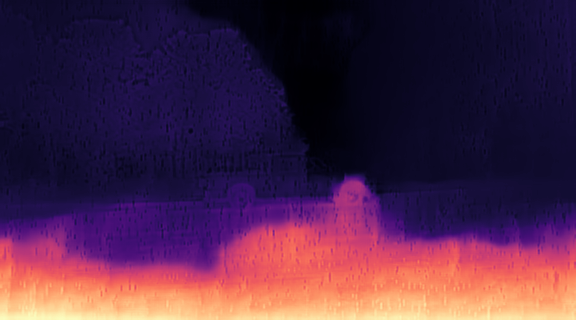} & 
  \includegraphics[width=0.191\linewidth]{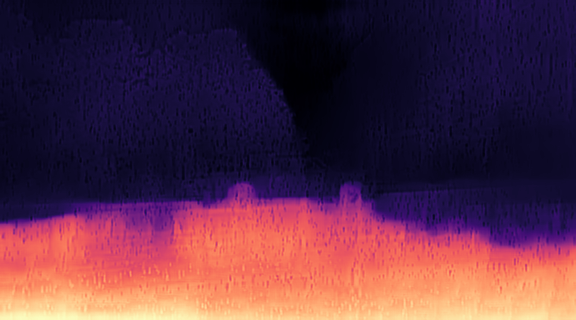} & 
  \includegraphics[width=0.191\linewidth]{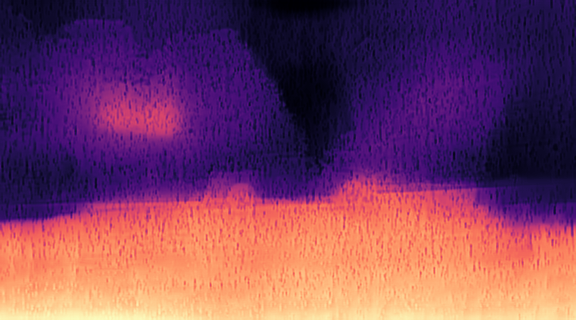} \\
	
  & Clear weather  & GAN only  & Moderate rain  & Heavy rain  & Shower rain \\
  &      &     & (50~mm/hr)     & (100~mm/hr)  & (200~mm/hr) \\
 \end{tabular}
 \caption{\textbf{Depth estimation on our GAN+PBR augmented nuScenes.} From left to right, the original image (clear), the GAN augmented image and three GAN+PBR images.}
 \label{fig:eval_depth_estim_detect_qualitative_hybrid}
\end{figure*}

\paragraph{Depth estimation.}
We evaluate the performance of the recent Monodepth2~\cite{godard2019digging} on the nuScenes-augment subset augmented with our GAN and GAN+PBR methods. 
As reported in fig.~\ref{fig:hybrid_depth_estim_perf}, for the same rain intensity between PBR and GAN+PBR images, the error is worse by a factor of 80--100\%. The same behavior was observed on other standard depth estimation metrics (absolute square error, RMSE, log RMSE), not reported here. Interestingly, the GAN augmentation affects only slightly the performance on depth estimation which might be because GAN translation keeps occlusion of the image to a minimum.
Qualitative results are shown in fig.~\ref{fig:eval_depth_estim_detect_qualitative_hybrid} for GAN and GAN+PBR. 

\section{Improving the robustness to rain}
\label{sec:improving}

We now wish to demonstrate the usefulness of our rain rendering pipeline for improving robustness to rain through extensive evaluations on synthetic and real rain databases. For the sake of coherence, the improvements are shown on the same tasks, algorithms, and test data from sec.~\ref{sec:real_data}.

\subsection{Training methodology}

While the ultimate goal is to improve robustness to rain, we aim at training a single model which performs well across a wide variety of rainfall rates (including clear weather). Having a single model is beneficial over employing, e.g., intensity-specific encoders~\cite{porav2019can}, since it removes the need for determining rain intensity from the input image. 
Because rain significantly alters the appearance of the scene, we found that training from scratch with heavy rain or random rainfall rates fails to converge.
Instead, we refine our \textit{untuned} models using curriculum learning~\cite{bengio2009curriculum} on rain intensity in ascending order (25, then 50, and finally 100mm/hr rain). The final model is referred as \textit{finetuned} and is evaluated against various weather conditions. Note that, for hybrid augmentation finetuning, the curriculum starts with the refinement on GAN images first and then go through the ascending rain intensities. The same images are used for all steps of the curriculum.

In order to avoid training and testing on the same set of images, in this section, we further divide the nuScenes-augment into train/test subsets of 1000 images each (ensuring they are taken from different scenes). 
Each algorithm is thus refined on the 1000 images from nuScenes-augment(train), and undergo a specific training process. For object detection, YOLOv2~\cite{redmon2017yolo9000} is trained each step at a learning rate of $0.0001$ and a momentum of 0.9 for 10 epochs with a burn-in of 5 epochs. For semantic segmentation, PSPNet~\cite{zhao2017pspnet} is trained with a learning rate of $0.0004$ and a momentum of 0.9 for 10 epochs. Finally, for depth estimation, Monodepth2~\cite{godard2019digging} is trained on triplets of consecutive images using a learning rate of $0.00001$ with the Adam optimizer for 10 epochs with $\beta = \left\{ 0.5, 0.999 \right\}$.  

\subsection{Improvement on synthetic rain}
\label{sec:perf_synth_data}

The synthetic evaluation is conducted on the set of 1000 images from nuScenes-augment(test), with rain up to 200mm/hr. Note again, for our hybrid GAN+PBR, 0mm/hr of rain correspond to the GAN-only augmented results.

\begin{figure}
 \centering
 \subfloat[Object detection~\cite{redmon2017yolo9000}]{\includegraphics[width=0.485\linewidth]{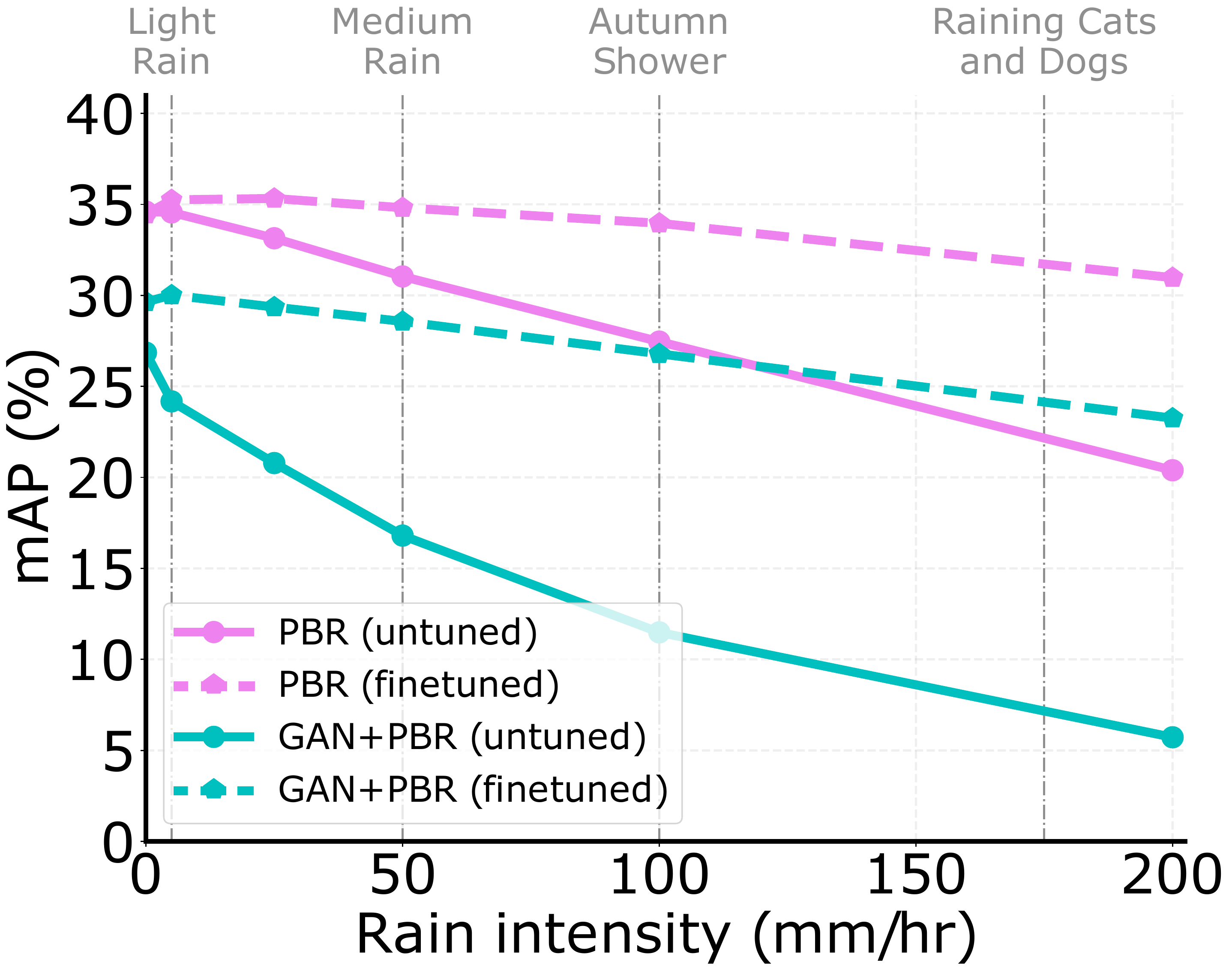}\label{fig:synth_eval_obj}} \hfill%
 \subfloat[Depth estimation~\cite{godard2019digging}]{\includegraphics[width=0.485\linewidth]{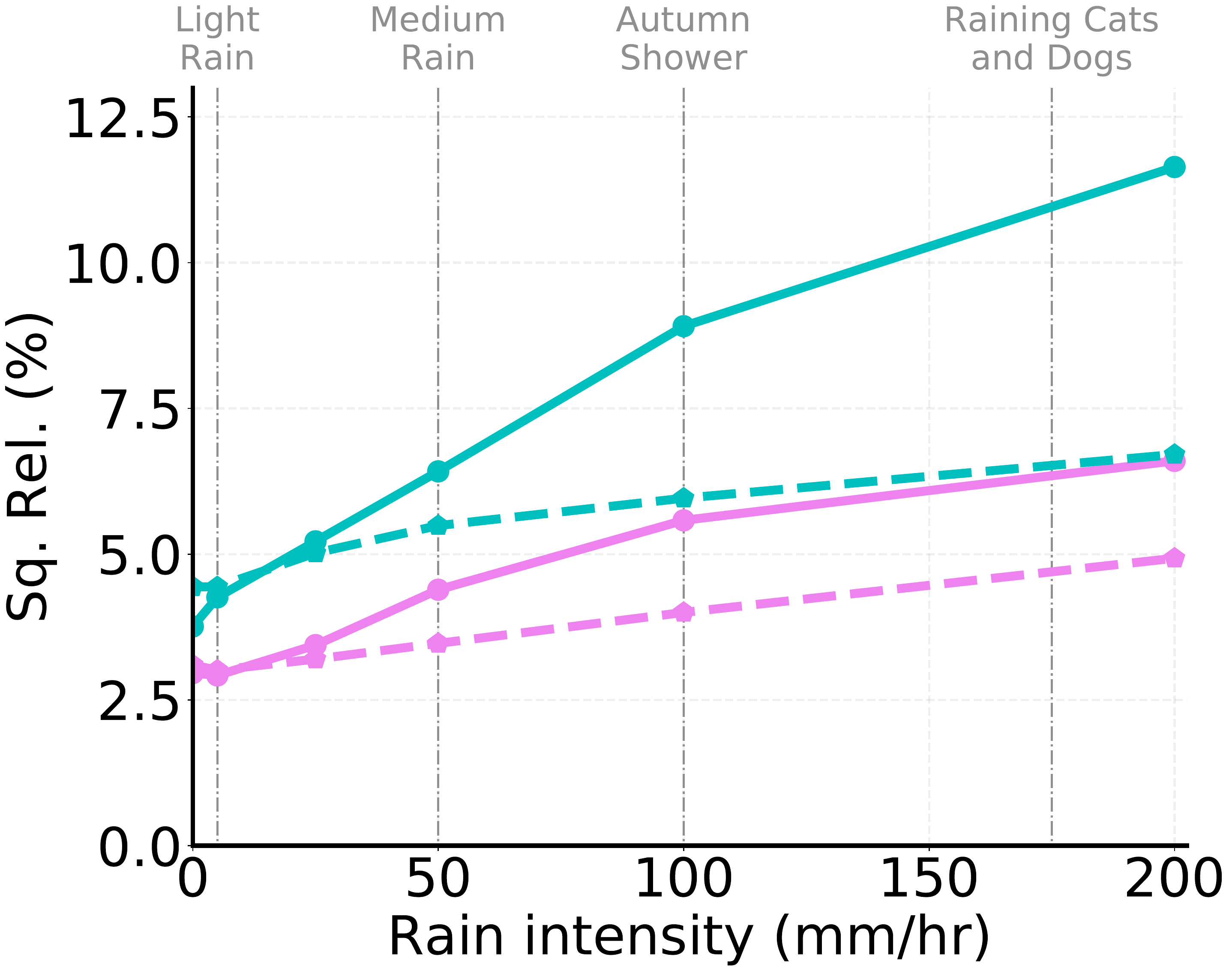}\label{fig:synth_eval_depth}} 
 
 \subfloat[Semantic segmentation~\cite{zhao2017pspnet}]{\includegraphics[width=0.485\linewidth]{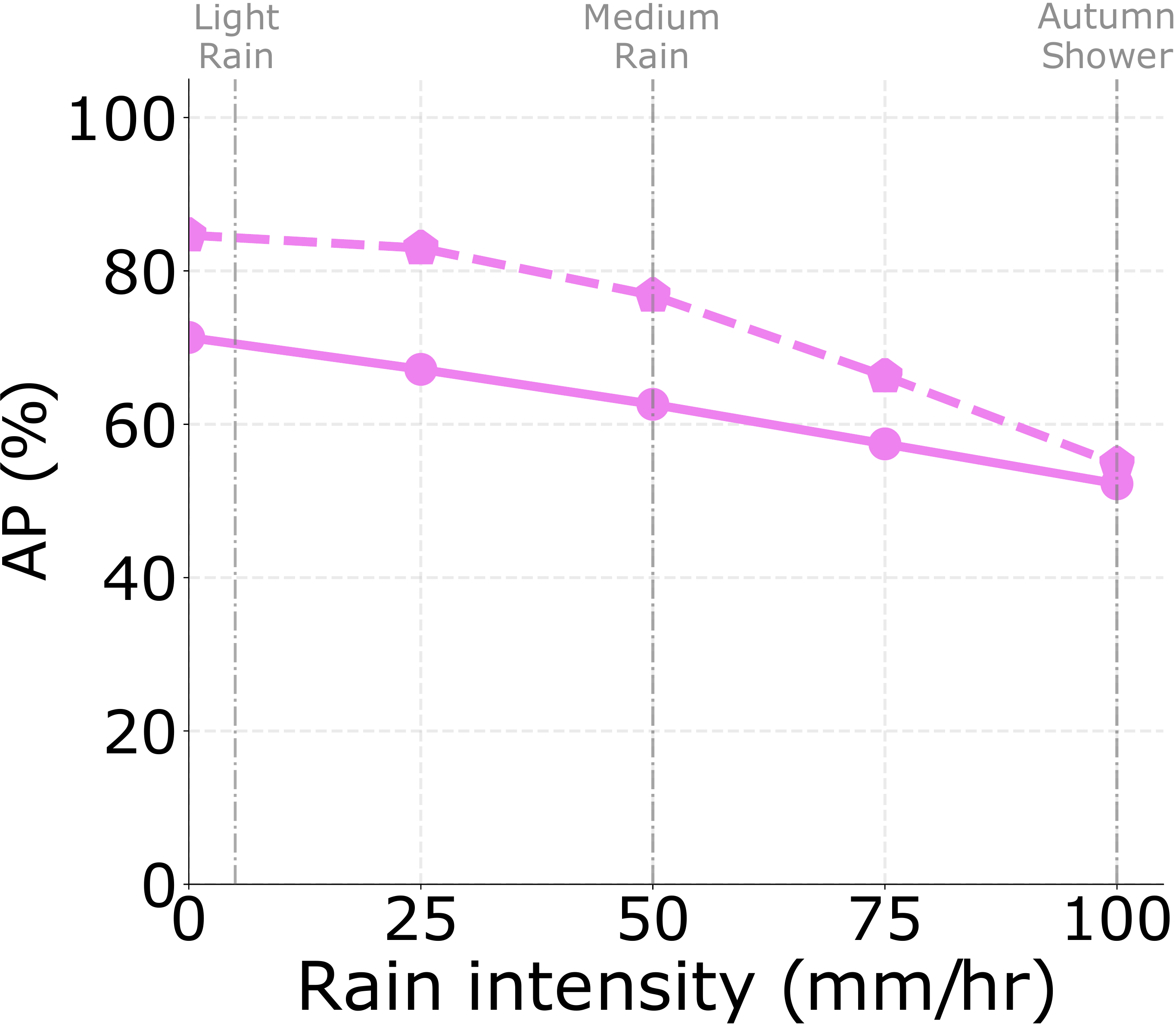}\label{fig:synth_eval_segm}}
 \caption{\textbf{Original (untuned) or finetuned performance on rain-augmented versions of nuScenes~\protect\subref{fig:synth_eval_obj}-\protect\subref{fig:synth_eval_depth} and Cityscapes~\protect\subref{fig:synth_eval_segm}.} Not only the finetuned models significantly outperform untuned models, but they exhibit a lower decrease with rain intensity, demonstrating increased robustness to both rain and clear weather.}
 \label{fig:synth_eval}
\end{figure}

Fig.~\ref{fig:synth_eval} shows the performance of our untuned and finetuned model for the three vision tasks on different augmented dataset. Fig.~\ref{fig:synth_eval_obj} and fig.~\ref{fig:synth_eval_depth} are for object detection (YOLOv2~\cite{redmon2017yolo9000}) and depth estimation (Monodepth2~\cite{godard2019digging}) on nuScenes-clear augmented data while fig.~\ref{fig:synth_eval_segm} is for the semantic segmentation (PSPNet~\cite{zhao2017pspnet}) on Cityscapes. We observe a significant improvement in both tasks and additional increase in robustness even in clear weather when refined using our augmented rain. Of interest, we also improve at the unseen 200mm/hr rain though the network was only trained with rain up to 100mm/hr.
The intuition here is that when facing adverse weather, the network learns to focus on strongest relevant features for all tasks and thus gain robustness. 

\paragraph{PBR.} For YOLOv2, the finetuned detection performance stays higher than its clear untuned counterpart in the 0-200mm/hr interval. Explicitly, it goes from 34.5\% to 31.0\% whereas the untuned model starts at 34.6\% and finishes at 20.4\%. For PSPNet, the segmentation exhibits a significative improvement when refined although at 100mm/hr the model is not fully able to compensate the effect of rain and drops to 54.0\% versus 52.0\% when untuned. Monodepth2 finetuning helps only for higher rain intensity level (+25mm/hr) and the error differences between 100mm/hr and 200mm/hr stay in the same ballpark (\textasciitilde{}1.2\%). This makes sense considering that since the occlusion created by rain streaks is minimum with a low rain intensity, the untuned model would not be strongly affected.

\paragraph{GAN+PBR.} In the case of YOLOv2, we notice a major difference between the hybrid  GAN+PBR untuned and finetuned performances. Indeed, the hybrid finetuned performance at 100mm/hr is at 21.7\% and only at a measly 7.5\% for the untuned model. The same goes for Monodepth2 for hybrid images performance with 5.7\% and 8.9\% at 100mm/hr for finetuned and untuned respectively. 
It is interesting to note that, for all tasks, performance evaluated on finetuned hybrid image decrease slower than for untuned models. This demonstrates again that more robust models are learned when finetuning with our rain translations.

\subsection{Improvement on real rain}
\label{sec:perf_real_data}

We evaluate the performance on real rain, using our nuScenes-rain(test) subset of images (see sec.~\ref{sec:real_data}).
Table~\ref{tab:impr_finetune} shows that our finetuning leads to performance increase in real rainy scenes compared to untuned performance in rain. We note for object detection (PBR: +20.7\%, GAN: +10.9\%, GAN+PBR: +21.0\%), for semantic segmentation (PBR: +36.9\%), and for depth estimation (PBR: 0.0\%, GAN: +3.8\%, GAN+PBR: +8.2\%) tasks. 
In clear weather, our finetuned model performs on par with the untuned version, sometimes even better. This boost in performance could be seen as the network learning to rely on more robust features, somehow invariant to rain streaks. 

Depth estimation underperformance for PBR finetuning can be explained by the learning loss of Monodepth2 which is, in short, a reprojection error which would not fare well with rain streaks as they do not reproject in consecutive frames. 
Interestingly, this problem does not seem to affect the GAN or GAN+PBR finetuned model, possibly because the GAN is being trained on split of nuScenes subsequently leading to finetuning images that are more resembling of the test set.
These results demonstrate the usefulness of our different rain rendering frameworks for real rain scenarios.

\begin{table}
 \centering
 \scriptsize
 \setlength{\tabcolsep}{0.125cm}
 \renewcommand{\arraystretch}{1.25}
 \stackunder[2pt]{\begin{tabular}{lcccccc}
  \toprule
  & \multicolumn{2}{c}{Object det. \cite{redmon2017yolo9000}} & \multicolumn{2}{c}{Semantic seg. \cite{zhao2017pspnet}} & \multicolumn{2}{c}{Depth est. \cite{godard2019digging}} \\
  & \multicolumn{2}{c}{\tiny{mAP (\%) $\uparrow$}} & \multicolumn{2}{c}{\tiny{AP (\%) $\uparrow$}} & \multicolumn{2}{c}{\tiny{Sq. err. (\%) $\downarrow$}} \\ 
  \cmidrule(lr){2-3}\cmidrule(lr){4-5}\cmidrule(lr){6-7}
  & Clear & Rain & Clear & Rain & Clear & Rain \\
  \midrule
  Untuned & 32.53 & 16.30 & \textbf{40.8} & 18.7 & 2.96 & 3.53 \\
  Finetuned \tiny{(PBR)} & \textbf{33.51} & 19.68 & 39.0 & \textbf{25.6} & 3.15 & 3.54 \\
  Finetuned \tiny{(GAN)} & 32.26 & 18.07 & * & * & \textbf{2.89} & 3.40 \\
  Finetuned \tiny{(GAN+PBR)} & 30.59 & \textbf{19.73} & * & * & 3.01 & \textbf{3.29} \\ 
  \midrule
  De-rained \tiny{DualResNet} & 32.60 & 18.30 & * & * & 2.25 & 3.09 \\
  \bottomrule
 \end{tabular}}{* Not evaluated due to lack of semantic labels for GAN training.}
 \vspace{.5em}
 \caption{\textbf{Improving performance of computer vision tasks on real nuScenes~\cite{caesar2019nuscenes} images.} These tasks are object detection (YOLOv2~\cite{redmon2017yolo9000}), semantic segmentation (PSPNet~\cite{zhao2017pspnet}), and depth estimation (Monodepth2~\cite{godard2019digging}). The last line shows performance with the untuned models after the de-raining~\cite{liu2019dual} process.}
 \label{tab:impr_finetune}
\end{table}

\subsection{De-raining comparison}
\label{sec:deraining}

We now compare to the strategy of de-raining images first and then running un-tuned vision algorithms. To this end, we used the state-of-the-art de-raining method DualResNet~\cite{liu2019dual}, finetuned using nuScenes-clear augmented with GAN+PBR to accommodate for the domain gap.

During the de-raining fine-tuning process, random batches of $\left\{25, 50, 100, 200\right\}$mm/hr paired with their non-augmented counterpart are generated. Except for a smaller learning rate ($10^{-5}$), we used the DualResNet default hyper-parameters (Adam optimizer, batch and crop size of 40 and 64 respectively).

\begin{figure}
 \centering
 \subfloat[Object detection~\cite{redmon2017yolo9000}]{\includegraphics[width=0.485\linewidth]{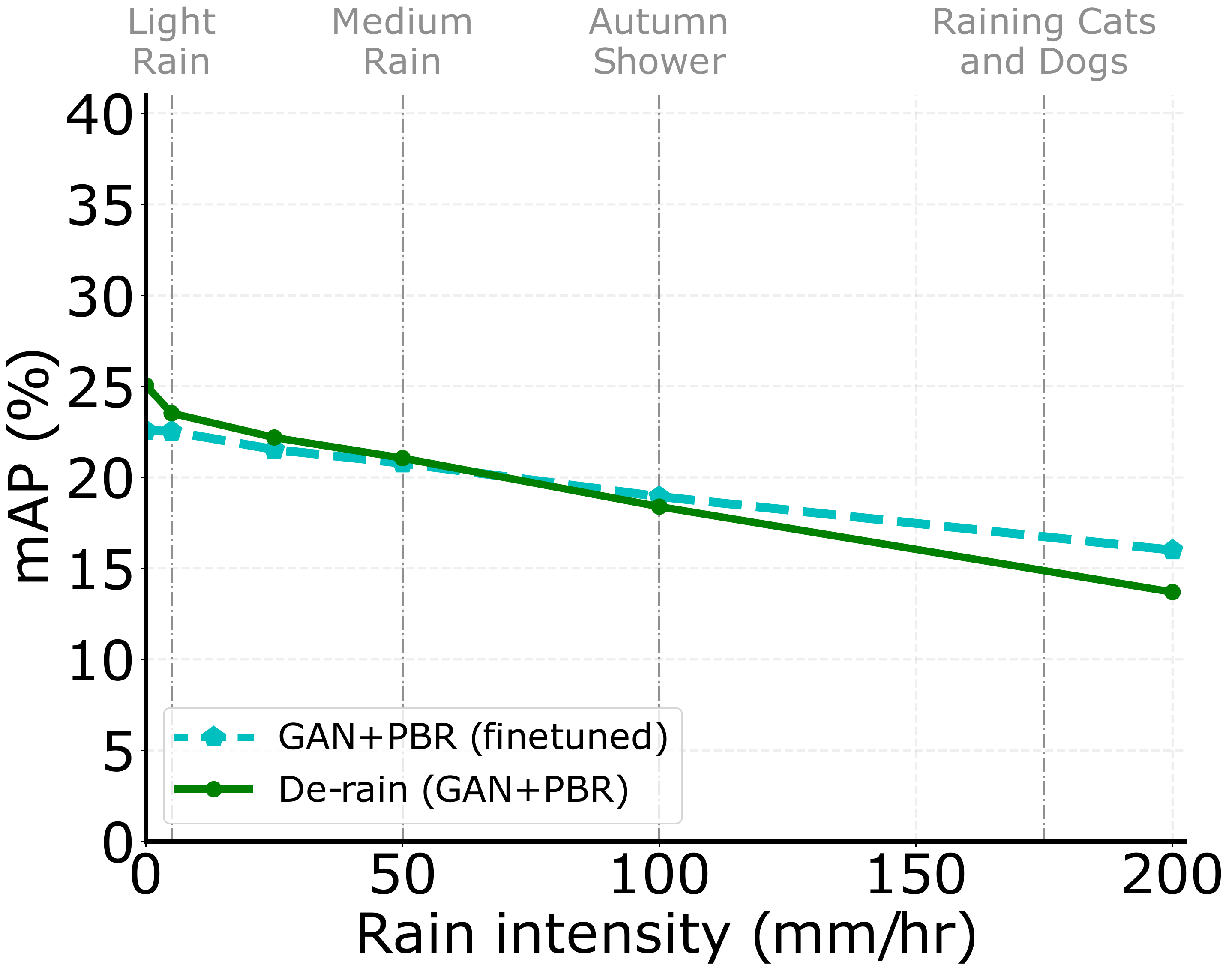} \label{fig:derain_obj_detect_perf}} \hfill%
    \subfloat[Depth estimation~\cite{godard2019digging}]{\includegraphics[width=0.485\linewidth]{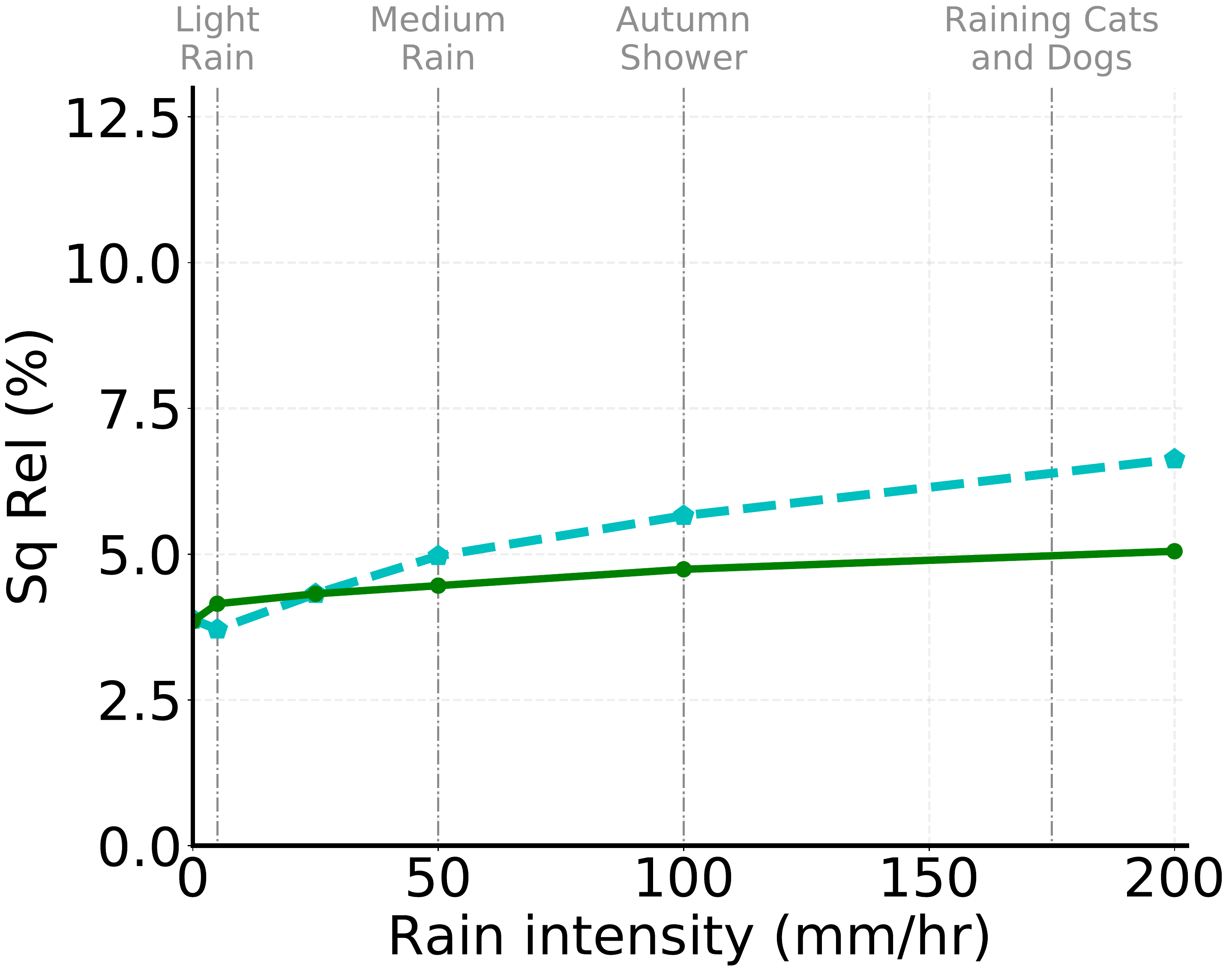} \label{fig:derain_depth_estim_perf}} 
 \caption{\textbf{Performance with varying rain intensities on de-rained GAN+PBR synthetic images.} The de-raining is performed with \cite{liu2019dual}. We see that the performance on both tasks decrease linearly for both \protect\subref{fig:derain_obj_detect_perf} object detection with YOLOv2~\cite{redmon2017yolo9000} and \protect\subref{fig:derain_depth_estim_perf} depth estimation with Monodepth2~\cite{godard2019digging}. Performance on both tasks are lower at low rain intensities (and the opposite at high rain intensities) compared to models finetuned with GAN+PBR synthetic images (cf. fig.~\ref{fig:synth_eval}).}
 \label{fig:deraining_synth_perf}
\end{figure}

With this de-raining finetuned model, we compare the performance of our ``untuned'' object detection (YOLOv2~\cite{redmon2017yolo9000}) and depth estimation (Monodepth2~\cite{godard2019digging}) models. Fig.~\ref{fig:deraining_synth_perf} shows the performance of the de-raining strategy compared to our ``rain-aware'' GAN+PBR finetuned models. Here, we observe that the rain-aware models offer improved performance for object detection over de-raining, while the latter improves depth estimation. This is likely due to the fact that streaks occlude the scene background, while de-raining acts as prior impainting thus easing depth estimation.

We also applied the same de-raining strategy to the real nuScenes images and report performance in the last row of table~\ref{tab:impr_finetune}. Again, for object detection on rainy images our rain-aware models perform better than de-raining. However, for depth estimation, the de-raining strategy is better for both clear and rainy images. This is consistent with the results obtained on synthetic data.

These experiments illustrate that de-raining is also a valid strategy that may even outperform ``rain-aware'' algorithms. However, this comes at the cost of having to perform two tasks, which may limit practical applications. On the long term, we believe rain-robust algorithms offer an exciting new research paradigm while avoiding the in-filling of occluded areas.

\section{Discussion}

In this paper, we presented the first intensity-controlled physical framework for augmenting existing image databases with realistic rain. This allows us to systematically study the impact of rain on existing computer vision algorithms on three important tasks: object detection, semantic segmentation, and depth estimation. 

\paragraph{Limitations and future work.}

While we demonstrated highly realistic rain rendering results, our approach still has limitations that set the stage for future work. 

For our PBR approach, the approximation of the lighting conditions~\ref{sec:rendering.fog-like} yielded reasonable results (fig.~\ref{fig:drop_cam_fov}), but it may under/over estimate the scene radiance when the sky is not/too visible. This approximation is more visible when streaks are imaged against a darker sky. More robust approaches for outdoor lighting estimation could potentially be used~\cite{holdgeoffroy-cvpr-19,zhang-cvpr-19}. 
Second, we make an explicit distinction between fog-like rain and drops imaged on more than 1 pixel, individually rendered as streaks. While this distinction is widely used in the literature~\cite{garg2007vision,garg2005does,de2012fast,Li_2018_ECCV}, it causes an inappropriate sharp distinction between fog-like and streaks. A possible solution would be to render all drops as streaks weighting them as a function of their imaging surface. However, our experiments show it comes at a prohibitive computation cost. 
Finally, rain streaks are added to image irrespective of the scene contents. Here, the depth estimate could be used to mask out streaks that appear behind objects and under the ground plane.
Another limitation is the computational cost of PBR. While this has no downside for benchmarking purpose as PBR may be run off-line, simulation requires increasing time with larger rainfall rates. With our current unoptimized implementation, the simulation of 1 / 25 / 50 / 100 mm/hr rainfall rates on Cityscapes requires 0.35 / 5.65 / 16.60 / 20.67 seconds respectively for the rain physics~\cite{de2012fast} and an additional 6.71 / 34.92 / 62.94 / 104.76 seconds for the rendering (times are per image, on single core, and averaged over 100 frames).
This restricts the usage of PBR to off-line processing though significant speed up could be obtained at the cost of additional optimization efforts.

The GAN employed also has limitations. First, while PBR is well-suited for videos since the rain simulator is temporally consistent, this is not the case for CycleGAN which does not guarantee temporal smoothness. Existing approaches such as \cite{bansal2018recycle} are alternatives, but GANs are known for their non-realistic physical outcome~\cite{xie2018tempogan}. Second, CycleGAN imposes a limit on the image resolution. Here, super-resolution networks such as SRResNet~\cite{dong2015image} or SRGAN~\cite{ledig2017photo} could potentially be used, or large-scale GANs such as BigGAN~\cite{brock2018large} are also an option. More importantly, GANs tend to have difficulty in generating rain on images of datasets different than which they are trained on since the learning process does not disentangle rain from scene appearance, demonstrating a strong domain dependence.

Finally, while our results demonstrate that fine-tuning on synthetically generated rain does improve performance on real rainy images (cf. sec.~\ref{sec:improving}), the improvements obtained are still quite modest. Further efforts are necessary to develop algorithms that are truly robust to challenging rainy conditions.

\section*{Acknowledgements}

This work was partially supported by the Service de coopération et d'action culturelle du Consulat général de France à Québec, as well as the FRQ-NT with the Samuel-de-Champlain grant. We gratefully thank Pierre Bourré for his priceless technical help, Aitor Gomez Torres for his initial input, and Srinivas Narasimhan for letting us reuse the physical simulator. We also thank the Nvidia corporation for the donation of the GPU used in this research.

\appendix
\section{Field of view of a drop in a sphere}
\label{sec:app-geom-drop-fov}

\begin{figure}
	\centering
	\includegraphics[width=1.0\linewidth]{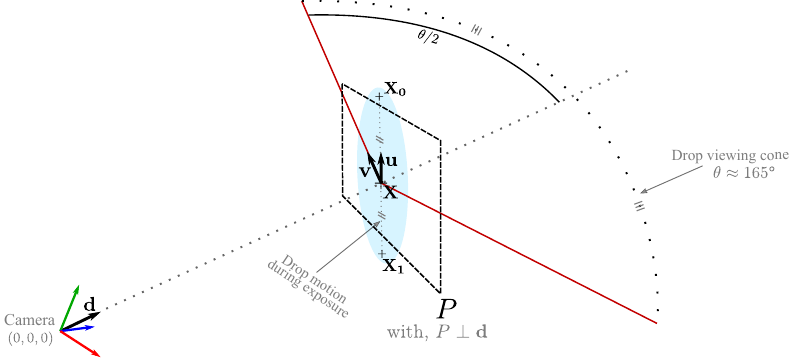}
	\caption{Geometrical construction to compute a drop FOV. Considering $\mathbf{X_0}$ and $\mathbf{X_1}$ the drop position at shutter opening and closing, respectively. We assume a constant drop position $\mathbf{X} = \frac{\mathbf{X_0}+\mathbf{X_1}}{2}$ (during the exposure time, a few milliseconds). Note that we drew only a slice of the drop FOV for simplicity but a full 3D visualization would show a full 3D cone. The drop FOV \textit{in} the environment map is the projection of the 3D drop FOV on the scene sphere of constant distance (refer to text for details).}
	\label{fig:drop_fov_geom}
\end{figure}

We estimate the field of view (FOV) of a drop when projected on a sphere to compute the radiance and chromaticity of each streak, as detailed in sec.~3.3 of the main paper. 
Despite its motion, we make the assumption of a constant field of view within a given exposure time. This is acceptable due to the short exposure time used here (i.e. 2ms for KITTI, 5ms for Cityscapes).
For each drop, the simulator outputs the start position (i.e. shutter opening) and end position (i.e. shutter closing) in both the 3D camera-centered and the 2D image coordinate frames.

We refer to the fig. \ref{fig:drop_fov_geom} for a geometrical illustration of the following.
Let us consider an imaged drop $D$, having 3D start position $\mathbf{X_0}$ and end position $\mathbf{X_1}$. We first compute $\mathbf{X} = \frac{\mathbf{X_0}+\mathbf{X_1}}{2}$ the \textit{assumed constant} position for which we will estimate the corresponding FOV. 
The position being camera-centered, the drop viewing direction is therefore $\mathbf{d}=\frac{\mathbf{X}}{||\mathbf{X}||}$. 

We compute the equation of the plane $P$ going through $\mathbf{X}$ and orthogonal to the viewing direction $\mathbf{d}$:
\begin{equation}
P = \mathbf{d}_x+\mathbf{d}_y+\mathbf{d}_z-\mathbf{d}\cdot\mathbf{X} = 0\,,
\end{equation}
where $\cdot$ is the dot product and select a random vector $\mathbf{u}$ (with $||\mathbf{u}|| = 1$) lying on $P$. 
Accounting for the field of view of the drop $\theta\approx165^\circ$ (according to \cite{garg2007vision}), we compute an arbitrary vector $\mathbf{v}$ on the viewing cone \textit{through} the drop
\begin{equation}
\mathbf{v} = \mathbf{d}\cdot\mathbf{R}_{\mathbf{u}}(\theta/2)\,,
\end{equation}
with $\mathbf{R}_{\mathbf{u}}(\theta/2)$ the 3x3 general rotation matrix of angle $\theta/2$ about vector $\mathbf{u}$. 
We use $\theta/2$ because the cone being symmetric along the viewing direction, the complete cone field of view obtain is $\theta$. 
The set $V'$ of vectors forming the viewing cone through the drop is obtained by the rotation of $\mathbf{v}$ all around the viewing direction. Formally,
\begin{equation}
V' = \{\mathbf{v}\cdot\mathbf{R}_{\mathbf{d}}(\alpha) ~| ~\forall ~\alpha \in [0, 2\pi[\}\,,
\end{equation}
with $\mathbf{R}_{\mathbf{d}}(\alpha)$ the rotation matrix of $\alpha$ around vector $\mathbf{d}$. In practice, $V'$ is a finite set of radially equidistant vectors (for computational reason we use $|V'| = 20$).

To compute the coordinates of the drop FOV in the environment, we assume a projection sphere $S$ of radius $10$m. Hence, we compute the set $Q = \{\phi(S, \mathbf{v'}) ~| ~\forall ~\mathbf{v'} \in V'\}$ of points where vectors intersect the environment sphere, considering only the positive viewing direction axis. 
Given that the sphere is centered to the camera position and all drops 3D positions are expressed in the camera referential, the intersection $\phi(S, \mathbf{v'})$ of a vector $\mathbf{v'}$ and sphere $S$ of radius $S_{\rho}$ is straight-forward with
\begin{equation}
\begin{split}
\phi(S, \mathbf{v'}) &= \mathbf{v'} + t\mathbf{d}\,\text{with},\\
t &= \frac{-b + \sqrt{b^2 - 4ac}}{2a}\,,\\
a &= \mathbf{d}_x^2 + \mathbf{d}_y^2 + \mathbf{d}_z^2\,,\\
b &= 2(\mathbf{d}_x\mathbf{v'}_x + \mathbf{d}_y\mathbf{v'}_y + \mathbf{d}_z\mathbf{v'}_z)\,,\\
c &= \mathbf{v'}_x^2 + \mathbf{v'}_y^2 + \mathbf{v'}_z^2 - S_{\rho}^2\,.
\end{split}
\label{eq:sphere-ray-intersection}
\end{equation}
Having computed $Q$, the finite set of 3D positions intersecting our environment sphere $S$, the set $Q'$ of spherical coordinates (azimuth, altitude) are obtained from simple Cartesian to spherical mapping, and directly translated to the environment map. 
Thus, $Q'$ is the projection of the drop FOV on the environment map.

Accounting for implementation details, one may note that $Q'$ is a discrete representation of the drop field of view contours. In practice, a polygon filling algorithm is used to obtain the drop FOV $F$, which we use for computing the photometry of a rainstreak (cf. sec.~3.3.2 of the main paper).

\section{Compositing a rain streak with different exposure time}
\label{sec:app-comp_diff_exp}

In their seminal work, Garg and Nayar~\cite{garg2005does} demonstrated that the streak appearance is closely related to the amount of time $\tau$ a drop stays on a pixel.
It is thus important to account for the difference of exposure time in the streak appearance database~\cite{garg2006photorealistic} when adding rain to existing images. 
Given that the appearance database does not provide enough calibration data to recompute the exact original $\tau_0$, we estimate it using observations made in appendix 10.3 of~\cite{garg2007vision}.
The latter states that for a constant exposure time $\tau$ can be safely approximated with $\sqrt{a}/50$ ($a$ the drop diameter, in meters), which we use to compute $\tau_0$ according to simulation settings in~\cite{garg2006photorealistic}.

Using the notation defined in eq.~(6) from the main paper, the radiance of streak $S'$ is corrected with
\begin{equation}
S'\frac{\tau_1}{\tau_0}\,,
\end{equation}
where $\tau_1$ is the time the current drop stays on a pixel, as obtained in a streak-wise fashion by the physical simulator. Noteworthy,~\cite{garg2007vision} also emphasizes that for a given streak the changes of $\tau$ across pixels are negligible, so $\tau$ can safely be assumed constant.

Finally, after normalization, the alpha of each streak is scaled according to $\tau_1$ and the targeted exposure time $T$. 
According to Garg and Nayar equations (cf. eq.~(18) from~\cite{garg2007vision}), the composite rainy image is an alpha blending of the background image $I_{bg}$ and the rain layer $I_r$. For pixel $\mathbf{x}$ corresponding to $\mathbf{x'}$ in the streak coordinates, it leads to:
\begin{equation}
\begin{split}
I_{rainy}(\mathbf{x}) &= \alpha{}I_{bg}(\mathbf{x}) + I_{r}(\mathbf{x'})\,,\\
 &= \frac{T-S'_{\alpha}(\mathbf{x'})\tau_1}{T}I(\mathbf{x}) + S'(\mathbf{x'})\frac{\tau_1}{\tau_0}\,.\\
\end{split}
\end{equation}
\section{Experiments data splits}
\label{sec:app-data-splits}

\begin{table}
\centering
\scriptsize
\setlength{\tabcolsep}{0.001\linewidth}
\renewcommand{\arraystretch}{1.25}
\begin{tabular}{L{0.225\linewidth}C{0.145\linewidth}C{0.184\linewidth}C{0.229\linewidth}C{0.209\linewidth}}
 \toprule
 &
  Kitti \cite{Geiger2012CVPR} & 
  Cityscapes \cite{Cordts2016Cityscapes} & 
  nuScenes-clear \cite{caesar2019nuscenes} & 
  nuScenes-rain \cite{caesar2019nuscenes} \\ 
 &
 7,481 & 
 2,995 & 
 24,134 & 
 4,996 \\
 \midrule
Effect of real rain (sec. \ref{sec:real_data}) &
\multirow{2}{*}{-} &
\multirow{2}{*}{-} &
\multirow{2}{*}{1,000 (test)} &
\multirow{2}{*}{*609 / 25 (test)} \\
Effect of synth. rain (sec. \ref{sec:impact_synth}) &
  \multirow{2}{*}{\textcolor{orange}{7,481 (test)}} &
  \multirow{2}{*}{\textcolor{orange}{2,995 (test)}} &
  \multirow{2}{*}{\textcolor{orange}{4,449 (test)}} &
  \multirow{2}{*}{-} \\
GAN training (sec.~\ref{sec:rain_setup}) &
 \multirow{2}{*}{-} &
 \multirow{2}{*}{-} &
 19,685 (train) 4,449 (test) &
 5,419 (train) 609~(test)\\
Improving synth. robustness (sec.~\ref{sec:perf_synth_data}) &
  \multirow{3}{*}{\shortstack[c]{\textcolor{orange}{1,000 (train)} \\ \textcolor{orange}{1,000 (test)}}} &
  \multirow{3}{*}{\shortstack[c]{\textcolor{orange}{1,000 (train)} \\ \textcolor{orange}{1,000 (test)}}} &
  \multirow{3}{*}{\shortstack[c]{\textcolor{orange}{1,000 (train)} \\ \textcolor{orange}{1,000 (test)}}} &
  \multirow{3}{*}{-} \\
Improving real robustness (sec.~\ref{sec:perf_real_data}) &
  \multirow{3}{*}{-} &
  \multirow{3}{*}{-} &
  \multirow{3}{*}{\shortstack[c]{\textcolor{orange}{1,000 (train)} \\ *1,000 / 25 (test)}} &
  \multirow{3}{*}{*609 / 25 (test)} \\
 \bottomrule
\end{tabular}
*Number of images for object\&depth / semantics
\caption{\textbf{Data splits for all experiments.} The splits are separated per context and datasets ; \textcolor{orange}{orange} indicates our rain augmented images. Notably, both the train and the test sets for improving the robustness to synthetic and real rain are sampled from the test set of the GAN training phase; this is valid since the computer tasks trained with these subsets have not seen images from neither set.}
\label{tab:splits}
\end{table}

Table~\ref{tab:splits} contains the minutiae of the data splits of the various experimental steps of this paper.

\end{document}